\documentclass[opre,nonblindrev]{informs4}

\OneAndAHalfSpacedXI 

\usepackage{enumerate}
\usepackage{url}
\usepackage{booktabs}
\usepackage{makecell}

\usepackage{tikz}
\usepackage{pgfplots}
\usepackage{pgfplotstable}
\pgfplotsset{compat=1.18}

\tikzset{
    chart/.style={
        legend label/.style={font={\scriptsize},anchor=west,align=left},
        legend box/.style={rectangle, draw, minimum size=5pt},
        axis/.style={black,semithick,->},
        axis label/.style={anchor=east,font={\tiny}},
    },
    bar chart/.style={
        chart,
        bar width/.code={
            \pgfmathparse{##1/2}
            \global\let\bar@w\pgfmathresult
        },
        bar/.style={very thick, draw=white},
        bar label/.style={font={\bf\small},anchor=north},
        bar value/.style={font={\footnotesize}},
        bar width=.75
    },
    pie chart/.style={
        chart,
    slice/.style={line cap=round, line join=round, very thick,draw=white},
    pie title/.style={font={\bfseries}},
    slice type/.style n args={3}{
        ##1/.style={pattern color=##2,pattern=##3},
        values of ##1/.style={}
        }
    }
}

\pgfdeclarelayer{background}
\pgfdeclarelayer{foreground}
\pgfsetlayers{background,main,foreground}
\usetikzlibrary{patterns}

\newcommand{\pie}[3][]{
    \begin{scope}[#1]
    \pgfmathsetmacro{\curA}{90}
    \pgfmathsetmacro{\r}{1}
    \def\c{(0,0)}
    \node[pie title] at (90:1.3) {#2};
    \foreach \v/\s in{#3}{
        \pgfmathsetmacro{\deltaA}{\v/100*360}
        \pgfmathsetmacro{\nextA}{\curA + \deltaA}
        \pgfmathsetmacro{\midA}{(\curA+\nextA)/2}

        \path[slice,\s] \c
            -- +(\curA:\r)
            arc (\curA:\nextA:\r)
            -- cycle;
        \pgfmathsetmacro{\d}{max((\deltaA * -(.5/50) + 1) , .5)}

        \global\let\curA\nextA
    }
    \end{scope}
}

\newcommand{\legend}[2][]{
    \begin{scope}[#1]
    \path
        \foreach \n/\s in {#2}
            {
                  ++(0,-10pt) node[\s,legend box] {} +(5pt,0) node[legend label] {\n}
            }
    ;
    \end{scope}
}

\tikzset{
  treenode/.style = {shape=rectangle,
                     draw, align=center},
  root/.style     = {treenode, font=\large},
  env/.style      = {treenode, font=\ttfamily\normalsize},
  dummy/.style    = {circle,draw}
}
\usepackage[breaklinks]{hyperref}
\usepackage{url}

\usepackage{float}
\usepackage[ruled,vlined,algo2e]{algorithm2e}
\usepackage{algpseudocode}
\usepackage{multirow}
\usepackage{subcaption}
\usepackage{makecell}
\usepackage{hyperref} 
\usepackage{bbm}
\usepackage{array}
\usepackage{cancel}
\usepackage{amsmath}
\usepackage{graphicx}
\usepackage{longtable}
\usetikzlibrary{shapes.geometric}

\newcolumntype{C}{>{$}c<{$}}
\newcolumntype{L}{>{$}l<{$}}
\newcolumntype{R}{>{$}r<{$}}
\newcommand{\bb}{ \mathbf{b} }

\newcommand{\bp}{ \mathbf{p} }

\newcommand{\bs}{ \mathbf{s} }

\newcommand{\bx}{ \mathbf{x} }

\newcommand{\by}{ \mathbf{y} }

\newcommand{\bz}{ \mathbf{z} }

\newcommand{\balpha}{ \boldsymbol{\alpha} }
\newcommand{\bgamma}{ \boldsymbol{\gamma} }
\newcommand{\brho}{\boldsymbol{\rho} }
\newcommand{\bmu}{ \boldsymbol{\mu} }

\newcommand{\bxi}{ \boldsymbol{\xi} }
\newcommand{\bbeta}{ \boldsymbol{\beta} }

\newcommand{\blambda}{ \boldsymbol{\lambda} }
\newcommand{\btheta}{ \boldsymbol{\theta} }

\definecolor{rosso}{RGB}{220,57,18}

\usepackage{natbib}
 \bibpunct[, ]{(}{)}{,}{a}{}{,}%
 \def\bibfont{\small}%
\TheoremsNumberedThrough     
\ECRepeatTheorems

\EquationsNumberedThrough    

\MANUSCRIPTNO{OPRE-2023-05-0248.R2} 

\begin{document}
\captionsetup[figure]{font=scriptsize}


\RUNAUTHOR{Adams et al.}

\RUNTITLE{Planning a Community Approach to Diabetes Care}

\TITLE{Planning a Community Approach to Diabetes Care in Low- and Middle-Income Countries Using Optimization}


\ARTICLEAUTHORS{%
\AUTHOR{Katherine B. Adams}
\AFF{Department of Operations and Analytics, University of Texas at San Antonio, San Antonio, TX 78249\\ \EMAIL{katherine.adams@utsa.edu}}
\AUTHOR{Justin J. Boutilier}
\AFF{Telfer School of Management, University of Ottawa, Ottawa, Canada, K1N 9B9 \\\EMAIL{boutilier@telfer.uottawa.ca}}

\AUTHOR{Sarang Deo}
\AFF{Max Institute of Healthcare Management, Indian School of Business, Gachibowli, Hyderabad, India, 50032\\ \EMAIL{sarang\_deo@isb.edu}}

\AUTHOR{Yonatan Mintz}
\AFF{Department of Industrial and Systems Engineering, University of Wisconsin-Madison, Madison, WI 53706\\ \EMAIL{ymintz@wisc.edu}} 

} 

\ABSTRACT{%

Diabetes is a global health priority, especially in low- and middle-income countries, where over 50\% of premature deaths are attributed to high blood glucose. Community Health Worker (CHW) programs can provide affordable and culturally tailored solutions for early detection and management of diabetes. We introduce an optimization framework to determine personalized CHW visits that maximize glycemic control at a community-level. Our framework explicitly models the trade-off between screening new patients and providing management visits to individuals who are enrolled in treatment. We account for patients' motivational states, which affect their decisions to enroll or drop out of treatment and, therefore, the effectiveness of the intervention. By estimating patients' health and motivational states, our model builds visit plans accounting for patients' tradeoffs when deciding to enroll in treatment, leading to reduced dropout rates and improved resource allocation. We apply our approach to generate CHW visit plans using operational data from urban slums in India. We find that our approach can reduce fasting blood glucose by up to 25\% with the same capacity as the best baseline method. Our experiments also demonstrate that our approach performs well with imperfect information.


}%



\maketitle




\vspace*{-2em}
\section{Introduction}\label{intro}

As of 2021, more than 460 million people are living with diabetes globally; a number that is projected to increase by 50\% in the next 20 years \citep{murray2020global}. Expanding access to diabetes screening and treatment is needed to meet the United Nations Sustainable Development Goal 3.4, which seeks to reduce premature mortality from noncommunicable diseases (NCDs) like diabetes by one third before 2030 \citep{cf2015transforming}. Diabetes disproportionately affects low- and middle-income countries (LMICs), where it is estimated than less than one in ten people with diabetes receive guideline-based diabetes treatment \citep{flood2021state}. In addition, it is estimated that 77.2\% of those with diabetes in LMICs have not achieved glycemic control \citep{manne2019health} and almost 90\% of individuals with undiagnosed diabetes reside in LMICs \citep{idf2021atlas}. The lack of a diagnosis often leads to poor glycemic control and, therefore, to increased risks of microvascular and macrovascular complications and mortality \citep{laiteerapong2019legacy}. Diabetes is a priority in urban areas in LMICs, where its prevalence is highest due to lifestyle changes driven by rapid urbanization \citep{ranasinghe2021prevalence}. 

Healthcare systems in LMICs lack the capacity required to deal with the burden of diabetes due to health workforce shortages \citep{scheffler2018forecasting}. However, there is growing evidence that workers with shorter training and fewer qualifications such as community health workers (CHWs) can reduce the burden of diabetes through early detection (e.g., screening) and case management (e.g., treatment initiation, education, etc.), especially for populations with low health literacy, limited access to health care, and poor social determinants of health \citep{jeet2017community, alaofe2017community, gyawali2021effectiveness}. This strategy has been proposed by the global health community and is known as task shifting, i.e., redistributing tasks from highly qualified professional health workers to CHWs for managing chronic diseases like diabetes \citep{world2007task}. Yet, there is an urgent need for increased research to improve the design and implementation of CHW programs so that they can be efficiently and effectively scaled-up to meet population needs in the long run \citep{world2018guideline}. 

Diabetes care is a challenging problem comprised of many different facets associated with detection, screening, and treatment of the disease. Due to the high rates of undiagnosed diabetes in LMICs, CHW programs must manage an explicit trade-off between providing screening visits (i.e., getting new patients on-board) and management visits (i.e., managing ongoing patients at monthly intervals). 
Regarding the provision of treatment, high patient attrition rates can lead to inefficiencies in planning and capacity allocation if unaccounted for by providers \citep{verevkina2014attrition, gucciardi2008factors}. When making treatment enrollment decisions, patients are affected by their internal motivational state, which plays an important role in initiating behaviors and selecting goal-oriented actions \citep{weinger2011effect, chew2014psychological}. Factors that can negatively impact patients' motivational states include social stigma, cognitive burden, economic burden, among others. Social stigma can lead patients to drop out of treatment, especially when receiving CHW home visits in communities where diabetes is stigmatized \citep{busza2018not, nyblade2017qualitative}. Similarly, the cognitive burden of treatment, which requires patients to learn how to manage their condition and implement changes in health and lifestyle behavior, may overwhelm patients and lead to drop outs \citep{heckman2015treatment}. Often, the effects of these adverse factors of enrollment are substantial enough to thwart potential health improvements provided by CHW interventions.

To jointly tackle the provider's screening and management visit decisions and patient enrollment decisions, we propose a novel optimization framework for coordinating CHW visits in the context of a behavioral diabetes intervention for urban environments in LMICs. Our framework is composed of a model for how patients may enroll in the program and a dynamic programming (DP) approach for finding the optimal intervention (visit) policy. By modeling patients as utility-maximizing agents that weigh beneficial and adverse treatment factors, providers can better predict the impact of their visit plans on patient enrollment and reduce drop out rates. 

In this paper, we develop an optimization framework for allocating CHW visits to patients in the context of behavioral diabetes interventions in LMICs. Through developing this framework, we provide three major contributions: 
\begin{enumerate}
    \item We extend existing models for chronic care by personalizing visits based on how patients interact with behavioral interventions delivered by CHWs (Section~\ref{sec:patient_model}). Our modeling approach captures both the disease progression of patients and their individual enrollment decisions based on their physical health state and their motivational state. While this model has several unobserved values, we provide a maximum likelihood method for estimating these quantities using basic data commonly collected by CHWs. We conduct numerical experiments using real data to demonstrate that the out-of-sample performance of our approach is comparable with leading machine learning methods (\ref{EC:incontrolprediction}).
   
    \item We develop a DP approach that leverages the patient model to allocate CHW visits such that the number of patients that have their blood glucose levels in control is maximized (Section~\ref{sec:provider_model}). Our model is designed for resource-constrained settings where the provider may need to trade-off managing patients that are currently enrolled and screening new patients to encourage them to join the intervention. We fully characterize the optimal policy for this model in  the single-patient case and examine the structure of the policy in the multi-patient resource constrained case. In the single-patient case, we establish a structural relationship between patient enrollment and glycemic control. For the multi-patient case, we provide a bound and prove convergence for a Lagrangian relaxation based approach and we prove the existence of a Whittle's Index type approach. We use these results to develop a diverse set of approximate DP approaches that can be implemented in practice and provide a tradeoff between equity and efficiency.
    
    \item We conduct a comprehensive set of experiments that validate our modeling and optimization framework using real-world data obtained from a CHW program that operates in Hyderabad, India (Section~\ref{sec:experiments}). We further assess the generalizability of our approach using simulated patient cohorts. Our experiments show that our approach can maintain similar performance to baseline methods with up to 50\% less capacity. For the case study with real data, our approach is able to reduce blood glucose by up to 25\% with the same capacity as baseline methods. Finally, we conduct a set of experiments to demonstrate that our approach performs well with imperfect information.

\end{enumerate}

\section{Literature}

Our work contributes to three major streams of literature: sequential resource allocation (Section~\ref{lit:SRA}), personalized healthcare (Section~\ref{lit:ph}), and global health operations (Section~\ref{lit:gho}). 

\subsection{Sequential Decision-Making for Resource Allocation}
\label{lit:SRA}

Sequential decision-making for resource allocation has been studied outside of health care for applications including retail operations \citep[e.g.,][]{acimovic2015making, farias2011irrevocable}, product development \citep[e.g.,][]{bhaskaran2021sequential, kornish2001pricing}, operations scheduling/reservations \citep[e.g.,][]{keyvanshokooh2021online, stein2020advance}, and machine maintenance \citep[e.g.,][]{glazebrook2006some, ruiz2020multi}. Within healthcare, authors have studied applications such as 
hepatitis C treatment in prisons \citep{ayer2019prioritizing}, planning community-based chronic disease care \citep{deo2013improving}, hepatocellular carcinoma screening \citep{lee2019optimal}, breast cancer screening \citep{ayvaci2012effect}, adherence to tuberculosis treatment \citep{mate2020collapsing, mate2021risk}, and HIV treatment under supply uncertainty \citep{deo2022optimal}.

There are similarities between our problem and the machine maintenance problem (e.g., machines deteriorate (disease progression) and require maintenance (management) visits). An important difference is that our setting requires modeling individual behavior (i.e., enrollment decisions) when building visit plans. Regarding healthcare applications, our work differs from prior literature in two ways. First, we explicitly model the trade-off between revisiting patients enrolled in treatment and screening new patients who are not enrolled in treatment. 
We note that \cite{deo2022optimal} also explored the trade-off in health outcomes when making operational decisions such as treatment initiation and continuity. In contrast to their work, we include a behavioral model within our resource allocation problem that accounts for patient decisions using a utility-maximizing framework. This approach accommodates not only heterogeneous patients in terms of disease progression and treatment effectiveness, but also their agency when making treatment decisions.





\subsection{Personalized Healthcare}
\label{lit:ph}


Personalized healthcare allows decision-makers to tailor care while accounting for heterogeneous disease incidence, progression, and varying efficacy of a range of prevention and treatment strategies. Early work sought to improve treatment decisions and outcomes by grouping patients by health state \citep[e.g.,][]{deo2015planning}, while more recent work often models disease progression and treatment effects at the individual level. Among common methodologies, prior research has applied Markov decision processes to optimize chronic disease screening \citep[e.g.,][]{skandari2021patient,hajjar2022personalized} and multi-armed bandit models for treating multiple myeloma, type 2 diabetes, multiple sclerosis, Warfarin dosing, and physical activity interventions \citep{negoescu2018dynamic, keyvanshokooh2019contextual, wang2019tumor, bastani2020online, mintz2020nonstationary}. Personalizing care for chronic diseases involves challenges related to the use of observational data \citep{denton2018optimization}; even so, several articles in the literature have identified policies to improve health outcomes and reduce expenditures compared to current guidelines. Examples include screening for cancer \citep{petousis2019using, alagoz2011optimizing, scherrer2015breast}, hypertension \citep{schell2016data}, diabetes \citep{wu2022optimizing}, and chronic kidney disease \citep{wu2024examining}. 
While optimizing screening guidelines is an important measure for several conditions, diabetes requires ongoing treatment that often spans several decades, leading to the need to model not only screening decisions, but also treatment. In contrast with models that personalize care to find the best care plan for each patient individually and with a single visit type, our model solves a personalized health care problem under resource-limited settings where a provider may need to trade-off offering care for different patients due to capacity constraints$-$including trading-off screening potential new patients and managing patients who are already enrolled in treatment. Additionally, our approach considers patients' agency, which manifests through their decisions to participate in the intervention (and impacts their outcomes). 

Regarding the second point, \cite{aswani2019behavioral} characterized patients’ responses in the context of a weight loss intervention. Similar to our approach, their framework used utility functions to model patients’ motivational states, allowing them to predict weight changes as a function of the step goals set for each individual and, therefore, set goals that optimized weight loss. \cite{mintz2023behavioral} used a similar behavioral analytics framework for online capacitated multi-agent systems where a single coordinator provides behavioral or financial incentives for each agent using a finite budget. The coordinator's objective depends on the states and decisions of all agents, whose utility functions are initially unknown.
In contrast to their work, our setting involves more than one visit type (the equivalent of their incentive), which requires the provider to split their resources between screening and management visits. These visit decisions affect not only patients' motivational states, but also the patient pool size that must be managed by the provider. 



\subsection{Global Health Operations}
\label{lit:gho}
Another related stream of literature applies optimization techniques to global health in resource-limited settings \citep[e.g.,][]{jonasson2017improving,parvin2018distribution,boutilier2020ambulance,boutilier2021improving}. 
Of particular interest is the stream of literature that applies optimization techniques to improve CHW interventions in LMICs. In this context, most prior research has focused on developing routing models to optimize CHW visits \citep[e.g.,][]{brunskill2010routing,cherkesly2019community}. Our approach does not include routing decisions because our study setting is urban slums with high population density where CHWs walk over short distances between visits. Additionally, our intervention is planned over medium- to long-term and involves tracking patients over time to optimize their health outcomes, which are explicitly modeled in our framework. 

\section{Study Context}

In this section, we provide additional details about our problem setting that are relevant to understand the modeling assumptions made in subsequent sections.

\subsection{Diabetes}

Diabetes is a metabolic disorder characterized by insufficient insulin production (decline in beta-cell function) or ineffective use of insulin (insulin resistance). These issues affect the body's ability to metabolize glucose for energy and lead to glucose accumulation in the bloodstream \citep{wysham2020beta}. Type 2 diabetes comprises 95\% of all diabetes cases and is diagnosed by measuring the blood glucose level with one of several metrics, including random blood glucose, oral glucose tolerance test, post-prandial blood glucose, fasting blood glucose (FBG), and HbA1c.

Upon diagnosis, several strategies can be used to care for diabetes patients, including prescribing medication, educating patients, and promoting healthy lifestyle choices. Many of these measures require providing health services at the individual level, a challenging issue for healthcare systems in LMICs due to lack of capacity caused by health workforce shortages \citep{scheffler2018forecasting}. Fortunately, some of these measures do not require medical training and can be effectively performed by CHWs \citep{world2007task}, who can be sought out from the communities that they will serve (allowing them to provide culturally-tailored care) and receive short training from the health system.

The literature indicates that a patient's adherence to their prescribed treatment and incorporation of self-care recommendations are impacted by psychosocial factors, which include social, psychological, and environmental influences to a person's health and well-being \citep{weinger2011effect}. In lieu of the terminology ``psychosocial factors" (which incorporates external factors), we focus on the patients and refer to their internal drive to treat their diabetes as their motivational state. CHWs can help patients increase their motivational state by emotionally supporting them and instilling in them ``a positive outlook, sense of resilience and well-being" \citep{chew2014psychological}. Furthermore, a Randomized Clinical Trial has shown that teaching diabetes patients cognitive behavioral strategies was more effective for glycemic control than group or individual education programs \citep{weinger2011effect}. This finding supports the need for long-term interventions where patients can be taught how medication use, diet, and exercise affect their blood glucose, how to set and achieve goals, and how to problem-solve when obstacles to self-care behavior implementation appear \citep{weinger2011effect}. Among adverse factors of treatment that can affect patients' motivational states and therefore impact their enrollment decisions and clinical outcomes are
social stigma and cognitive burden. The use of stigmatizing phrases to describe people with diabetes has been reported in several countries and includes terms such as ``sick and disabled", ``contagious", ``self-inflicting", among others \citep{abdoli2018discussion} -- a depiction of the stigma that can lead patients to refuse CHW home visits \citep{busza2018not, nyblade2017qualitative}. Therefore, CHW programs must model patient health outcomes and account for patient enrollment behavior to determine the optimal timing and type of visit to provide to each patient so that programs most effectively use their limited resources.

\subsection{Diabetes in India}

India is the country with the second highest burden of diabetes in the world -- an estimated 74.2 million people \citep{lin2020global}, with a higher prevalence in urban than in rural areas \citep{ranasinghe2021prevalence}. Between 2017 and 2018, the National Noncommunicable Disease Monitoring Survey (NNMS) collected data on NCD risks (including anthropometric measurements) and health-seeking behaviors from a national sample of adults from 12,000 households. Through the NNMS, researchers determined that the prevalence of type 2 diabetes and impaired FBG in India was 9.3\% and 24.5\% respectively. Among those with diabetes, 45.8\% were aware, 36.1\% were on treatment and 15.7\% had their FBG in control \citep{mathur2022prevalence}. Researchers estimate that the number of individuals with diabetes is projected to increase by 68\% by 2045 and that over 53\% of individuals living with diabetes are undiagnosed \citep{idf2021atlas}.

In 2010, the Government of India launched the National Program for Prevention and Control of Cancer, Diabetes, Cardiovascular diseases and Stroke to better prepare the country for the rise in NCD prevalence. However, a recent study that included a nationally representative survey found that both private and public health facilities are not adequately prepared to handle the burden of NCDs. By assessing the availability of trained human resources, essential medicines and technologies for
diabetes, cardiovascular and chronic respiratory diseases, the study found that the percentage of primary care facilities capable of managing these three NCDs varied between 1.1\% in rural public to 9.0\% in urban private facilities \citep{krishnan2021preparedness}.


Regarding the economic impact of diabetes to individuals, estimates of the mean expenditure on diabetes and its complications in India range between 15,535 and 76,279 Indian Rupees (INR) (209.30-1017.05 USD) per year \citep{sathyanath2022economic,kazibwe2021household}. On average, expenditures from outpatient services constituted approximately 3-5\% of individual annual incomes. However, this number increased to 21\% for those experiencing complications or requiring hospitalization, representing a catastrophic health expenditure \citep{sathyanath2022economic}. The high out-of-pocket expenditure is explained by the high percentage of health care provided by the private sector (over 70\%) \citep{kumar_2022} and a very small percentage of private pre-paid insurance spending (2.4\%) \citep{dieleman2017evolution}. At a national level, researchers estimate a loss to the Gross Domestic Product of 176.6 trillion INR (2.6 trillion USD; 9.8 trillion USD in terms of purchasing power parity) due to lost productivity in 2017 alone \citep{banker2021impact}. 









\subsection{Diabetes Care Provider -- NanoHealth}
\label{sec:context-nanohealth}

We collaborated with and obtained data from NanoHealth, a social enterprise and former Hult Prize\footnote{The Hult Prize is an annual, year-long competition that crowd-sources ideas from university level students after challenging them to solve a pressing social issue around topics such as food security, water access, energy, and education.} winner based in Hyderabad, India. Hyderabad is the capital of the state of Telangana and the fourth largest city in India with a population of 7 million \citep{census:2011}, including more than 1.7 million people living in 1400 urban slums \citep{hyderabad2012slums, telangana2015}. NanoHealth operates a diabetes screening and management program in low-income households in urban slums. The population they serve comprises individuals in the unorganized sector of the economy working as drivers, daily wage earners, domestic helpers, vendors, and self-employed professionals. The average family income of these residents ranges between 15,000 and 30,000 INR per month, which is equivalent to 200-400 USD. For reference, the average monthly family income in India is estimated to be 23,000 INR (307 USD) \citep{choudhury_2022}. 

NanoHealth employs a team of CHWs, referred to as \emph{Saathis} (Hindi for companion), who are members of the communities they serve and have received basic training on diabetes care. When the data used in this paper were collected (2015-2018), NanoHealth operated in 52 urban slums (and surrounding communities) and had an average of 30 active CHWs per day providing screening visits to new patients and monthly follow-up visits to those enrolled in their treatment plan. CHWs were assigned densely populated catchment areas with a population of approximately 5,000, which can be easily covered on foot to conduct both screening and management visits. Each CHW conducted an average of 144 visits per month ($\approx$5 per day) and was equipped with a ``Doc-in-the-Bag” kit that included a weighing scale, measuring tape, blood glucose monitor, and blood pressure/heart rate cuffs. A mobile tablet was used to record patients’ responses to a questionnaire about lifestyle, demographics, symptoms of common ailments, and to record certain anthropomorphic measurements and vitals \citep{deo2021community}. Screening visits were provided to identify patients with diabetes who were undiagnosed, while management visits were provided to patients who were screened, identified as high risk based on the anthropometric measurements collected, and decided to enroll in NanoHealth's treatment program \citep{boutilier2021risk} 

\section{Patient Model}
\label{sec:patient_model}
In this section, we describe the model of prospective and current patients' decisions to enroll or remain enrolled in the CHW-based intervention. We assume that each patient makes these decisions based on the perceived utility (reduction in FBG) and disutility (adverse factors such as social stigma and cognitive burden weighted by their perceived importance) of staying enrolled. 

Let $\mathcal{P}$ be the set of patients that can be enrolled or are already enrolled in the CHW intervention and let $i \in \mathcal{P}$ denote the index of a particular patient. Let the time index be given by $t \in \mathcal{T} = \{0,...,N\}$. We let $z_{i,t}$ indicate the enrollment status (binary) of patient $i$ at time $t$, where $z_{i,t}=1$ indicates that the patient is enrolled in the intervention. Screening visits are only conducted for patients where $z_{i,t}=0$; while management visits are only conducted for patients where $z_{i,t}=1$. At each period $t$, the provider's decisions -- whether or not to visit a patient -- are denoted by the indicator variables $y_{i,t}$, which equal 1 if the provider chooses to send a CHW to visit patient $i$ at time $t$ and 0 otherwise. This variable represents a screening visit for an unenrolled patient and a management visit for an enrolled patient. 

We assume that the states and actions occur in the following order.
Within each period $t$, we assume that the decision to visit a patient $i$, $y_{i,t}$, precedes the patient's enrollment decision in that same period, $z_{i,t}$. This assumption implies that the provider can influence $z_{i,t}$ through $y_{i,t}$, potentially affecting patient $i$'s enrollment decision within the same period. Updates to patient states occur in the following period and can be impacted by both $y_{i,t}$ and $z_{i,t}$. 

We model the disease state using the log of patients' FBG, denoted by $b_{i,t} \in \mathcal{B}$, where $\mathcal{B} \subset \mathbb{R}_+$ is a closed and bounded interval. We use the FBG as our primary metric because it provides a suitable trade-off between diagnostic accuracy and practical feasibility, and is the metric most commonly used in LMICs \citep{hoyer2018utility, zhao2013fasting}. Measuring the FBG requires a fast (i.e., only water consumption) for 8 to 12 hours, which is typically done overnight. The log transform is used to capture the fact that the disease progression is proportional to the current FBG level \citep{derendorf1999modeling,lee2018outcome}. 
We assume that $b_{i,t}$ evolves according to a set of linear dynamics of the form: 
\begin{equation}
    b_{i,t+1} = b_{i,t} + p_i -\mu_i z_{i,t} - \alpha_i y_{i,t} z_{i,t} + \xi_{i,t},
    \label{eq:fbg_dynamics}
\end{equation}
where $p_i \in \mathcal{B}$ represents the disease progression in one time period, $\mu_i\in \mathcal{B}$ represents the impact of enrollment (on FBG), $\alpha_i\in \mathcal{B}$ represents the impact of a management visit (on FBG), and $\{\xi_{i,t}\}_{t\in \mathcal{T}}$ are i.i.d disturbance terms such that $\mathbb{E}\xi_{i,t} = 0$ and $\mathbb{E}\xi_{i,t}^2 < \infty$. To rule out the trivial case of an ineffective intervention (which does not reduce patients' FBG), we assume that $p_i < \mu_i + \alpha_i$. Supporting this assumption, a retrospective study found that a community-based diabetes management intervention successfully induced a retention-mediated reduction in FBG, indicating that the magnitude of the combined enrollment and management visit effects is robust enough to consistently prevent the FBG from increasing beyond the benefits they provide \citep{deo2021community}. To interpret the dynamics equation, let $\tilde{b}_{i,t} = \exp(b_{i,t})$ be the true FBG. Then, we can write the dynamics as $\tilde{b}_{i,t+1} = \tilde{b}_{i,t} \exp(p_i -\mu_i z_{i,t} - \alpha_i y_{i,t} z_{i,t} + \xi_{i,t})$. Consider the case where $p_i > 0$, $z_{i,t}=0$, $y_{i,t}=0$, and $\xi_{i,t}=0$, which implies that a patient's rate of change in FBG $\Tilde{b}_{i,t+1}/\Tilde{b}_{i,t}$ is equal to $\exp(p_i)$. Therefore, for an unenrolled patient whose enrollment or visit status don't change, the rate of change in the FBG will be constant and their FBG will be equal to $\Tilde{b}_{i,t+n}=\Tilde{b}_{i,t}\exp(np_i)$ after $n$ periods. 

Next, we consider adverse factors experienced by enrolled patients such as the social stigma of enrollment or the cognitive burden of participation \citep{busza2018not,heckman2015treatment}. To develop these model components, we draw from planned behavior theory (PBT), a social-cognitive model that has been effective in predicting several health behaviors, including diet, physical activity, condom use, drug use, and health screening behaviors \citep{rich2015theory, vo2015testing, roncancio2015using}. In the first meta-analysis applying PBT to predict treatment adherence in people with chronic diseases, \cite{rich2015theory} found that PBT accounted for 33\% of the variance in intentions and 9\% in the variance of behavior related to treatment adherence. Perceived behavioral control was the strongest predictor of intention, indicating that a patient's perception of their self-efficacy is related to their resulting behaviors. Furthermore, authors have identified an ``attenuating effect of past behavior on theory relations" \citep{hagger2016using}, which informs our choice of using discount factors for adverse factors updates. 

Let $s_{i,t} \in \mathcal{S}$ represent the adverse factors of enrollment experienced by patient $i$ by period $t$, where $\mathcal{S} \subset \mathbb{R}_+$ is a closed and bounded interval, and $s_{i,0} \in \mathcal{S}$ represent the baseline level of adversity that a patient will experience from participating in the intervention. We assume that this adversity level evolves according to the following linear dynamics: 
\begin{equation}
   s_{i,t+1} = z_{i,t}(\gamma(s_{i,t}-s_{i,0}) + s_{i,0}) + \beta_i y_{i,t} z_{i,t}, \label{eq:disutil_dynamics}
\end{equation}
where $\gamma \in (0,1)$ is a discount factor on the previous period's adversity, and $\beta_i \in \mathcal{S}$ is the impact on adverse factors from a management visit. Intuitively, this equation states that the adversity of a patient will increase when they are visited by a CHW, but the adverse impact of a visit will decay exponentially over time. Figure~\ref{fig:adv_factors1} illustrates this effect, where there is an increase in adverse factors in period 1 after a management visit is received in period 0, but eventually $s$ returns to the baseline level $s_0$ (in the case where no other management visits are received). The rate of exponential decay depends on $\gamma$, with lower values of $\gamma$ leading to a faster decay. We use this structure because it is reflective of how being visited by a CHW may increase social stigma as perceived by the community \citep{abdoli2018discussion, busza2018not}, but as time progresses, the memory (and thus the associated stigma) of the visit will fade \citep{ritchie2015pancultural}.

\begin{figure}[!ht]
    \centering
\begin{subfigure}{0.48\textwidth}
    \centering
\pgfplotstableread[col sep=comma,]{adverse_factors_data.csv}\datatable
\begin{tikzpicture}
\begin{axis}[
    width=\textwidth,
    height=5.5cm,
    xtick=data,
    xticklabels from table={\datatable}{Period},
    ticklabel style = {font=\footnotesize},
    ylabel near ticks,
    ylabel style={inner sep=1pt},
    xlabel style={font=\footnotesize},
    legend style={at={(0.97,0.48)},anchor=south east},
    legend style={font=\footnotesize},
    ylabel={Adverse factors ($s$)}, 
    xlabel={Period ($t$)}]
    
    \addplot [mark=o, blue!80 ] table [x expr=\coordindex, y={Example 1}]{\datatable};
    \addlegendentry{$s_0=0.5, \: \beta=0.5$}
    
    \addplot [mark=square, red!80] table [x expr=\coordindex, y={Example 2}]{\datatable};
    \addlegendentry{$s_0=0.5, \: \beta=1$}
    
    \addplot [mark=diamond, black!50 ] table [x expr=\coordindex, y={Example 3}]{\datatable};
    \addlegendentry{$s_0=1, \: \beta=0.5$}
    
    \addplot [mark=triangle, violet!80] table [x expr=\coordindex, y={Example 4}]{\datatable};
    \addlegendentry{$s_0=1, \: \beta=1$}
\end{axis}
\end{tikzpicture}
\caption{}
\label{fig:adv_factors1}
\end{subfigure}
\begin{subfigure}{0.48\textwidth}
    \centering
\pgfplotstableread[col sep=comma,]{perception_adverse_factors_data.csv}\datatable
\begin{tikzpicture}
\begin{axis}[
    width=\textwidth,
    height=5.5cm,
    xtick=data,
    xticklabels from table={\datatable}{Period},
    ylabel near ticks,
    ylabel style={inner sep=0.0pt},
    legend style={at={(0.97,0.05)},anchor=south east},
    legend style={font=\footnotesize},
    ticklabel style = {font=\footnotesize},
    xlabel style={font=\footnotesize},
    ylabel={Perc. of adv. factors ($\theta$)},
    xlabel={Period ($t$)}]
    
    \addplot [mark=o, orange!90!black] table [x expr=\coordindex, y={Example 1}]{\datatable};
    \addlegendentry{$\theta_0=0.5, \: \lambda=0.5$}
    
    \addplot [mark=square, green!40!black] table [x expr=\coordindex, y={Example 2}]{\datatable};
    \addlegendentry{$\theta_0=0.5, \: \lambda=1$}
    
    \addplot [mark=diamond, red!80 ] table [x expr=\coordindex, y={Example 3}]{\datatable};
    \addlegendentry{$\theta_0=1, \: \lambda=0.5$}
    
    \addplot [mark=triangle, teal] table [x expr=\coordindex, y={Example 4}]{\datatable};
    \addlegendentry{$\theta_0=1, \: \lambda=1$}
\end{axis}
\end{tikzpicture}
\caption{}
\label{fig:adv_factors2}
\end{subfigure}
\vspace{5pt}
\caption{(a) Illustration of the change in adverse factors ($s$) after a management visit in period 0 for varying $s_0$ and $\beta$ (for $\gamma=0.2$); (b) Illustration of the change in perceived importance of adverse factors ($\theta$) after a management visit in period 0 for varying $\theta_0$ and $\lambda$ (for $\rho=0.2$).}
\end{figure}

An important aspect of diabetes management is educating patients on disease progression and healthy lifestyle choices \citep{nazar2016effectiveness}. As such, we assume that a patient's perception of the adverse factors of enrollment may decrease with the number of management visits received since enrolling in the intervention \citep{fisher2014impact}. To model this, we introduce the state $\theta_{i,t} \in \Theta$, where $\Theta\subset \mathbb{R}_+$ is a closed interval, to represent the patient's perceived importance of adverse factors and assume that $\theta_{i,t}$ evolves according to the following linear dynamic equation:
\begin{equation}
    \theta_{i,t+1} = \rho (\theta_{i,t} - \theta_{i,0}) + \theta_{i,0} - \lambda_i y_{i,t} z_{i,t}, \label{eq:percep_dynamics}
\end{equation}
where $\theta_{i,0} \in \Theta$ denotes the steady-state level of perceived importance of adverse factors for patient $i$, $\lambda_i \in \Theta$ denotes the decrease in the perceived importance of adverse factors that occurs each time a management visit is received as shown in Figure~\ref{fig:adv_factors2}, and $\rho \in (0,1)$ is a discount factor on the previous period's perception. Even though the actual adverse factors of enrollment, $s_{i,t}$, may increase when a management visit is received, we assume that the patient's understanding of the importance of treatment and the reassurance provided by the CHWs will decrease the relevance of this factor each time a management visit is received. In the following section, the actual disutility of enrollment experienced by a patient is implemented through the product of $\theta_{t}$ (their current perceived importance of adverse factors) and $s_{t+1}$ (adverse factors of enrollment in the following period).


\subsection{Patient Utility Function and Enrollment Decision}
%

We assume that patients have knowledge of their FBG level ($b_{i,t}$), adverse factors of enrollment ($s_{i,t}$), and the perceived importance of adverse factors ($\theta_{i,t}$) when making enrollment decisions 
because patients' enrollment depends on subjective notions of the utilities associated with their health and adverse treatment factors when enrolled compared to when not enrolled \citep{paige2016patient}. We use this assumption to predict future patient enrollment through a framework where patients weigh the pros and cons based on the relative importance given to a potential FBG improvement (and the corresponding improvement in quality of life) at the cost of increased social stigma/cognitive burden. While our baseline model assumes that patients are aware of their own health status and perceived adverse factors, this assumption is not essential to the algorithm's effectiveness. We show in Section~\ref{sec:exp_w_imperfect_info} and in Appendices~\ref{EC:clusteringprediction} and \ref{EC:imperfinfo} that our proposed approach remains robust even when these inputs are noisy or partially unknown, suggesting that the method can generalize to more realistic behavioral settings.

Patients are also assumed to make decisions based on a myopic utility maximizing framework. Some social scientists believe that these decisions are rational with respect to heavily discounted future outcomes; an effect that has been previously observed in healthcare decision-making \citep{cawley2004economic}. Moreover, previous work \citep{aswani2019behavioral,mintz2023behavioral} has shown that despite making these assumptions, resulting models still have strong predictive and prescriptive performance. Therefore, we assume that patients only consider their benefit of enrollment in the immediate next period, $t+1$, when deciding whether to enroll/stay enrolled in period $t$. 

We denote the utility function of patient $i$ by $U_i:\mathbb{R}^4\times \{0,1\} \mapsto \mathbb{R}$. We assume that each patient's utility function has the structure $U_i = U_{i,b} + U_{i,s}$, where $U_{i,b}$ is the net utility associated with the state $b_{i,t}$ and $U_{i,s}$ is the net utility associated with the state $s_{i,t}$. This additive decomposition reflects the assumption -- common in health behavior models -- that individuals evaluate benefits and burdens as separable components of decision-making, consistent with frameworks like the Health Belief Model and cost-utility analysis conventions \citep{koulouvari2025applications, keeney1993decisions}. An alternative model would be to consider multiplicative utilities $\tilde{U}_i = \tilde{U}_{i,b}\tilde{U}_{i,s}$, however this can still be reformulated to the additive structure we propose by interpreting $\tilde{U}_i := \exp{(U_i)},\tilde{U}_{i,b}:= \exp{(U_{i,b})}, \tilde{U}_{i,s}:=\exp{(U_{i,s})} $, that is thinking of the patient's objective being additive in the log utilities. Thus we focus our modeling on the additive case for the remainder of the paper.
Specifically, we assume that these functions are of the form:
\begin{align}
   & U_{i,b}(b_{i,t},z_{i,t},\xi_{i,t}) = -b_{i,t+1}(b_{i,t},z_{i,t}, \xi_{i,t};\:y_{i,t}),  \label{eq:fbg_util}\\
    &U_{i,s}(\theta_{i,t}, s_{i,t+1},z_{i,t}) = -\theta_{i,t} s_{i,t+1}(s_{i,t},z_{i,t};\:y_{i,t}). \label{eq:s_util}
\end{align}
Note that $U_{i,b}$ and $U_{i,s}$ are functions of future patient states because the patient's decisions (and the intervention) can only impact their future states. Intuitively, $U_{i,b}$ implies that patients want to reduce $b_{i,t}$ in the future, while $U_{i,s}$ implies that patients want to reduce the adverse factors of future enrollment, scaled by their perception of the importance of the adverse factors. Thus, patients choose whether to enroll (or remain enrolled) by balancing these two parts of utility $U_i$. 

Then, the patient's enrollment decision at time $t$ is given by:
\begin{equation}
    z_{i,t} \in (z_{i,t-1} + y_{i,t} - z_{i,t-1}y_{i,t}) \cdot \argmax_{z \in \{0,1\}} \ \mathbb{E}_\xi \ U_i(b_{i,t+1},s_{i,t+1},\theta_{i,t},\xi_{i,t}, z).
    \label{eq:patient_prob}
\end{equation}

Intuitively, this approach implies that a patient's binary enrollment decision, $z_{i,t}$, maximizes their overall  expected utility, where $U_{i}$ is a function of the utility associated with their future health states ($b_{i,t+1}$), adverse factors ($s_{i,t+1}$), and perception of adverse factors ($\theta_{i,t}$), where the expectation is taken over the distribution of $\xi_{i,t}$. We include the boolean ``OR'' operator $(z_{i,t-1} + y_{i,t} - z_{i,t-1}y_{i,t})$ to ensure that a patient's prior enrollment status and their present enrollment decision are consistent. A patient can only choose to be enrolled in the current period if they were enrolled in the previous period \emph{or} if they were not enrolled in the previous period but received a screening visit.

To characterize patients' decisions we introduce the variable $B_{i,t} := U_i(b_{i,t+1}, s_{i,t+1}, \theta_{i,t}, \xi_{i,t},  1) - U_i(b_{i,t+1}, s_{i,t+1}, \theta_{i,t}, \xi_{i,t}, 0)$
representing the net benefit of enrollment to patient $i$. This leads to the following characterization of the patients' enrollment decisions:
\begin{proposition}
\label{prop:benefit}
Given the patient utility model presented by equations \eqref{eq:fbg_util},\eqref{eq:s_util},\eqref{eq:patient_prob}, and the patient state dynamic model described in Section \ref{sec:patient_model}, at time $t$, patient enrollment decisions are given by:
\begin{equation}
    z_{i,t} = (z_{i,t-1} + y_{i,t} - z_{i,t-1}y_{i,t}) \cdot \mathbbm{1}[B_{i,t} \geq 0], \label{eq:enroll_condition}
\end{equation}
where $\mathbbm{1}:\mathbb{R} \mapsto \{0,1\}$ is an indicator function. Moreover, the benefit $B_{i,t}$ has the following closed form:
\begin{equation}
    B_{i,t} = \mu_i - \theta_{i,t}(\gamma(s_{i,t} - s_{i,0})+s_{i,0}) + (\alpha_i - \theta_{i,t} \beta_i)y_{i,t}.
\end{equation}

\end{proposition}

The complete proof of this proposition is found in Appendix~\ref{app:proofs}, but we provide a brief sketch here. We can interpret the benefit of enrollment as the sum of the fixed effect of enrollment ($\mu_i - \theta_{i,t} (\gamma( s_{i,t} -  s_{i,0}) + s_{i,0})$) and the variable/visit effect ($\alpha_i - \theta_{i,t} \beta_i$). The benefit function is obtained by adding the net utility of enrollment with respect to the FBG state, $b_{i,t+1}$, and adverse factors, $s_{i,t+1}$. The equations for $b_{i,t+1}$ and $s_{t+1}$ are then used to obtain a benefit function in terms of the present states, $b_{i,t}$ and $s_{i,t}$. Note that patient enrollment requires not only the utility of enrollment to be equal to or exceed the utility of non-enrollment ($B_{i,t} \geq 0$), but also enrollment in the previous period ($z_{i,t-1}=1$) or a screening visit in the present period ($y_{i,t}=1$ if $z_{i,t-1}=0$). Also note that, although $B_{i,t}$ is the difference of two random functions, because they are both functions of the same random variable (RV) in an affine term, $B_{i,t}$ is a deterministic quantity. 

\begin{remark}
According to the assumptions of the model, when $B_{i,t} = 0$ the patient will be indifferent between enrolling and not enrolling in the intervention. Without loss of generality, we assume that the tie is broken in favor of enrolling in the intervention. 
\end{remark}

The characterization of patient enrollment decisions in Proposition~\ref{prop:benefit} will later allow us to define when visiting a patient is beneficial for their enrollment and how patient enrollment relates to the optimal visit policy from the provider's perspective.

\subsection{Parameter Estimation}
\label{sec:parameter-estimation}
In this section, we present a procedure that enables the provider to predict patient enrollment decisions for a proposed visit plan. Our estimation procedure offers two key advantages over standard, out-of-the-box ML approaches. First, it enables seamless integration with the prescriptive optimization framework for CHW visit planning, ensuring that estimated patient parameters directly inform visit planning decisions. Second, it maintains consistency with patient trajectories and underlying dynamics. This latter feature is particularly important for leveraging the longitudinal visit data, and for capturing how patient states evolve over time. By aligning estimation with the trajectory-based state update equations, the approach preserves the causal and temporal structure of the problem, thereby supporting both accurate prediction and effective prescription. At each time period, the provider observes only the enrollment status of the patient ($z_{i,t}$), a record of whether or not the patient was visited ($y_{i,t}$), and a noisy signal of the patient's FBG -- including measurement noise and biological variability in the FBG itself \citep{bonora2011pros}. Let $\Bar{b}_{i,t}$ be the noisy observation of the log FBG for patient $i$ in period $t$. We assume that the relationship between  $\Bar{b}_{i,t}$ and $b_{i,t}$ is given by: $\Bar{b}_{i,t} = b_{i,t} + \epsilon_{i,t}$, where $\epsilon_{i,t}$ are i.i.d. random variables representing the disturbance in the observations such that $\mathbb{E}[\epsilon_{i,t}] = 0$ and $\mathbb{E}[\epsilon_{i,t}^2] < \infty$, with density function $f_\epsilon: \mathbb{R} \mapsto \mathbb{R}_+$. Note that this is a standard modeling assumption that can be satisfied by various families of distributions including Normal and Laplace distributions \citep{billingsley1961lindeberg}. Since each patient's parameters are estimated individually using their longitudinal visit history, we drop the patient index $i$ for the remainder of this section. 

Let $\mathcal{K} \subset \mathcal{T}$ be the set of time periods for which there is an observation of the log-FBG for a particular patient (e.g., when a visit occurred). Then, we can use a maximum likelihood estimation (MLE) approach to jointly estimate all unknown model parameters, which include FBG progression ($p$), visit effect on FBG ($\alpha$), enrollment effect on FBG ($\mu$), visit effect on adverse factors ($\beta$), visit effect on perception of adverse factors ($\lambda$), and two discount factors ($\gamma$ and $\rho$). This approach also allows us to obtain FBG estimates for missing periods $t' \in \mathcal{T} \setminus \mathcal{K}$ through the use of our state dynamics equations \eqref{eq:fbg_dynamics}, \eqref{eq:disutil_dynamics}, \eqref{eq:percep_dynamics}, and \eqref{eq:enroll_condition}, an important feature for longitudinal datasets that commonly have missing data (e.g., due to a patient not being home or not being in a fasting state when a CHW visits). We show that the MLE problem can be formulated as a mixed integer linear program:
\begin{proposition}
\label{prop:mle_form}
The MLE problem \eqref{eq:mle_packed} can be formulated as the following constrained optimization problem:
\begin{subequations}
\label{eq:mle_prob}
\begin{align}
    \underset{\bb,\bs,\btheta,\beta, \alpha, \mu, p, \lambda, \rho, \gamma, \bxi}{\mathrm{minimize}} \: & \sum_{t \in \mathcal{K}} \log f_\epsilon(\bar{b}_t - b_t) + \sum_{t\in\mathcal{T}} \log f_\xi( \xi_t) \\
    \mathrm{subject\,to}\quad &  b_{t+1} = b_{t} + p - \mu z_{t} -\alpha y_{t} z_{t} + \xi_t, \quad  \forall  t \in \mathcal{T}, \label{eq:mle_b_const_1} \\
    & s_{t+1} = \gamma z_{t} (s_t - s_0) + z_t s_0 + \beta y_{t} z_{t}, \quad \forall t \in \mathcal{T}, \label{eq:mle_s_const}\\
    & \theta_{t+1} = \rho (\theta_{t} - \theta_{0}) + \theta_{0} - \lambda y_{t} z_{t}, \quad \forall t \in \mathcal{T}, \label{eq:mle_theta_const}\\
    & B_{t} \geq -M(1 - z_{t}), \quad \forall t \in \mathcal{T}, \label{c_bta}\\
    & B_{t} \leq M z_{t}, \quad \forall t \in \mathcal{T}, \label{eq:mle_b_const_2} \\
    & b_{t} \in \mathcal{B},s_{t} \in\mathcal{S},\theta_t \in \Theta, \quad \forall t \in \mathcal{T}, \label{c_last} \\
    & \beta \in \mathcal{S}, \alpha, \mu, p \in \mathcal{B}, \lambda \in \Theta, \rho, \gamma \in (0,1),
\end{align}
\end{subequations}
\end{proposition}
where $B_{t} = \mu - \theta_{t}(\gamma(s_{t}-s_{0}) + s_{0}) + (\alpha - \theta_{t} \beta)y_{t}$ and $f_\xi:\mathbb{R} \mapsto \mathbb{R}_+$ is the density function for $\xi$ (RV). 

The proof begins decomposing the likelihood function presented in \ref{eq:mle_packed} and subsequently applying the log to both sides, allowing the log of the products to be separated as a sum of logarithms. The term including $\bar{b}_t$ is then rewritten using the definition of the disturbance model and the remaining terms correspond to deterministic system dynamics.

The objective function may be nonlinear depending on the choice of disturbance distribution $\epsilon_{i,t}$, which we established assumptions for in Section 4.2 (i.i.d., $\mathbb{E}[\epsilon_{i,t}]=0$, and $\mathbb{E}[\epsilon_{i,t}^2]<\infty$). See Remark~\ref{remark-convexity} for choices of disturbance distribution that lead to convex objective functions. Furthermore, the products of continuous decision variables in Constraints \eqref{eq:mle_s_const}, \eqref{eq:mle_theta_const}, \eqref{c_bta} and \eqref{eq:mle_b_const_2} are nonlinear and cannot be solved directly by commercial solvers.  We used a coarse grid search to handle the products of continuous variables in the parameter estimation and found that it gave strong computational results. See~\ref{EC_patparamest} for details on our coarse grid search procedure. 

\begin{remark}
\label{remark-convexity}
    Problem~\eqref{eq:mle_prob} can be expressed as a linear or convex optimization problem for fixed $\gamma$, $\rho$, $s_0$, and $\beta$ under the assumption that $\epsilon_{i,t}$ is Laplace or Normally distributed. In both cases, standard convex optimization techniques are guaranteed to yield a global optimum.
\end{remark}


While, in general, grid search may result in multiple optima, using MLE results in a solution on the grid that minimizes the KL divergence between the grid parameters and the true distribution \citep{bickel2015mathematical}. Given this restriction, we can still show that the parameter estimates given by our procedure will approach this value as we obtain more data. Let $(\bb^*,\bs^*,\btheta^*,\beta^*,\alpha^*,\mu^*,p^*,\lambda^*)$ be the closest set of parameters to the true system parameters subject to $s_0,\beta$ being confined to the gird of the grid search. Further let $(\hat{\bb},\hat{\bs},\hat{\btheta},\hat{\beta},\hat{\alpha},\hat{\mu},\hat{p},\hat{\lambda})$ be the minimizers of  \eqref{eq:mle_prob}. Then, under mild conditions, we obtain the following result:
\begin{proposition}
\label{prop:mle_consist}
    As $|\mathcal{K}|,|\mathcal{T}| \rightarrow \infty$,  $(\hat{\bb},\hat{\bs},\hat{\btheta},\hat{\beta},\hat{\alpha},\hat{\mu},\hat{p},\hat{\lambda}) \xrightarrow{p} (\bb^*,\bs^*,\btheta^*,\beta^*,\alpha^*,\mu^*,p^*,\lambda^*)$,
\end{proposition}
where $\xrightarrow{p}$ denotes convergence in probability \citep{bickel2015mathematical}. This result states that our estimation procedure is statistically consistent, that is, as we collect additional participant observations, our estimates get as close as possible to the true parameter values. We present an in depth discussion and proof of this result in Appendix~\ref{EC:prop:mle_consist}.
An alternative method for solving Problem \eqref{eq:mle_prob} would be to implement a Reformulation-Linearization Technique \citep{sherali1990hierarchy, bestuzheva2024efficient} along with McCormick envelopes \citep{mccormick1976computability} to obtain a convex relaxation of the bilinear terms. While this may simplify the computation and avoid the need for grid search, using such a relaxation would not allow us to obtain the result of Proposition~\ref{prop:mle_consist}. This is because a key aspect of the proof of this proposition is obtaining a global optimal solution to \eqref{eq:mle_prob} within fixed numerical precision. Thus, if the relaxation gap is too large, this would remove the consistency guarantee. 

We evaluate the out-of-sample performance of \eqref{eq:mle_prob} for predicting the probability of \emph{in control} FBG. We compare our approach with standard machine learning models and find that our method has similar performance, but offers two clear advantages: it can be readily integrated within the optimization framework and we have theoretical guarantees for the consistency of the estimations. See Appendix~\ref{EC:incontrolprediction} for more details.


\section{CHW Provider Problem}
\label{sec:provider_model}

In this section, we describe the dynamic programming framework used to optimize the decisions of the CHW provider (e.g., NanoHealth), while accounting for individual patient enrollment decisions. The goal of the provider is to maximize glycemic control (i.e., the number of individuals whose FBG is less than a given threshold) for the entire community under consideration. Throughout this section, we assume that the provider has full knowledge of the patient parameters described in Section~\ref{sec:patient_model} for all patients in the targeted community, even for those who have not been screened yet. In practice, pilot trials allow planners to obtain data that can be used as initial estimates and forecast disease progression for unscreened patients \citep[e.g.,][]{deo2018evaluation}. We demonstrate the feasibility of such an approach using a two-phase machine learning framework in Appendix~\ref{EC:clusteringprediction}.

In accordance with prevailing clinical guidelines, we define glycemic control for each patient through the use of a threshold due to a nonlinear association between FBG and the risk of complications (e.g., coronary heart disease and ischaemic stroke), with substantially higher risks for FBG levels above 125 mg/dL \citep{world2016global, emerging2010diabetes}.

\subsection{Single Patient Problem}
\label{sec:prov_prob_single_patient}

To build intuition, we first consider a single-patient problem (Section~\ref{sec:prov_prob_multi_patient} considers the multi-patient problem) and for brevity, we drop the patient index $i$. Using the dynamics described in Section~\ref{sec:patient_model}, the state of a patient in period $t\in \mathcal{T}$ is given by $x_t := (b_t,s_t,\theta_t,z_{t-1})$ and the provider's action is given by $y_t$, which represents the decision of whether or not to visit the patient in period $t$. 
Let $\delta$ denote the maximum tolerable FBG. Then, the total reward is given by: 
$g_t(x_t) = g_t (b_t, s_t, \theta_t, z_{t-1}) = \mathbbm{1}(b_t \leq \delta), \forall t\in \mathcal{T}.$ The total reward for a planning horizon with $N$ periods is then $\sum_{t=0}^N \mathbbm{1}(b_t \leq \delta)$. However, since $b_t$ is a random variable (because of dependence on $\xi$), the provider cannot directly maximize this value and must instead maximize the expected total number of periods where the patient's FBG is in control, which is given by $\sum_{t=0}^N \mathbb{E}_\xi[\mathbbm{1}(b_t \leq \delta)] = \sum_{t=0}^N \mathbb{P}_\xi(b_t \leq \delta)$.

Using this reward function and the patient dynamics, the optimization model to determine the optimal timing of CHW visits can be written as: 
\begin{subequations}
\label{prob:single-patient}
\begin{align}
    \underset{\bz, \by, \bb,\bs,\btheta }{\mathrm{maximize}} & \quad \sum_{t\in \mathcal{T}} \mathbb{P}_\xi(b_{t} \leq \delta)    \\ 
    \mathrm{subject\,to} \quad & \eqref{eq:mle_b_const_1} - \eqref{c_last}, \\ 
    & z_{t} = (z_{t-1}+y_{t}-z_{t-1}y_{t}) \cdot \mathbbm{1}(B_{t}(y_{t}) \geq 0), \quad \forall t \in \mathcal{T},\\ \label{SPP_4}
    & y_{t},z_{t} \in \{0,1\}, \quad \forall  t \in \mathcal{T}, 
\end{align}
\end{subequations}
where $B_{t} = \mu - \theta_{t}(\gamma(s_{t}-s_{0}) + s_{0}) + (\alpha - \theta_{t} \beta)y_{t}$. The objective function maximizes the expected total number of periods where the patient's FBG is \emph{in control} (i.e., less than or equal to $\delta$). 

Due to the stochasticity in the FBG update equation and the presence of bilinear terms between two continuous variables ($\theta_t$ and $s_t$) in constraints ~\eqref{c_bta} and \eqref{eq:mle_b_const_2}, we leverage DP techniques to obtain an optimal policy for this formulation. Let $V_t(b_t,s_t,\theta_t,z_{t-1})$ for $t\in \mathcal{T}$ represent the optimal value-to-go function, i.e., the optimal expected number of periods the patient will be in control from period $t$ to period $N$. The DP equations can be written as: 
\begin{flalign}
    V_t(b_t,s_t,\theta_t,z_{t-1})=&\mathbbm{1}(b_t \leq \delta) + \label{dp-single-patient-last-period}&&\\\nonumber & \max_{y_t\in\{0,1\}}\bigg\{\mathbb{E}_\xi\big[ \, V_{t+1}\big( b_t+p-\mu z_t(y_t) -\alpha y_t z_{t}(y_t) + \xi_t,  z_t(y_t)(\gamma(s_t-s_0)+s_0+\beta y_t),  &&\\\nonumber &
      \rho(\theta_t-\theta_0)+\theta_0-\lambda y_t z_{t}(y_t),  
     z_t(y_t)\big)\big]\bigg\}, \forall t\in \mathcal{T}=\{0,...,N-1\},  &&\\\nonumber
    V_N(b_N,s_N,\theta_N,&z_{N-1}) = \mathbbm{1}(b_N \leq \delta).  
\end{flalign}

The value function in a general period $t$ is determined by the immediate reward for having the FBG in control in period $t$ and the expected value function in period $t+1$, where each state is updated according to the system dynamics equations explained in Section~\ref{sec:patient_model}. In the terminal condition, the provider accrues one final reward if the FBG is in control. Note that, in contrast to the optimization formulation in \eqref{prob:single-patient}, where $b_t$, $s_t$, $\theta_t$, $z_{t}$, and $y_t$ are decision variables, the only decision/control variable in the DP equations is $y_t$. While $z_t$ is in reality a patient decision, it is treated as a state derived from state $z_{t-1}$, other states in period $t$ via the benefit function ($B_t$), and provider visit decision $y_t$. This modeling choice allows us to analyze the effect of provider decisions on patient enrollment, thereby leading to more efficient behavioral health interventions.

\subsubsection{Structural Results.} \label{single_struc}

In this section, we characterize the relationship between $V_t(b_t,s_t,\theta_t,z_{t-1})$ and $z_{t-1}$ to guide the provider on how to employ visit strategies that improve glycemic control.
Before we present the theorem and corresponding proof of this result, we first establish some lemmas pertaining the relationship between $z_t$ and $z_{t-1}$, and $V_N$ and $z_{N-1}$.

\begin{lemma}
\label{lemma-1}
$z_{t}(z_{t-1}, y_{t}) = (z_{t-1}+y_{t}-z_{t-1}y_{t}) \cdot \mathbbm{1}(B_{t}(y_{t}) \geq 0)$ is nondecreasing in $z_{t-1}, \forall t\in \mathcal{T}$.
\end{lemma}

The result is obtained by evaluating the binary variable $z_{t}$ for the two possible enrollment states $z_{t-1}$. Lemma~\ref{lemma-1} demonstrates that, if a patient is enrolled in the intervention in period $t-1$, it is more likely 
they will stay enrolled in future periods irrespective of the CHW visit. Next, we consider the structure of the terminal value function.
\begin{lemma}
\label{lemma-2}
    With probability one, $V_N(b_N,s_N,\theta_N,z_{N-1}) = \mathbbm{1}(b_N \leq \delta)$ is nondecreasing in $z_{N-1}$.
\end{lemma}

The proof consists of evaluating the threshold function $V_N$ for different values of $z_{N-1}$ and relies on the fact that, in period $N$, the value function only depends on the FBG state. Lemma~\ref{lemma-2} demonstrates that, if a patient is enrolled in the penultimate time period, they are more likely to have their FBG in control than if they are not. With these lemmas, we can establish our primary structural result with the following theorem, which has its proof completed by induction. The base case follows from the fact that $V_N(z_{N-2})$, the evaluation of $V_N$ for a given disturbance sequence at a particular state in time $N-2$, is a composition of nondecreasing functions $V_N(b_N,s_N,\theta_N,z_{N-1})$ and $z_{N-1}(z_{N-2},y_{N-1})$), and therefore also nondecreasing. The induction step follows a similar process, in addition to evaluating the maximum of monotonically nondecreasing functions. 

\begin{theorem}
\label{thm-1}
    Given the dynamic programming equations in \eqref{dp-single-patient-last-period}, $V_t(b_t,s_t,\theta_t,z_{t-1})$ is nondecreasing in $z_{t-1}, \forall t\in \mathcal{T}$. Since $z_{t-1}\in\{0,1\}$, this is equivalent to $V_t(b_t,s_t,\theta_t,0) \leq V_t(b_t,s_t,\theta_t,1), \forall t\in \mathcal{T}$ with probability one.
\end{theorem}

The result above is non-intuitive because it implies that the provider should focus on keeping patients enrolled to achieve better glycemic control, even if that means that some patients should receive less visits to remain enrolled -- a situation that may occur when a patient has a large increase in social stigma/cognitive burden for each visit.

\subsubsection{Optimal Visit Policy.}
\label{sec-3.2}

We now prove structural results for the optimal CHW provider policy. Recall that patient enrollment is given by $z_{t} = (z_{t-1} + y_{t} - z_{t-1}y_{t}) \cdot \mathbbm{1}[B_{t}(y_{t}) \geq 0]$, (where we view benefit as an explicit function of the visit), and that Theorem~\ref{thm-1} implies that a patient will never be worse off if they are enrolled in the intervention.

\begin{theorem}\label{thm:suff_nec_cond}
Given the patient model in Section \ref{sec:patient_model}, a policy that solves the dynamic programming equations in \eqref{dp-single-patient-last-period} must satisfy the following necessary and sufficient conditions:
\begin{enumerate}[(i)]
    \item Visit the patient (i.e., $y_t = 1$) when 
        $z_{t-1} = 1$, $B_{t}(0) < 0$, and $B_{t}(1) \geq 0$ or when
        $z_{t-1} = 0$ and $B_{t}(1) \geq 0$.
        
    \item Never visit the patient (i.e., $y_t =0$) when 
        $z_{t-1} = 1$, $B_{t}(0) \geq 0$, and $B_{t}(1) < 0$.
    
    \item Indifferent between visiting and not visiting when 
        $z_{t-1} = 1$, $B_{t}(0) <0$, and $B_{t}(1) < 0$ or when $B_{t}(0) \geq 0$ and $B_{t}(1)\geq 0$ or when $z_{t-1} = 0$ and $B_{t}(1) < 0$.
\end{enumerate}
\end{theorem}


The proof of the theorem proceeds by showing that the value function is strictly increasing in $y_{t-1}$ for case $(i)$, strictly decreasing for case $(ii)$, and constant  for case $(iii)$. Using the result from Theorem~\ref{thm-1}, which relates the value function ($V_t$) to the enrollment status in the previous period ($z_{t-1}$), the only remaining task is to evaluate the relationship between $z_t$ and $y_t$ (which would imply a similar relationship between $z_{t-1}$ and $y_{t-1}$). Theorem~\ref{thm:suff_nec_cond} establishes the conditions for an optimal policy. Specifically, it determines that a patient should only be visited if the visit will lead to an enrollment or prevent a patient from dropping out of treatment; that a patient should never be visited if the visit will lead them to drop out of treatment; and that a provider is indifferent if the visit cannot prevent a patient from dropping out, if a patient will remain enrolled regardless of whether a visit is received, or if an unenrolled patient will not enroll in treatment if visited.

Note that because patient decisions are outcomes of thresholding functions, and similarly, the CHW provider value function is an affine combination of thresholding functions, there may be multiple optimal policies that solve the DP equations. For example, if all patients' FBG levels are always in control or out of control for any number of CHW visits received, it would be optimal to visit everyone, visit no one, or any other solution in-between. To choose among these optimal policies, we look ahead to the multi-patient setting where CHW resources are likely to be constrained. Hence, an advantageous characteristic of an optimal policy is to visit a patient as little as possible while ensuring they remain enrolled in the program and maintain their FBG under control. Thus, we show that the following policy is optimal:

\begin{corollary}
\label{cor-1}
An optimal policy that solves the CHW provider problem described in Section~\ref{sec:provider_model} is given by:
\begin{equation*}
    y_t(x_t) = \begin{cases}
        1, \:\: \text{if } B_t(y_t=1) \geq 0\text{ and } \Big( B_t(y_t=0) < 0 \\ 
        \hspace*{9em} \text{ or } B_t(y_t=0) \geq 0 \text{ and } \big(z_{t-1} = 0 \\ 
        \hspace*{16.7em} \text{ or } z_{t-1} = 1 \text{ and }  B_t(y_t=1) - B_t(y_t=0) > 0 \big) \Big),  \\
        0, \:\: otherwise.
    \end{cases}
\end{equation*}
\end{corollary}

Clearly, the policy outlined in Corollary~\ref{cor-1} meets the conditions in Theorem~\ref{thm:suff_nec_cond}. Intuitively, this policy only assigns a CHW to visit a patient if the visit is necessary for enrollment or to avoid drop out, or if the visit provides a strict improvement in the patient's utility -- visiting a patient in this case can be interpreted as avoiding a future screening visit by providing a proactive maintenance visit. Driven by the condition $B_t(y_t = 1) - B_t(y_t = 0) > 0$ in Corollary~1, this proactive visit might prevent drop outs, and therefore, the need to re-screen patients, which could waste visiting resources. Figure~\ref{fig:visit-decision-tree} provides a graphical representation of the policy structure. 
\begin{figure}[!ht]
    \centering
    \scalebox{.65}{
    \begin{tikzpicture}[node distance={40mm}, thick, main/.style = {draw, circle, inner sep=1.25pt, fill=black}, 
    dbox/.style = {diamond, draw, minimum width = 3.6cm, minimum height = 3.2cm, align=center, execute at begin node=\setlength{\baselineskip}{3ex}}, 
    rect/.style = {rectangle, draw, minimum width = 2.3cm, minimum height = 0.8cm}] 
        \node [dbox] (split1)  {$B_t(y_t=1)$};
        \node [rect] (novisit1) [below left of=split1] {Don't visit};
        \node [dbox] (split2) [below right of=split1] {$B_t(y_t=0)$};
        \node [rect] (novisit2) [below left of=split2] {Visit};
        \node [dbox] (split3) [below right of=split2] {$z_{t-1}$};
        \node [rect] (novisit3) [below left of=split3] {Visit};
        \node [dbox] (split4) [below right of=split3] {$B_t(y_t=1)-$ \\ $B_t(y_t=0)$};
        \node [rect] (novisit4) [below left of=split4] {Visit};
        \node [rect] (split5) [below right of=split4] {Don't visit};
        \draw[-, black] (split1)--(split2) node[pos=0.75,above=1em] {$\geq 0$};
        \draw[-, black] (split1)--(novisit1) node[pos=0.75,above=1em] {$< 0$};
        \draw[-, black] (split2)--(split3) node[pos=0.75,above=1em] {$\geq 0$};
        \draw[-, black] (split2)--(novisit2) node[pos=0.75,above=1em] {$< 0$};
        \draw[-, black] (split3)--(split4) node[pos=0.75,above=1em] {$=1$};
        \draw[-, black] (split3)--(novisit3) node[pos=0.75,above=1em] {$=0$};
        \draw[-, black] (split4)--(split5) node[pos=0.75,above=1em] {$\leq 0$};
        \draw[-, black] (split4)--(novisit4) node[pos=0.75,above=1em] {$> 0$};
    \end{tikzpicture}}
    \caption{Graphical representation of the structure of the optimal policy given in Corollary~\ref{cor-1}.}
    \label{fig:visit-decision-tree}
\end{figure}
\subsection{Multi-Patient Problem}
\label{sec:prov_prob_multi_patient}

In this section, we extend our results to the multi-patient problem faced by CHW organizations like NanoHealth. CHW organizations are typically resource-limited, especially those operating in global health contexts such as India. As a result, we limit the maximum number of patients that can be visited in each period; in reality, this amount can be derived from the number of CHWs and their total daily working hours. Let $C$ represent the number of available CHW visits per period. Recall that $\mathcal{P}$ represents the set of patients and note that we now add back patient indices to our variables. The multi-patient provider problem can be written as:
\begin{subequations}
\begin{align}
    \underset{\bz, \by, \bb,\bs,\btheta }{\mathrm{maximize}} & \quad \sum_{i\in \mathcal{P}}\sum_{t\in \mathcal{T}} \mathbb{P}_\xi(b_{i,t} \leq \delta)    \\
    \mathrm{subject\,to} \quad &  \sum_{i\in \mathcal{P}} y_{i,t} \leq C, \quad \forall t \in \mathcal{T}, \label{MPP_1} \\ 
    & b_{i,t+1} = b_{i,t} + p_i - \mu_i z_{i,t} -\alpha_i y_{i,t} z_{i,t} + \xi_{i,t}, \quad  \forall i \in \mathcal{P}, t \in \mathcal{T},  \\
    & s_{i,t+1} = \gamma z_{i,t} (s_{i,t} - s_{i,0}) + z_{i,t} s_{i,0} + \beta_i y_{i,t} z_{i,t}, \quad \forall i \in \mathcal{P}, t \in \mathcal{T}, \\
    & \theta_{i,t+1} = \rho (\theta_{i,t} - \theta_{i,0}) + \theta_{i,0} - \lambda_i y_{i,t} z_{i,t}, \quad \forall i \in \mathcal{P},t \in \mathcal{T},\\
    & z_{i,t} = (z_{i,t-1}+y_{i,t}-z_{i,t-1}y_{i,t}) \cdot \mathbbm{1}(B_{i,t}(y_{i,t}) \geq 0), \quad \forall i \in \mathcal{P},t \in \mathcal{T},\\
    & B_{i,t} \geq -M(1 - z_{i,t}), \quad \forall i \in \mathcal{P},t \in \mathcal{T}, \\
    & B_{i,t} \leq M z_{i,t}, \quad \forall i \in \mathcal{P},t \in \mathcal{T},  \\
    & y_{i,t},z_{i,t} \in \{0,1\}, \quad \forall  i \in \mathcal{P},t \in \mathcal{T},  \\
    & b_{i,t},s_{i,t},\theta_{i,t} \geq 0, \quad \forall  i \in \mathcal{P}, t \in \mathcal{T},
\end{align}
\label{eq:multi_patient_problem}
\end{subequations}
where $B_{i,t} = \mu_i - \theta_{i,t}(\gamma(s_{i,t}-s_{i,0}) + s_{i,0}) + (\alpha_i - \theta_{i,t} \beta_i)y_{i,t}.$ The objective function maximizes the expected total number of patient periods where FBG is in control. Constraint~\eqref{MPP_1} limits the number of visits in each period to $C$. The remaining constraints are the same as in the single patient problem, except that there is a set of constraints for each patient. Aside from constraint~\eqref{MPP_1} the problem is completely separable by patient. We exclude travel times as, in our context of densely populated urban areas, these are negligible.

Similar to the single patient problem, we leverage DP to solve the model.
Let $\mathcal{V}_t(\bb_t,\bs_t,\btheta_t,\bz_{t-1})$ represent the optimal value-to-go function, where all boldface variables are vectors of length $|\mathcal{P}|$ with each entry corresponding to the state values for a given patient at time $t$ and $\odot$ represents component-wise multiplication. Furthermore, the disturbance vector $\bxi_t$ has independent identically distributed components with the same distribution and assumptions as $\bxi_t$ in the previous sections. For notational simplicity, let $h: \mathbb{R}_+^{|\mathcal{P}| \times 3}\times \{0,1\}^{ |\mathcal{P}| \times 2} \mapsto \mathbb{R}_+^{|\mathcal{P}|\times 3}\times \{0,1\}^{|\mathcal{P}|}$ be defined as the vector-valued system dynamics function: \begin{equation}
    \begin{pmatrix} \bb_{t+1} \\ \bs_{t+1} \\ \btheta_{t+1} \\ \bz_{t} \end{pmatrix} = h(\bb_t,\bs_t,\btheta_t,\bz_{t-1},\by_t)= \begin{pmatrix} \bb_t+\bp-\bmu \odot\bz_t(\by_t) -\balpha \odot\by_t \odot\bz_{t}(\by_t) + \bxi_t \\ \bz_t(\by_t)\odot(\bgamma\odot(\bs_t-\bs_0)+\bs_0+\bbeta\odot \by_t)\\ \brho(\btheta_t-\btheta_0)+\btheta_0-\blambda \odot\by_t \odot\bz_{t}(\by_t) \\ (\bz_{t-1} + \by_t - \bz_{t-1}\by_t) \odot \mathbbm{1}_{\mathbf{B}_t(\by_t) \geq 0 } \end{pmatrix},
\end{equation}
where the vector $\mathbbm{1}_{\mathbf{B}_t(\by_t) \geq 0 } \in \{0,1\}^{|\mathcal{P}|}$ is a component-wise indicator such that its $i^\text{th}$ entry is defined as $\mathbbm{1}_{\mathbf{B}_t(\by_t) \geq 0, i } = \mathbbm{1}\big(B_{i,t}(y_{i,t}) \geq 0\big)$, and $\bz_t(\by_t)$ indicates the last entry in the composite vector on the right-hand side. Note that $\bz_t$ is not a decision made by the provider (and by extension the optimization model), but rather a dependent state that is derived from other states (i.e., $\bz_{t-1}$, $\bs_{t}$, $\btheta_{t}$) and the visit decision (control) $\by_t$. In other words, in our bi-level framework, $\by_t$ and $\bz_t$ interact in a leader-follower dynamic, with the provider anticipating but not dictating the patient’s response. We will use the convention that $h_i:\mathbb{R}_+^{3}\times \{0,1\}^2 \mapsto \mathbb{R}_+^{3}\times \{0,1\}$ is the component of $h$ that corresponds to the dynamics of patient $i$. Using this notation, the DP equations are:
\begin{equation}
\begin{aligned}
    &\mathcal{V}_t(\bb_t,\bs_t,\btheta_t,\bz_{t-1}) = \sum_{i \in \mathcal{P}} \mathbbm{1}(b_{i,t} \leq \delta) + \max_{\by_t \in \{0,1\}^{|\mathcal{P}|}, \by_t^\top \mathbbm{1} \leq C} \mathbb{E}_{\bxi} \mathcal{V}_{t+1}\big(h(\bb_t,\bs_t,\btheta_t,\bz_{t-1},\by_t)\big), \quad \forall t \in \mathcal{T}, \\
    &\mathcal{V}_N(\bb_N,\bs_N,\btheta_N,\bz_{N-1}) = \sum_{i \in \mathcal{P}} \mathbbm{1}(b_{i,N} \leq \delta), \label{eq:joint_dp_N}
\end{aligned}  
\end{equation}
where $\mathcal{T} = \{0,...,N-1\}$. We can obtain an initial structural result for the optimal visit policy of the multi-patient problem using the optimal policy from the single-patient problem to define a set that represents the patients that would be visited in each period if there was no capacity constraint. To build intuition for why this would be of interest, consider the case when the multi-patient problem is uncapacitated, that is, $C \geq |\mathcal{P}|$. In this case, the problem is completely separable by patient and can thus be decomposed into $|\mathcal{P}|$ single-patient problems that can be solved using the results in Section~\ref{sec:prov_prob_single_patient}. Thus, in this case, we would only visit patients if we would visit them in the single-patient setting. We call this group of patients the set of patients of interest.

\begin{definition}
Let $\mathcal{I}_t$, the set of patients of interest in period $t\in \mathcal{T}$, be defined as:
\begin{align*}
\label{def-patientsofinterest}
    \mathcal{I}_t := \Bigg\{i \in \mathcal{P} \;\Big|& \big(B_{i,t}(y_{i,t}=1) \geq 0 \big) \text{ and } \\ & \big((B_{i,t}(y_{i,t}=0) < 0) \:  \text{ or } \: (z_{i,t-1}=0) \text{ or } (z_{i,t-1}=1 \text{ and } B_t(y_t=1)-B_t(y_t=0) > 0)\big) \Bigg\}.
\end{align*}
\end{definition}
Another property that will be key for our analysis of the multi-patient problem is additive decomposability. This property is key in the analysis of weakly-coupled MDPs and restless multiarmed bandit problems \citep{adelman2008relaxations, brown2023strength, topaloglu2009using} and imply that the value function of the entirety of the system (i.e., the entire cohort of patients) can be written as a sum of component functions that are only a function of each individual subsystem state (i.e., each individual patient state). Formally we write this as Definition~\ref{def:add_decomp}.

\begin{definition}
\label{def:add_decomp}
    Consider a weakly-coupled dynamical system with $n$ individual subsystems. Let $x_{i,t} \in \mathcal{X}$ be the state of system $i \in \{1,...,n\}$ at some point in time $t$, and let $\bx_t = (x_{1,t},x_{2,t},...,x_{n,t}) \in \mathcal{X}^n$ be the complete system state at time $t$. Then we say the value function of the system at time $t$ given by $\mathcal{V}_t:\mathcal{X}^n \mapsto \mathbb{R}$ is additively decomposable if and only if there exist individual functions $\mathcal{V}_{i,t}: \mathcal{X} \mapsto\mathbb{R}$ such that $\mathcal{V}_{t}(\bx_t) = \sum_{i=1}^n \mathcal{V}_{i,t}(x_{i,t})$ for all states $\bx_t \in \mathcal{X}^n$.
\end{definition}

Additive decomposability is a desirable property since it would simplify the computation for the decision maker to extract a system optimal policy from the individual optimal policies of the subsystems. Thus, if the multi-patient problem value functions in  \eqref{eq:joint_dp_N} had this property, we could apply the results of the previous section to compute a multi-patient policy. While it is trivial to see that the value function for period $N$ satisfies this property, we will show that, for $t < N$, this property does not generally hold. Consider the single-stage problem of finding an optimal visit schedule. For notational brevity, let $V_{i,N}^j = \mathbb{E}_{\bxi}\big[ \mathcal{V}_{i,N}(b_{i,N},s_{i,t},\theta_{i,N},z_{i,N-1}(y_{i,N-1} = j))\big]$ and $\mathcal{I}_N^j = \{i \in \mathcal{I}_N | y_{i,N}=j\}$ where $j \in \{0,1\}$. Then our result can be stated as follows:
\begin{lemma}
\label{lemma-6}
    For period $N-1$, the value function $\mathcal{V}_{N-1}(\bb_{N-1},\bs_{N-1},\btheta_{N-1},\bz_{N-1})$ is such that there exist $\mathcal{V}_{i,N-1}(\bb_{N-1},\bs_{N-1},\btheta_{N-1},\bz_{N-1})$ where $\mathcal{V}_{N-1}(\bb_{N-1},\bs_{N-1},\btheta_{N-1},\bz_{N-1}) = \sum_{i \in \mathcal{P}} \mathcal{V}_{i,N-1}(\bb_{N-1},\bs_{N-1},\btheta_{N-1},\bz_{N-1})$. Moreover, the optimal policy at time $N-1$ is given by
    \begin{align*}
        \by^*_{N-1} &\in \argmax_{\big\{i \in \mathcal{I}_{N-1} | \sum_{i\in \mathcal{I}_{N-1}} y_{i,N-1}\leq C\big\}} \Bigg(\sum_{i\in \mathcal{P}}(V^1_{i,N} - V^0_{i,N})y_{i,N-1} \Bigg).
    \end{align*}
\end{lemma}
Note that there are two key results in this Lemma: $(i)$ the component value function $\mathcal{V}_{i,N-1}$ depends on the states of all patients for each $i \in \mathcal{P}$; and $(ii)$ only patients from the set of the patients of interest should be visited. The first result can be shown by reformulating the single stage problem as an integer program with totally unimodular constraints \citep{wolsey1999integer}. This means that the solution to its linear programming relaxation is integral, and thus strong duality holds. Then, using a Lagrangian relaxation, the component functions can be formed. The proof of the second result is then completed by considering two cases: $|\mathcal{I}_t|\leq C$ and $|\mathcal{I}_t|>C$. The proof of the first case follows directly from Theorem~\ref{thm-1} and Corollary~\ref{cor-1}, and the second case is shown through a proof by contradiction that uses an exchange argument. Note that this result shows that $\mathcal{V}_{N-1}$ is not additively decomposable, since the resulting component functions $\mathcal{V}_{i,N-1}$ are not functions of  individual patient states but of all patient states. This is because each component is a function of the Lagrange multiplier of the capacity constraint, which is itself a function of the full state of the system. This means that using direct dynamic programming would be computationally complex, leading to the need for approximate approaches.


\subsubsection{Lagrangian Relaxation.} \label{sec:lagrange} One approach for approximating a solution to the multi-patient problem is to consider a Lagrangian relaxation of the problem. This form of relaxation approach is common in developing methods for computing policies for weakly-coupled MDPs \citep{adelman2008relaxations, brown2023strength}.
First, consider the following relaxation of
Equation~\eqref{eq:multi_patient_problem} for a particular set of multipliers $\blambda = \{\lambda_1,...,\lambda_{N-1}\}$:
\begin{subequations}
\begin{align}
  \mathcal{L}(\blambda) =\underset{\bz, \by, \bb,\bs,\btheta }{\mathrm{maximize}} & \quad \sum_{i\in \mathcal{P}}\sum_{t\in \mathcal{T}}\mathbb{P}_\xi (b_{i,t} \leq \delta) + \sum_{t \in \mathcal{T}} \lambda_t \bigg(C- \sum_{i \in \mathcal{P}} y_{i,t}\bigg), \\
\mathrm{subject\,to} \quad & (\bb_{t+1},\bs_{t+1},\btheta_{t+1},\bz_t) = h(\bb_t,\bs_t, \btheta_t, \bz_{t-1},\by_t), \quad \forall t \in \mathcal{T},\\
& y_{i,t},z_{i,t} \in \{0,1\}, \quad \forall  i \in \mathcal{P},t \in \mathcal{T},  \\
    & b_{i,t},s_{i,t},\theta_{i,t} \geq 0, \quad \forall  i \in \mathcal{P}, t \in \mathcal{T}.
\end{align}
\label{eq:lagrangian_joint_formulation}
\end{subequations}

The dual solution is given by $\mathcal{L}^* =\underset{\blambda \geq 0}{\mathrm{minimize}} \:\mathcal{L}(\blambda)$. For a given value of $\blambda$, we can calculate $\mathcal{L}(\blambda)$ using the following dynamic programming equations:
\begin{equation}
\begin{aligned}
    &\mathcal{V}^{\blambda}_t(\bb_t,\bs_t,\btheta_t,\bz_{t-1}) = \sum_{i \in \mathcal{P}} \mathbbm{1}(b_{i,t} \leq \delta) \\ 
    &+ \max_{\by_t \in \{0,1\}^{|\mathcal{P}|}}  \mathbb{E}_{\bxi} \Bigg\{\mathcal{V}^{\blambda}_{t+1}\big(h(\bb_t,\bs_t,\btheta_t,\bz_{t-1},\by_t)\big) + \lambda_t \bigg(C - \sum_{i\in \mathcal{P}} y_{i,t}\bigg) \Bigg\},  \quad \forall t \in \mathcal{T}, \\
&\mathcal{V}^{\blambda}_N(\bb_N,\bs_N,\btheta_N,\bz_{N-1}) = \sum_{i \in \mathcal{P}} \mathbbm{1}(b_{i,N} \leq \delta). \label{eq:lagrange_joint_dp_N}
\end{aligned}  
\end{equation}
Here $\mathcal{L}(\blambda) = \mathcal{V}^{\blambda}_1(\bb_1,\bs_1,\btheta_1,\bz_0)$. There are several key properties of this relaxation:
\begin{proposition}
\label{prop:lagrange_struct}
    For any $\blambda \in \mathbb{R}_+^{|\mathcal{T}|}$, the following properties hold $\forall t \in \mathcal{T}$ and $\bb,\bs,\btheta,\bz \in \mathbb{R}_+^{3\times |\mathcal{P}|}\times\{0,1\}^{|\mathcal{P}|}$.
    \begin{enumerate}
        \item\label{prop:lag-part1} (Weak duality) $\mathcal{V}_t(\bb_t,\bs_t,\btheta_t,\bz_{t-1}) \leq \mathcal{V}_t^{\blambda}(\bb_t,\bs_t,\btheta_t,\bz_{t-1})$.
        \item\label{prop:lag-part2} (Decomposition) $\mathcal{V}_t^{\blambda}$ are additively decomposable. That is: $\mathcal{V}_t^{\blambda}(\bb_t,\bs_t,\btheta_t,\bz_{t-1}) = C \sum_{k = t}^N\lambda_k + \sum_{i \in \mathcal{P}} \mathcal{V}_{i,t}^{\blambda}(b_{i,t},s_{i,t},\theta_{i,t},z_{i,t-1})$ where:
        \begin{equation}
        \label{eq:langrange_singele_patient}
            \mathcal{V}_{i,t}^{\blambda}(b_{i,t},s_{i,t},\theta_{i,t},z_{i,t-1}) = \mathbbm{1}(b_{i,t}\leq \delta) + \max_{y_{i,t} \in \{0,1\}} \mathbb{E}_\xi \mathcal{V}^{\blambda}_{i,t+1}\big(h_i(b_{i,t},s_{i,t},\theta_{i,t},z_{i,t-1},y_{i,t})\big) - \lambda_t y_{i,t}.
        \end{equation}
        \item (Convexity) \label{prop:lag-part1}$\mathcal{V}_t^{\blambda}(\bb_t,\bs_t,\btheta_t,\bz_{t-1})$ is convex in $\blambda$.
    \end{enumerate}
\end{proposition}

The result of this proposition can be thought of as a finite-time and continuous state space analogue to Section 2.3 from \cite{adelman2008relaxations} and Proposition 3.2 from \cite{brown2023strength}. We provide a proof for completeness in Appendix~\ref{app:proofs}. The key difference between this result and prior discrete state results is that, while $\mathcal{V}_t^\lambda$ is convex in our case, it may not be piecewise linear. This is because we assume $\xi_{i,t}$ are continuous random variables, meaning that when taking an expectation we are smoothing over any non-differentiable points from the value function of the future period. Indeed, adding the restriction that $\xi_{i,t}$ only has positive measure on a finite set would recover the piecewise linear structure of the value function. The key insight from this result is that using the Lagrangian relaxation allows the decision maker to compute a single policy based on a collection of single-patient problems. This property will be key in establishing the existence of a dual optimal policy. For simplicity, recalling our prior notation, let $V_{i,t}^{{\blambda},j} = \mathbb{E}_{\bxi}\Big[ \mathcal{V}^{\blambda}_{i,t+1}\big(h_i(b_{i,t},s_{i,t},\theta_{i,t},z_{i,t-1},y_{i,t} = j)\big)\Big]$.
\begin{proposition}
\label{prop:lagrange_single_struct}
    For any $\blambda \in \mathbb{R}^{|\mathcal{T}|_+},b_{i,t},s_{i,t},\theta_{i,t} \geq 0$, $z_{i,t-1} \in \{0,1\}$, and $t \leq N-1$, the dual policy is given by $y^{\blambda}_{i,t}(b_{i,t},s_{i,t},\theta_{i,t},z_{i,t-1},\blambda) = \mathbbm{1}(V_{i,t}^{{\blambda},1} - V_{i,t}^{{\blambda},0} - \lambda_t \geq 0 )$. Moreover: 
    \begin{equation}
        \mathcal{V}^{\blambda}_{i,t}(b_{i,t},s_{i,t},\theta_{i,t},z_{i,t-1}) = \mathbbm{1}(b_{i,t} \leq \delta) + V_{i,t}^{{\blambda},0} + (V_{i,t}^{{\blambda},1} - V_{i,t}^{{\blambda},0} - \lambda_t)^+,
    \end{equation} 
    where $(x)^+ = \max\{0, x\}$ for any $x \in \mathbb{R}$.
\end{proposition} 
This proposition shows that, for a participant $i \not\in \mathcal{I}_t$, the dual policy $y_{i,t}^{\blambda}$ should be set to zero because $\lambda_t \geq 0$. Moreover, it provides a method for computing $\mathcal{V}^{\blambda}_{i,t}$ recursively for a particular set of initial states. Using this result we can  use a subgradient approach similar to \cite{topaloglu2009using}. We present the details of such an approach in Appendix~\ref{app:subgradient-approach}. Alternatively, by discretizing the state of the system, we can approximate a solution to the Lagrangian dual using the linear programming formulation presented in \cite{brown2023strength}. While this approach provides faster solution times then the subgradient approach, it is subject to discretization error.

Using a similar technique to \cite{adelman2008relaxations}, we can obtain a bound on the looseness of the relaxation for the special case where we allow $\lambda_t$ to be state-dependent. 
\begin{proposition}
\label{prop:lagrange_relax_gap}
    Suppose $\lambda_t:\mathbb{R}^{3}\times\{0,1\} \mapsto \mathbb{R}$ and define the relaxation: 
    \begin{equation}
            \mathcal{V}_{i,t}^{\blambda(\cdot)}(b_{i,t},s_{i,t},\theta_{i,t},z_{i,t-1}) = \mathbbm{1}(b_{i,t}\leq \delta) + \max_{y_{i,t} \in \{0,1\}} \mathbb{E}_\xi \mathcal{V}^{\blambda(\cdot)}_{i,t+1}(h_i(b_{i,t},s_{i,t},\theta_{i,t},z_{i,t-1},y_{i,t})) - \lambda_t(\cdot) y_{i,t}
        \end{equation}
        Further define $\blambda^*(\cdot) = \argmin \mathcal{L} \big(\blambda(\cdot)\big)$. Then for any $t \in \mathcal{T}$ and $\bb,\bs,\btheta,z \in \mathbb{R}^{3 \times |\mathcal{P}|}\times\{0,1\}^{|\mathcal{P}|}$: $\mathcal{V}_t^{\blambda^*(\cdot)}(\bb,\bs,\btheta,z) - \mathcal{V}_t(\bb,\bs,\btheta,z) \leq 2(N-t)$.
\end{proposition}

This result is significant since the derived bound agrees with the order and structure of the bound in \cite{adelman2008relaxations}, but adapted for a finite-horizon setting instead of a discounted infinite-horizon setting. Note that if our state space were discrete, this proposition would apply to our non-state-dependent Lagrangian relaxation using existing results  due to our capacity constraint being totally unimodular \citep{brown2023strength}. However, since our state space is continuous, this may not necessarily hold. That said, when using approximate DP computationally, this would indicate the Lagrangian relaxation proposed is a sharp approximation up to discretization error. Moreover, this proposition directly implies that the Lagrangian relaxation forms an asymptotically optimal policy with respect to the number of participants.

\begin{corollary}
    \label{cor:lagrange_ass_opt} Given the assumptions of Proposition \ref{prop:lagrange_relax_gap},
    as the participant population $|\mathcal{P}| \rightarrow \infty$, for all $t \in \mathcal{T}$: 
    \begin{equation}
        \frac{\mathcal{V}_t^{\blambda^*(\cdot)}(\bb,\bs,\btheta,z) - \mathcal{V}_t(\bb,\bs,\btheta,z) }{|\mathcal{P}|} \rightarrow0
    \end{equation}
\end{corollary}
The proof of the corollary is trivial and follows directly as a consequence of the bound in Proposition~\ref{prop:lagrange_relax_gap} being independent of $|\mathcal{P}|$. The implication of this corollary is that, for large patient populations, we would expect the value function computed by the relaxation to be quite close to the true value function. This means that we would expect the policy generated by the relaxation to be close to the optimal policy for large patient cohorts.

\subsubsection{Index Policy.} \label{sec:whittles_index} An alternative to the Lagrangian relaxation, which involves separate multipliers for each decision epoch, is a Whittle's style relaxation \citep{whittle1988restless}. The key difference here is adding the restriction to the Lagrangian relaxation shown in \eqref{eq:lagrangian_joint_formulation} that $\lambda_t = \omega$ for all $t \in \mathcal{T}$. Essentially, this is the equivalent of averaging the capacity constraints across all time periods. The resulting relaxation takes the following form:
\begin{subequations}
\begin{align}
  \mathcal{W}(\omega) =\underset{\bz, \by, \bb,\bs,\btheta }{\mathrm{maximize}} & \quad \sum_{i\in \mathcal{P}}\sum_{t\in \mathcal{T}}\mathbb{P}_\xi (b_{i,t} \leq \delta) + \frac{\omega}{N}\sum_{t \in \mathcal{T}} \bigg(C- \sum_{i \in \mathcal{P}} y_{i,t}\bigg) \\
\mathrm{subject\,to} \quad & (\bb_{t+1},\bs_{t+1},\btheta_{t+1},\bz_t) = h(\bb_t,\bs_t, \btheta_t, \bz_{t-1},\by_t) \quad \forall t \in \mathcal{T}\\
& y_{i,t},z_{i,t} \in \{0,1\}, \quad \forall  i \in \mathcal{P},t \in \mathcal{T},  \\
    & b_{i,t},s_{i,t},\theta_{i,t} \geq 0, \quad \forall  i \in \mathcal{P}, t \in \mathcal{T},
\end{align}
\label{eq:lagrangian_whittles}
\end{subequations}

Since the multipliers are more restrictive than in \eqref{eq:lagrangian_joint_formulation} it is immediately clear that $\mathcal{W}(\omega^*) = \min_{\omega\geq0} \mathcal{W}(\omega) \geq \min_{\blambda \geq 0} \mathcal{L}({\blambda})$. While this relaxation is looser, it does provide a more computationally tractable heuristic for choosing which patients to visit. In particular, we can show that this will result in an index-based policy that will allow us to rank patients at each time period by said index and visit them according to this priority. To establish these properties, consider the dynamic programming formulation of \eqref{eq:lagrangian_whittles}.
\begin{equation}
\begin{aligned}
    &\mathcal{V}^{\omega}_t(\bb_t,\bs_t,\btheta_t,\bz_{t-1}) = \sum_{i \in \mathcal{P}} \mathbbm{1}(b_{i,t} \leq \delta) \\ &+ \max_{\by_t \in \{0,1\}^{|\mathcal{P}|}}  \mathbb{E}_{\bxi} \Bigg\{\mathcal{V}^{\omega}_{t+1}\big(h(\bb_t,\bs_t,\btheta_t,\bz_{t-1},\by_t)\big) + \frac{\omega}{N} \bigg(C - \sum_{i\in \mathcal{P}} y_{i,t}\bigg) \Bigg\}, \quad \forall t \in \mathcal{T}, \\
    &\mathcal{V}^{\omega}_N(\bb_N,\bs_N,\btheta_N,\bz_{N-1}) = \sum_{i \in \mathcal{P}} \mathbbm{1}(b_{i,N} \leq \delta). \label{eq:whittles_joint_dp_N}
\end{aligned}  
\end{equation}

In the formulation above, note that the maximum is taken with respect to the visit decision $\by_t$, which appears both in $\mathcal{V}_{t+1}^\omega(\cdot)$, where it affects the transition of states $\bb_t$, $\bs_t$, $\btheta_t$, and $\bz_{t-1}$ via dynamics function $h(\cdot)$, and in Whittle's-style relaxation term associated with the capacity constraint. Using a similar reasoning to the Lagrangian relaxation, we can obtain the following result for the structure of the value functions in \eqref{eq:whittles_joint_dp_N}:
\begin{proposition}
\label{prop:whittles_struct}
    For any $\omega\geq 0$ the following properties hold $\forall t \in \mathcal{T}$ and $\bb,\bs,\btheta,\bz \in \mathbb{R}_+^{3\times |\mathcal{P}|}\times\{0,1\}^{|\mathcal{P}|}$.
    \begin{enumerate}
        \item (Weak duality) $\mathcal{V}_t(\bb_t,\bs_t,\btheta_t,\bz_{t-1}) \leq \mathcal{V}_t^{\omega}(\bb_t,\bs_t,\btheta_t,\bz_{t-1})$
        \item (Decomposition) $\mathcal{V}_t^{\omega}$ are additively decomposable. That is: $\mathcal{V}_t^{\omega}(\bb_t,\bs_t,\btheta_t,\bz_{t-1}) = \frac{(N-t)C \omega}{N} + \sum_{i \in \mathcal{P}} \mathcal{V}_{i,t}^{\omega}(b_{i,t},s_{i,t},\theta_{i,t},z_{i,t-1})$ where:
        \begin{equation}
        \label{eq:whittles_single_patient}
            \mathcal{V}_{i,t}^{\omega}(b_{i,t},s_{i,t},\theta_{i,t},z_{i,t-1}) = \mathbbm{1}(b_{i,t}\leq \delta) + \max_{y_{i,t} \in \{0,1\}} \mathbb{E}_\xi \mathcal{V}^{\omega}_{i,t+1}\big(h_i(b_{i,t},s_{i,t},\theta_{i,t},z_{i,t-1},y_{i,t})\big) -\frac{\omega}{N} y_{i,t}
        \end{equation}
        \item (Indexability) $\mathcal{V}_{i,t}^{\omega}(b_{i,t},s_{i,t},\theta_{i,t},z_{i,t-1})$ is convex and monotonically nonincreasing in $\omega$, moreover it is bounded.
    \end{enumerate}
\end{proposition}

Indexability is a key property in the study of restless multiarmed bandits since it ensures the existence of an index. Essentially, it means that the set of states for which a patient would be visited at a given time period increases as $\omega$ decreases. Thus there exists an $\omega_i^*(b_{i,1},s_{i,1},\theta_{i,1},z_{i,0}) = \min\{\omega :  \mathcal{V}_{i,0}^{\omega_c}(b_{i,1},s_{i,1},\theta_{i,1},z_{i,0}) - \mathbbm{1}(b_{i,1} \leq \delta) = \omega\}$, which can be interpreted as the smallest reward the decision would be willing to accept to be indifferent between visiting and not visiting a participant. These $\omega_i^*$ can then be thought of as the index values for each of the participants. At each decision epoch, the decision maker would rank participants in the set $\mathcal{I}_t$ by these indices and choose the first $C$. The index itself can be computed using a bisection procedure presented in Algorithm \ref{alg:whittle}. 
\begin{algorithm2e}[H]
    \DontPrintSemicolon
    \KwIn{ $\bb_1,\bs_1,\btheta_1,\bz_{0}, \epsilon$}\;
    \For{$i \in \mathcal{P}$}{
    Set $\underline{\omega} = 0, \overline{\omega} = \mathcal{V}_{i,0}^0(b_{i,1},s_{i,1},\theta_{i,1},z_{i,0}) $ \;
    \While{$\overline{\omega} -\underline{\omega} > \epsilon$}{
    Set: $\omega_c \gets \frac{\overline{\omega} +\underline{\omega}}{2}$\;
    Compute: $\mathcal{V}_{i,0}^{\omega}(b_{i,1},s_{i,1},\theta_{i,1},z_{i,0})$\;
    \eIf{$\mathcal{V}_{i,0}^{\omega_c}(b_{i,1},s_{i,1},\theta_{i,1},z_{i,0}) - \mathbbm{1}(b_{i,1} \leq \delta)\geq \omega_c $}{
    $\underline{\omega} \gets \omega_c$ \;
    }{
    $\overline{\omega} \gets \omega_c$ \;
    }
    }
    Set: $\omega_i \gets \frac{\overline{\omega}+\underline{\omega}}{2}$
    }
    \Return $\{\omega_i\}_{i\in \mathcal{P}}$
    \caption{Bisection Procedure for Index-Based Relaxation \label{alg:whittle}}
    \end{algorithm2e}

Note that $\omega_i^*$ must be upper bounded by the unconstrained optimal number of visits, meaning it must lay in a compact interval. This combined with the indexability property ensuring the value functions are monotonic implies that the bisection search will terminate in finite time when computing the index. Like the previous Lagrange case, it is still clear that participants ought not to be visited unless they are in the set of the patients of interest previously described. Another observation of the index policy above is that each index $\omega_{i,t}$ is upper bounded by the total number of periods in control that can be achieved by a single-patient policy. Thus an optimistic index policy would be to substitute these values in for the indices and sort participants by these levels. Unlike the previous Lagrangian case, it is not straightforward to establish that the Whittle's index policy is asymptotically optimal. While like the full Lagrangian relaxation, $\mathcal{W}(\omega^*)$ can be shown to be asymptotically optimal with respect to $|\mathcal{P}|$, using a similar argument to that in \cite{weber1990index}, this does not provide a guarantee on the index based policy that comes from this relaxation. Existing asymptotic results for the Whittle's index critically rely on the state space being discrete and the contraction properties of infinite-horizon sequential decision-making problems for their proofs, therefore they are not readily applicable in our setting \citep{verloop2016asymptotically,gast2023exponential}. That said the Whittle's index provides a more computationally tractable method for selecting patient visits, especially under partial information, and as per our results in Section \ref{sec:experiments} performs well computationally.

\subsubsection{Implementation of the Optimal Visit Policy.}
\label{sec:implementation-opt-policy}

To solve the multi-patient problem, we implement a solution approach that we refer to as the \emph{Enrollment Algorithm (EA)}. We use this approach due to the challenges of solving the multi-patient problem given our continuous state space and lack of additive decomposability of the value function.
Intuitively, the EA incorporates the results from Corollary~\ref{cor-1} to first identify $\mathcal{I}_t$ for each period $t$ (see Figure~\ref{fig:visit-decision-tree} for a graphical representation of the conditions defining $\mathcal{I}_t$). If $|\mathcal{I}_t| \leq C$, then we set $y_{i,t}=1 \:\forall i \in \mathcal{I}_t$. If $|\mathcal{I}_t| > C$, then we use one of six tie-breaking rules to rank the patients and select the first $C$ patients to be visited. The six tie-breaking rules used to sort patients $i \in \mathcal{I}_t$ (and the intuition behind them) are:
\begin{enumerate}
    \item \emph{EA ascending FBG:} Order patients by ascending $b_{i,t}$. The provider may have more patients with their FBG in-control if they focus on patients with lower FBG (easier to control).
    \item \emph{EA descending FBG:} Order patients by descending $b_{i,t}$. The provider may have a greater impact if they can flip patients with high FBG from out of control to in-control, especially when the patients with lower FBG have slow disease progression (in-control even if unvisited).
    \item \emph{EA value-to-go:} Estimate the value-to-go of each patient by solving a single-patient problem using the one-step look-ahead policy that is based on Corollary~\ref{cor-1}. Order patients by descending value-to-go. The value-to-go for each patient is determined based on the number of periods that the patient would have their FBG in control if we were to implement the optimal policy for a single-patient problem, that is, $\tilde{\mathcal{V}}_{i,t}=\sum\limits_{t' \in \mathcal{T} | t' \geq t} \mathbbm{1}(b_{i,t'} \leq \delta)$. The provider chooses the patients that would have a greater number of periods in control in the single patient problem.
    \item \emph{EA value-to-go/visits:} Estimate the value-to-go and the number of visits required in the remaining periods for each patient by solving a single-patient problem. Order patients by descending ratio of value-to-go divided by the number of visits, $\tilde{\mathcal{V}}_{i,t}/L_{i,t}$. We calculate the number of visits ($L_{i,t}$) that the patient would require from the present period until the end of the horizon under the optimal single-patient problem policy, that is, $L_{i,t}=\sum\limits_{t' \in \mathcal{T} | t' \geq t} y_{i,t'}$. The provider divides the total number of periods in control for each patient by the number of visits required to obtain a ratio (i.e., return/cost). The ratios are then ranked in descending order. 
    \item \emph{EA Lagrangian}: Prior to solving the multi-patient problem, we numerically solve the Lagrangian relaxation in \eqref{eq:lagrange_joint_dp_N} for a limited number of periods. The value function is then used as an approximation of the value to go in a roll-out policy \citep{bertsekas2012dynamic}.
    \item \emph{EA Whittle's index}: At each period $t$, the Whittle's indices are calculated for each patient $i \in \mathcal{I}_t$ using the bisection procedure in Algorithm~\ref{alg:whittle} and ranked in descending order.  While the relaxation using the Whittle's index is upper bounded by the Lagrangian relaxation (due to the restriction that all multipliers must be the same for all periods), it is easily computed for each patient individually and can be parallelized.
    %
\end{enumerate}

\section{Numerical Experiments}
\label{sec:experiments}

In this section, we describe the data and experimental setup used to evaluate our approach. Section~\ref{sec:experiments-setup-overall} describes the overall simulation framework including how we incorporated uncertainty in disease progression, the baseline approaches used for comparison, and the evaluation metrics used. Section~\ref{sec:experiments-setup-nanohealth} presents the results from simulations performed with NanoHealth's data. Section~\ref{sec:experiments-setup-scenarios} describes how we generated artificial patient cohorts, which we refer to as \emph{expanded scenarios}, to evaluate the generalizability of our approach, and Section~\ref{sec:scen-results} presents the results from the expanded scenarios. We evaluate the performance of our methods with partial information in Section \ref{sec:exp_w_imperfect_info}. Finally, Sections~\ref{sec:results-manag-implications} and \ref{sed:limit_and_etend} describe broad managerial implications obtained through our experiments, as well as limitations and potential extensions of our methods.

\subsection{Simulation Framework}
\label{sec:experiments-setup-overall}

Unless otherwise stated, we solve the multi-patient CHW visitation problem described in Section~\ref{sec:provider_model} using a planning horizon of $N=60$ periods, corresponding to five years of monthly visits (in line with NanoHealth operations). We conducted simulations for capacity levels ranging from 5\% to 100\% in 5\% increments, where the capacity level is expressed as a percentage of the per period visit capacity ($C$) divided by the number of patients in the community being served ($|\mathcal{P}|$). All simulations were run using Python 3.7 on a PC with 16 GB of RAM and an Intel Core i5 processor.  The run times for all experiments are reported in Table~\ref{table:computation-times-full-info}.

\subsubsection{Data.} \label{sec:data-description} Our dataset includes anonymous longitudinal data about a patient cohort treated by NanoHealth, which includes 378 patients that received at least six visits through NanoHealth's intervention in Hyderabad, India. For each patient, our data includes an FBG reading each time they were visited by a CHW (for both screening and management visits).

\subsubsection{Uncertainty in Disease Progression.}
\label{sec:experiments-setup-uncertain-disease-progression}

We simulated $\xi_{i,t}$ as normally distributed with mean zero and standard deviation $\sigma$, where $\sigma$ was estimated using historical data from NanoHealth by calculating the sample standard deviation of the change in FBG between successive periods across all patients. We solve 10 replications of each instance each with a different random seed. 

\subsubsection{Baseline Heuristics.} We use four naive baseline heuristics as benchmarks:
\label{sec:baseline-heuristics}

\begin{itemize}
    \item \emph{Visit-no-one:} No patients are visited during the planning horizon.
    \item \emph{Visit-everyone:} Every patient is visited at every period.
    \item \emph{Descending FBG:} Visit $C$ patients ranked by descending FBG.
    \item \emph{Ascending FBG:} Visit $C$ patients ranked by ascending FBG.
\end{itemize}
Note that the last two baseline heuristics differ from their EA counterparts in that patients are not filtered by identifying set $I_t$ prior to ranking in each period $t$.

\subsubsection{Evaluation Metrics.}
\label{sec:experiments-setup-metrics}
To compare our solution algorithms, we introduce the concept of patient-periods in control (PPC), i.e., the number of patients with their FBG in control (under the threshold $\delta$) at the end of each period. By computing the total number of PPC across all periods, we can easily compare the relative performance of each heuristic for the full planning horizon. We also compare our algorithms using the number of screening visits, total enrollment, and FBG. 



\subsection{NanoHealth Case Study}
\label{sec:experiments-setup-nanohealth}

In this section, we describe the experiments and results using a real patient cohort. 
To obtain personalized patient parameters required to solve the multi-patient provider problem, we first solved the parameter estimation problem described in Section~\ref{sec:parameter-estimation} for each patient using their actual visit history. See Appendix~\ref{EC_patparamresults} for details. According to the parameter estimation results, there is significant heterogeneity in patient parameters (e.g., disease progression, enrollment effect, visit effect) and our simulation results indicate that visit allocation decisions and glycemic control are both improved through a planning approach that models individual patient decisions. 


\subsubsection{PPC Performance.}
\label{sec:nanohealth-ppc-performance}

\begin{figure}[ht]
\centering
\begin{subfigure}[ht]{0.45\textwidth}
\centering
\includegraphics[width=\textwidth, trim={0 0.5cm 0 0}, clip]{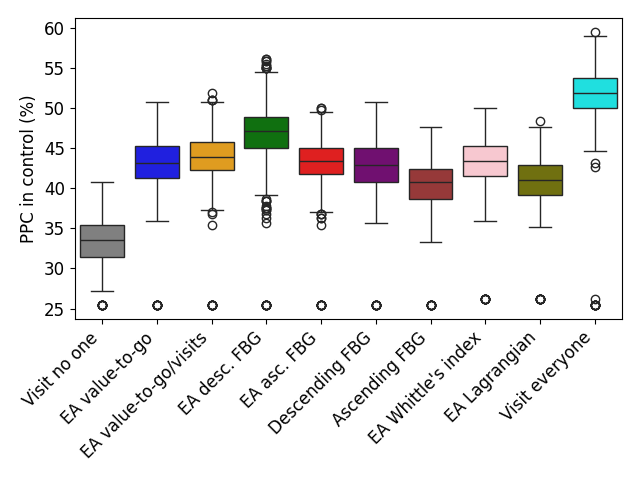}
\caption{\label{fig:nanohealth-ppc-boxplot-cap151}}
\end{subfigure}
\hfill
\begin{subfigure}[ht]{0.53\textwidth}
\centering
\includegraphics[width=\textwidth, trim={0 0 0 0.4cm}, clip]{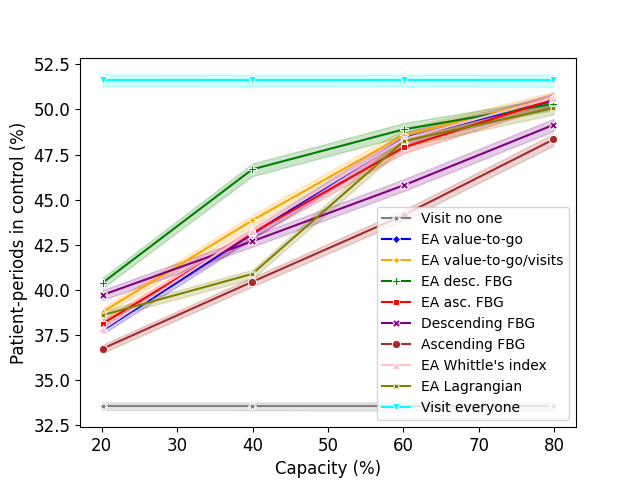} 
\caption{\label{fig:nanohealth-ppc-lineplot}}
\end{subfigure}\vspace{10pt} 
\caption{(a) Boxplots showing the interquartile range of PPC for each heuristic at 40\% capacity; (b) Line graph of PPC for each heuristic for varying capacity levels (5\% increments) with shaded 95\% confidence intervals. \label{fig:results-nanohealth-ppc}}
\end{figure}

Figure~\ref{fig:nanohealth-ppc-boxplot-cap151} shows boxplots of PPC for each algorithm when the visit capacity per period corresponds to 40\% of the total number of patients in the cohort. We chose this capacity level because we observe the largest divergence in PPC performance across the different algorithms. We observe that EA with descending FBG achieves the best performance with an average (standard deviation) PPC of 46.7\% (2.1\%), followed by EA value-to-go/visits with a PPC of 43.8\% (1.8\%). These algorithms have a relative improvement upon the best benchmark heuristic (ascending FBG) of 9.4\% and 2.7\%, respectively. 

Figure~\ref{fig:nanohealth-ppc-lineplot} shows the PPC at varying capacity levels in 20\% increments. We observe that the different algorithms are comparable at 20\% and 80\% capacities (excluding the benchmarks visit-no-one and visit-everyone). As the capacity increases from 20\%, the EA with descending FBG has the fastest gain in performance, reaching a 9.4\% better performance at 40\% capacity with respect to the best benchmark heuristic. Furthermore, to achieve at least 45\% PPC, the EA with descending FBG requires approximately 40\% capacity, while the ascending FBG heuristic requires approximately 60\% capacity, a 20\% absolute (50\% relative) increase.

\subsubsection{Visit Types and Patient Enrollment.}
\label{sec:nanohealth-visit-types}

\begin{figure}[ht]
\centering
\begin{subfigure}{0.48\textwidth}
\centering
\includegraphics[width=\textwidth, trim={0.6cm 0 1.55cm 0.25cm}, clip]{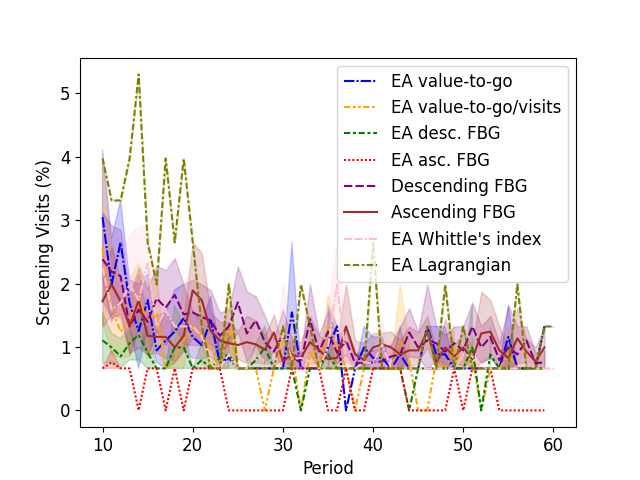}
\caption{\label{fig:nanohealth-screening-cap151}}
\end{subfigure}\vspace{10pt} 
\hfill
\begin{subfigure}{0.48\textwidth}
\centering
\includegraphics[width=\textwidth, trim={0.4cm 0 1.55cm 0.25cm}, clip]{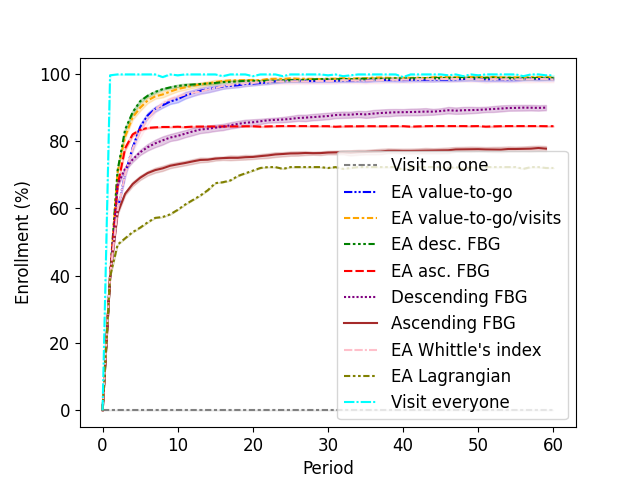}
\caption{\label{fig:nanohealth-enrollment-cap151}}
\end{subfigure}
\caption{(a) Line graph showing the proportion of CHW visits that were screening visits at 40\% capacity starting at period 10 with shaded 95\% confidence intervals; (b) Line graph showing the proportion of patients enrolled in treatment at 40\% capacity with shaded 95\% confidence intervals. \label{fig:results-nanohealth-visit-types}}
\end{figure}

Figure~\ref{fig:results-nanohealth-visit-types} illustrates the long-term patterns in screening and patient enrollment. Figure~\ref{fig:nanohealth-screening-cap151} displays the proportion of CHW visits that were screening visits across the planning horizon. The first 10 periods in the planning horizon are omitted since the short-term screening patterns are similar for all heuristics. We observe that the EA implementations have a lower percentage of visits that are dedicated to screening (except for the Lagrangian), in addition to cyclical behavior which suggests that some patients may drop out of treatment and require additional screening visits to re-enroll. Furthermore, we observe that the benchmark heuristics (descending FBG and ascending FBG) maintain a higher overall percentage of screening visits. However, Figure~\ref{fig:nanohealth-enrollment-cap151}, which shows the proportion of enrolled patients, demonstrates that these two benchmark heuristics also have the lowest percentage of patients enrolled in treatment during most of the planning horizon. Intuitively, these results suggest that the screening visits conducted by the benchmark heuristics are not allocated as efficiently as the screening visits conducted by the EA implementations. In fact, most EA heuristics have an enrollments surpassing 95\%, while the enrollment for the descending FBG and ascending FBG heuristics never surpasses 81\% and 93\%, respectively.

\subsubsection{FBG Distribution.}
\label{sec:nanohealth-fbg-dist}

\begin{figure}[ht]
    \centering
    \includegraphics[width=0.45\textwidth]{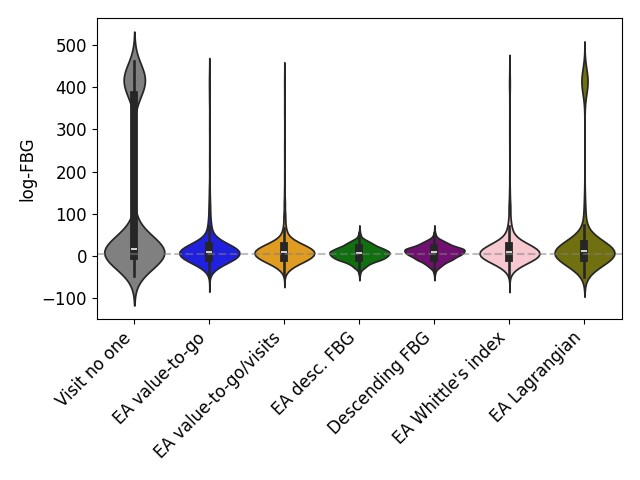}
    \caption{Violin plots showing the distribution of log-FBG values at the end of the planning horizon for seven different heuristics at 40\% capacity (NanoHealth). The gray dashed line indicates the threshold $\delta$.}
    \label{fig:nanohealth-fbg-dist-violin}
\end{figure}

Figure~\ref{fig:nanohealth-fbg-dist-violin} shows violin plots for the distribution of log-FBG levels at the end of the planning horizon. The visit-no-one benchmark shows a bimodal distribution due to the fast FBG increase experienced by some patients, and slow increase or stabilization in FBG experienced by other patients, leading to two distinct peaks. Because we modeled disease progression proportionally to current FBG levels, an artifact of our model is that some patients present extremely high FBG levels that would not occur in practice. In these cases, life-threatening conditions would likely occur before those FBG levels were reached.

The visit-no-one policy achieves a median and 90th percentile log-FBG of 114.1 and 114.6, respectively. Although the Lagrangian approximation has a similar shape with a long tail, the median and 90th percentile are 47.3 and 47.8 (relative improvements of 58.5\% and 58.3\%), respectively. The best baseline method was descending FBG with a median and 90th percentile of 7.3 and 7.9, respectively. EA descending FBG performed best overall with a median and 90th percentile of 5.5 and 6.3, improving upon the best baseline method (ascending FBG) by 24.5\% and 21.1\% in terms of median and 90th percentile, respectively.

\subsection{Expanded Scenarios' Setup}
\label{sec:experiments-setup-scenarios}

We evaluate the generalizability of our approach through simulation experiments with expanded scenarios (i.e., artificial patient cohorts) that have varying patient parameters. For each scenario, we evaluate the performance of the four EA implementations (described in Section~\ref{sec:implementation-opt-policy}) and the four baseline heuristics (described in Section~\ref{sec:baseline-heuristics}). To generate the scenarios, we first created five patient groups derived from observations of real world data and conversations with our collaborator -- specifically, each group is represented by a ``patient-type" that includes seven parameters (see Table~\ref{table:clusters-and-scenarios}). The five groups (A, B, C, D, and E) have their parameter centroids represented in Figure~\ref{fig:group_centroids}. We combined two patient types per scenario, totaling 10 scenarios with 50\% of patients from each group.  After extensive computational experiments, we picked a subset of two illustrative scenarios (denoted Scenarios 2 and 3), each one with 378 patients. We also built a scenario with 20\% of patients from each group (denoted Scenario 1), totaling 756 patients. Figure~\ref{fig_artscen} provides a graphical representation of the composition of the three scenarios using the patient groups. 


\begin{figure}[!ht]
    \centering
   \begin{tikzpicture}[font=\sffamily]
    \begin{axis}[
        width  = 0.9*\textwidth,
        height = 5cm,
        major x tick style = transparent,
        major y tick style = transparent,
        ybar=2*\pgflinewidth,
        bar width=6pt,
        ymajorgrids = true,
        ylabel = {Parameter level},
        symbolic x coords={$p$,$\mu$,$\alpha$,$\theta_0$,$\lambda$,$s_0$,$\beta$},
        xtick = data,
        scaled y ticks = false,
        enlarge x limits=0.1,
        ymin=0,
        legend cell align=left,
        legend style={font=\footnotesize\sffamily},
        legend style={
                at={(0.98,0.95)},
                anchor=north east,
                column sep=1ex
        }
    ]
        \addplot[style={blue,fill=blue,mark=none}]
            coordinates {($p$, 0.05) ($\mu$,0.025) ($\alpha$,0.1) ($\theta_0$,0.7) ($\lambda$,0.5) ($s_0$,1) ($\beta$,0.3)};

        \addplot[style={red,fill=red,mark=none}]
             coordinates {($p$,5) ($\mu$,4) ($\alpha$,2) ($\theta_0$,0.7) ($\lambda$,0.5) ($s_0$,0.2) ($\beta$,1.5)};

        \addplot[style={green!60!black,fill=green!60!black,mark=none}]
             coordinates {($p$,5) ($\mu$,2) ($\alpha$,4) ($\theta_0$,0.7) ($\lambda$,0.5) ($s_0$,0.2) ($\beta$,1.5)};

        \addplot[style={violet,fill=violet,mark=none}]
             coordinates {($p$,7.5) ($\mu$,4) ($\alpha$,2) ($\theta_0$,0.7) ($\lambda$,0.5) ($s_0$,0.2) ($\beta$,1.5)};

        \addplot[style={orange,fill=orange,mark=none}]
             coordinates {($p$,0.05) ($\mu$,0.025) ($\alpha$,0.35) ($\theta_0$,2) ($\lambda$,1.5) ($s_0$,0.2) ($\beta$,1.5)};

        \legend{Group A,Group B,Group C,Group D,Group E}
    \end{axis}
\end{tikzpicture}

    \caption{Visualization of the group centroids used in truncated normal sampling procedure.}
    \label{fig:group_centroids}
\end{figure}

We build each scenario (i.e., artificial community) in three steps. First, we sample the patient state dynamics parameters from each group described in Table~\ref{table:clusters-and-scenarios}. 
Patient parameters are assumed to be independent and to have constant variance across different groups, therefore only the mean vector differs for each group and the parameters can be sampled using truncated normal distributions. We use a lower bound of zero for the truncation based on the assumption that the parameters being sampled are non-negative. Second, we set the discount factors, $\rho_i$ and $\gamma_i$, to 0.2 for all patients $i \in \mathcal{P}$ in all scenarios because the majority of real world patients had $\rho=\gamma=0.2$ after solving the parameter estimation problem in  Section~\ref{sec:parameter-estimation} (over 93\% and over 95\% for $\rho$ and $\gamma$, respectively). Recall that both $\gamma$ and $\rho$ are contained in the interval $(0,1)$, where 0 indicates that a state immediately returns to its steady state level ($s_{i,0}$ and $\theta_{i,0}$ for $s_{i,t}$ and $\theta_{i,t}$, respectively), while 1 indicates that state at time $t$ is fully determined by the immediate prior state ($s_{i,t-1}$ and $\theta_{i,t-1}$), with the addition or subtraction of $\beta_i$ or $\lambda_i$ in the case of an increase in adverse factors or a decrease in perception of adverse factors, respectively. The empirical results indicate that most patients tend to quickly return to their steady-state levels of $s_{i,0}$ and $\theta_{i,0}$. Third, we sample initial FBG values (i.e., $b_{i,0}$) using a normal distribution fit to the initial FBG values from the NanoHealth cohort. We confirmed that initial FBG was approximately normally distributed using a Kolmogorov–Smirnov at a significance level of 0.05.

\begin{figure}[!ht]
\centering
\sffamily
\begin{tikzpicture}
[
    pie chart,
    slice type={clusterD}{teal!50}{bricks},
    slice type={clusterE}{red!35!yellow!95!black}{crosshatch dots},
    slice type={clusterG}{red!70}{north east lines},
    slice type={clusterJ}{blue!40}{checkerboard},
    slice type={clusterP}{red!80!yellow!70}{horizontal lines},
    pie values/.style={font={\small}},
    scale=1.35
]

    \pie[shift={(0.3cm, 0cm)},values of caffe/.style={pos=1.1}]{Scenario 1}{20/clusterD,20/clusterE,20/clusterG,20/clusterJ,20/clusterP}
    \pie[shift={(3.3cm, 0cm)},values of caffe/.style={pos=1.1}]%
        {Scenario 2}{50/clusterE,50/clusterJ}
    \pie[shift={(6.3cm, 0cm)},values of caffe/.style={pos=1.1}]%
        {Scenario 3}{50/clusterE,50/clusterP}

    \legend[shift={(8.5cm,1cm)}]{{Group A}/clusterD}
     \legend[shift={(8.5cm,0.7cm)}]{{Group B}/clusterE}
      \legend[shift={(8.5cm,0.4cm)}]{{Group C}/clusterG}
    \legend[shift={(8.5cm,0.1cm)}]{{Group D}/clusterJ}
    \legend[shift={(8.5cm,-0.2cm)}]{{Group E}/clusterP}

\end{tikzpicture}
\caption{Graphical illustration of the composition of the artificial scenarios. \label{fig_artscen}}
\end{figure}




\subsection{Expanded Scenarios' Results}
\label{sec:scen-results}

\subsubsection{PPC Performance.}
\label{sec:scen-ppc-performance}

\begin{figure}[ht] 
    \centering
    \begin{subfigure}{0.32\textwidth}
        \centering
        \includegraphics[width=\linewidth, trim={0.6cm 0 1.55cm 0.25cm}, clip]{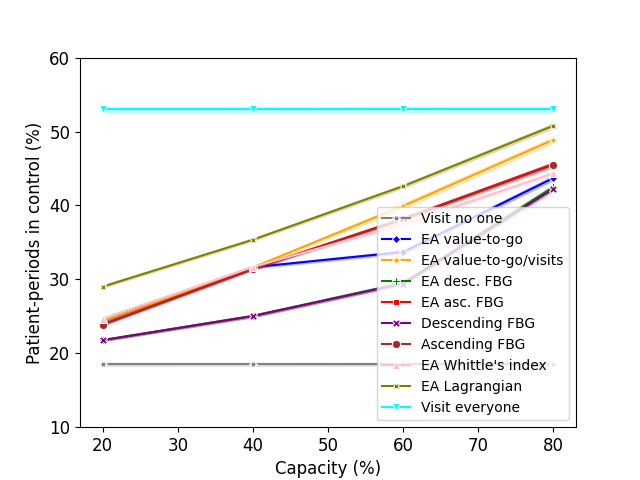}
        \caption{Scenario 1 \label{fig:scen1-ppc-varying-capacity}}
    \end{subfigure}
    \begin{subfigure}{0.32\textwidth}
        \centering
        \includegraphics[width=\linewidth, trim={0.6cm 0 1.55cm 0.25cm}, clip]{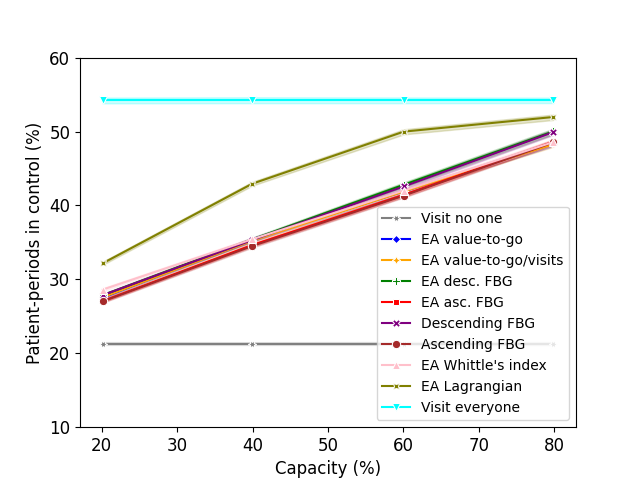}
        \caption{Scenario 2 \label{fig:scen2-ppc-varying-capacity}}
    \end{subfigure}
    \begin{subfigure}{0.32\textwidth}
        \centering
        \includegraphics[width=\linewidth, trim={0.6cm 0 1.55cm 0.25cm}, clip]{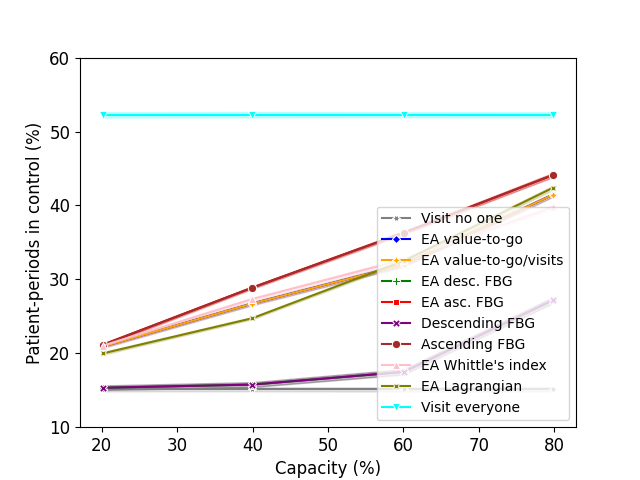}
        \caption{Scenario 3 \label{fig:scen3-ppc-varying-capacity}}
    \end{subfigure}
    \vspace{10pt}
    \caption{Line graphs for different scenarios showing glycemic control at various capacity levels with shaded 95\% confidence intervals. \label{fig:scen-glycemic-control-varying-capacities}}
\end{figure}

Figure~\ref{fig:scen-glycemic-control-varying-capacities} shows one plot per scenario for the percentage of patient-periods in control (PPC) as a function of capacity. For Scenario 1, the EA Lagrangian approximation performed best across all capacity levels, with an absolute PPC improvement over the best baseline method (descending FBG) of between 4.0 and 5.3, depending on the capacity level. For Scenario 2, the EA Lagrangian approximation also performed best across all capacity levels, with an absolute PPC improvement over the best baseline method (ascending FBG) of between 2.0 and 7.6, depending on the capacity level. For Scenario 3, EA ascending FBG and the baseline ascending FBG had the same performance at all capacity levels. 

Alternatively, we can can compare the capacity required for the best naive heuristic to perform comparably to the best EA implementation. For example, in Scenario 1, we can see that the best EA implementation is able to achieve approximately 35\% PPC with 40\% capacity, while the best baseline method (Descending FBG) requires roughly 60\% capacity to achieve the same PPC (a relative increase of 50\%).


By analyzing Figure~\ref{fig:scen-glycemic-control-varying-capacities} across different scenarios, we note that the EA Lagrangian and EA value-to-go/visits achieve a superior performance for Scenarios 1 and 2, while EA ascending FBG and ascending FBG have a superior performance to their counterparts (EA descending FBG and descending FBG) at lower capacities, which suggests that when the capacity is highly constrained, the PPC performance can be improved the most by focusing on patients with the lowest FBG levels (easier to maintain in control). This effect is clear for scenarios where patients have a fast disease progression (e.g., Scenario 2). Also note that EA ascending FBG and ascending FBG have a comparable performance for Scenario 2, whereas EA ascending FBG performs significantly better for Scenarios 1 and 3. This phenomenon is explained by the fast disease progression ($p$), moderate enrollment effect ($\mu$), and high increase in adverse factors ($\beta$) for both groups that compose scenario 2 (B and D). Because $\mu < p$, being enrolled is not enough to deter the disease progression for these patients. Management visits have to be planned carefully to improve PPC while avoiding drop outs due to the high $\beta$, which the EA ascending FBG accomplishes by filtering the patients of interest at each period prior to ranking by ascending FBG. In addition to reducing drop out rates, our EA implementations seek to allocate limited resources efficiently by visiting an enrolled patient who would not drop out if unvisited only if it provides a strict increase in their benefit function (see Theorem~\ref{thm:suff_nec_cond} and Corollary~\ref{cor-1}).

Our PPC results in Sections~\ref{sec:experiments-setup-nanohealth} and \ref{sec:scen-results} show differing findings with respect to the best performing EA heuristic for each cohort/scenario. While EA Descending FBG performed best for NanoHealth, EA Lagrangian performed best for Scenarios 1 and 2, and EA Ascending FBG performed best for Scenario 3. In Section~\ref{sec:alg-perf-insights}, we propose insights into the differing performances.

\subsubsection{Visit Types and Patient Enrollment.}
\label{sec:scen-visit-types}
\begin{figure}
    \centering
    \begin{subfigure}{0.32\textwidth}
        \centering
        \includegraphics[width=\linewidth, trim={0.3cm 0 1.55cm 0.25cm}, clip]{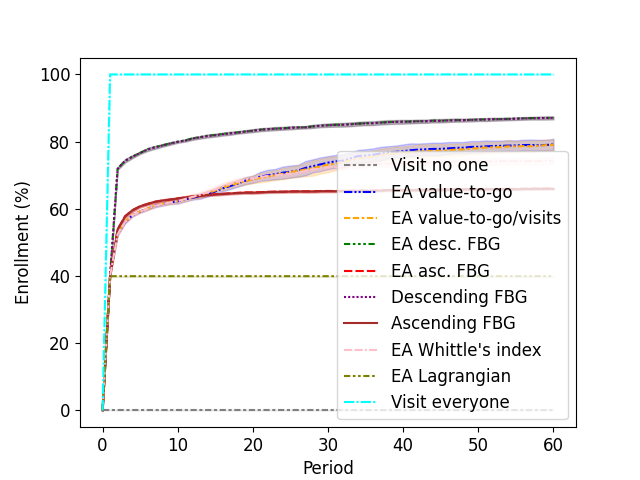}
        \caption{Scenario 1 \label{fig:scenario1-enrollment-over-time}}
    \end{subfigure}
    \begin{subfigure}{0.32\textwidth}
        \centering
        \includegraphics[width=\linewidth, trim={0.3cm 0 1.55cm 0.25cm}, clip]{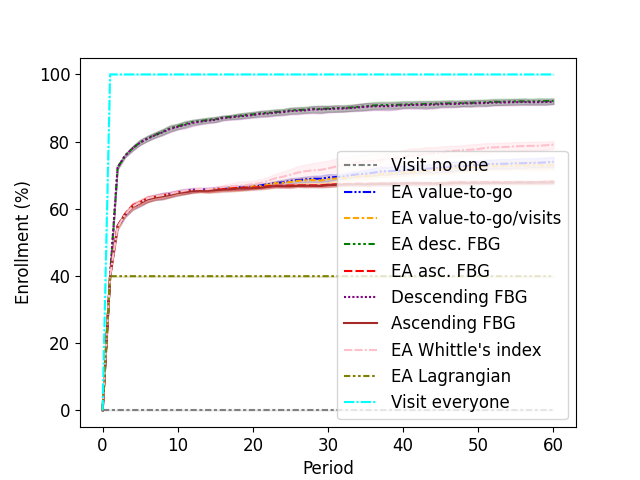}
        \caption{Scenario 2 \label{fig:scenario2-enrollment-over-time}}
    \end{subfigure}
    \begin{subfigure}{0.32\textwidth}
        \centering
        \includegraphics[width=\linewidth, trim={0.3cm 0 1.55cm 0.25cm}, clip]{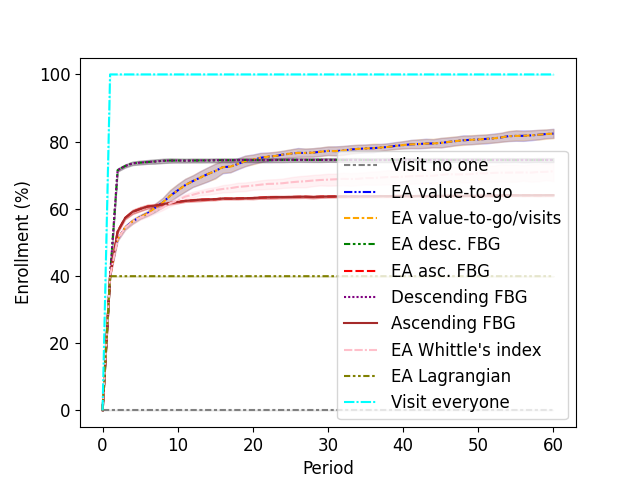}
        \caption{Scenario 3\label{fig:scenario3-enrollment-over-time}}
    \end{subfigure}
    \vspace{10pt}
    \caption{Line graphs for different scenarios showing the enrollment percentages over time at 40\% capacity with shaded 95\% confidence intervals.}
    \label{fig:enrollment_over_time_cap}
\end{figure}

Figure~\ref{fig:enrollment_over_time_cap} displays the proportion of enrolled patients for different scenarios throughout the planning horizon at 40\% capacity. As expected, enrollment increases rapidly during the first 5 periods. The best performing implementation from Scenarios 1 and 2 above (EA Lagrangian) maintains roughly 40\% enrollment across all scenarios, suggesting that there exists an enrollment ``sweet spot". Intuitively, this makes sense for two reasons. First, if we enroll too many patients, we may not have the capacity to conduct management visits causing patients to drop out (and then need to be re-screened). Second, if we do not enroll enough patients, we may not make a large enough impact and we may have excess capacity and over-visit patients, causing them to drop out.

    

\subsubsection{FBG Distribution.}
\label{sec:scen-fbg-dist}

\begin{figure}
    \centering
    \begin{subfigure}{0.32\textwidth}
        \centering
        \includegraphics[width=\linewidth, trim={0.5cm 0 0 0}, clip]{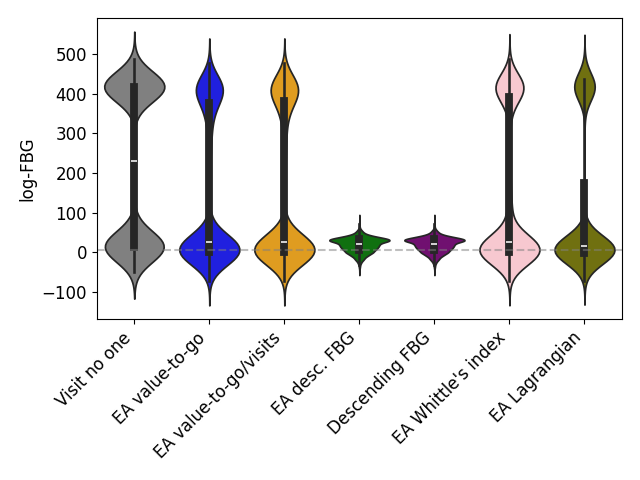}
        \caption{Scenario 1 \label{fig:scenario1-violin-cap302}}
    \end{subfigure}
    \begin{subfigure}{0.32\textwidth}
        \centering
        \includegraphics[width=\linewidth, trim={0.5cm 0 0 0}, clip]{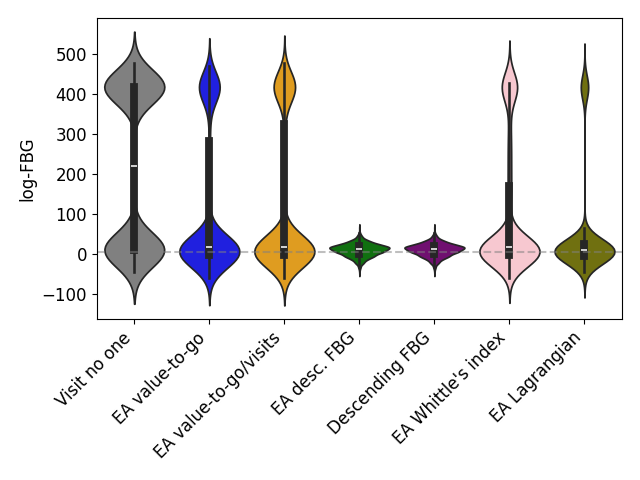}
        \caption{Scenario 2 \label{fig:scenario2-violin-cap151}}
    \end{subfigure}
    \begin{subfigure}{0.32\textwidth}
        \centering
        \includegraphics[width=\linewidth, trim={0.5cm 0 0 0}, clip]{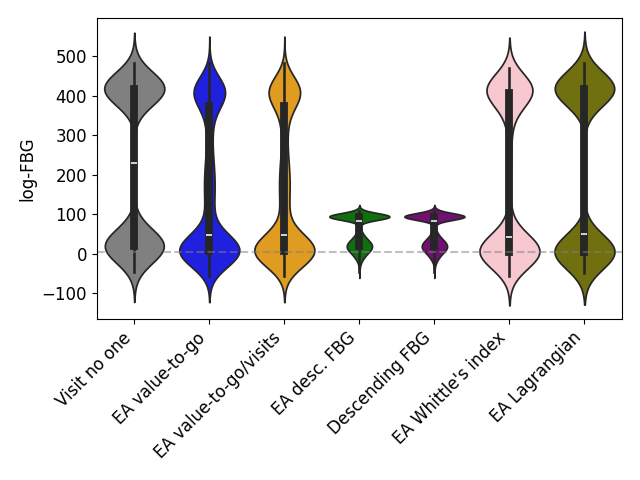}
        \caption{Scenario 3\label{fig:scenario3-violin-cap151}}
    \end{subfigure}
    \vspace{10pt}
    \caption{Violin plots showing the distribution of log-FBG values at the end of the planning horizon for different scenarios and seven heuristics at 40\% capacity. The gray dashed line indicates the threshold $\delta$.}
    \label{fig:scenarios-violin-plots}
\end{figure}

Figure~\ref{fig:scenarios-violin-plots} displays violin plots of the log-FBG values for different scenarios. For all three scenarios, descending FBG and EA descending FBG had the same median and 90th percentile performance. The median and 90th percentile were 18.9 and 19.8 for scenario 1, 10.4 and 10.9 for scenario 2, and 56.9 and 57.3 for scenario 3. Both of these policies exhibited significant improvement (more than 50\% reduction) over the visit no one policy. The best method for PPC (EA Lagrangian) did not perform well, likely because some people need to be ignored (i.e., never visited) to maximize those who are in control. By contrast, the best performing method was ranked by descending FBG, which prioritizes visiting individuals with high FBG and reducing the tail of the FBG distribution at the cost of fewer total individuals in control.

\subsection{Algorithm Performance Insights}
\label{sec:alg-perf-insights}

We discuss insights related to the top performing algorithms -- EA Lagrangian, EA Ascending FBG, and EA Descending FBG -- across all simulated patient cohorts/scenarios (see Figure~\ref{fig:scen-glycemic-control-varying-capacities}). The comparative performance of these heuristics can be understood through the interaction between system congestion, the manageability of patient conditions, and the underlying bi-level optimization structure. Our formulation models the provider's objective to maximize glycemic control, while patients decide whether to enroll based on their utility, which reflects perceived treatment benefit. This dual perspective naturally maps onto two key concepts: system congestion and condition manageability.

System congestion, captured by the ratio $\frac{|\mathcal{I}_t|}{C}$, reflects the number of patients worth visiting (from the provider's standpoint) relative to available capacity. These patients are selected to maximize downstream enrollment, and thus $I_t$ implicitly encodes the patient-side utility maximization. When congestion is high, 
feasible allocations are tightly coupled across time and resources, and the EA Lagrangian heuristic leverages dual variables as shadow prices to coordinate these interdependencies. 
It performs implicit triage, prioritizing patients or requests with higher marginal contributions to glycemic control while deprioritizing others whose inclusion would impose a disproportionate cost on feasibility. This selective focus enables efficient use of scarce capacity, though it often results in lower overall enrollment (see Figure~\ref{fig:enrollment_over_time_cap}).

However, EA Ascending FBG can also excel under high congestion when patient manageability is heterogeneous and visit effects ($\alpha$) are strong for a subset of patients. For example, in Scenario~3, where disease progression ($p$) is aggressive and enrollment effects ($\mu$) are modest, EA Ascending FBG outperformed other heuristics by concentrating visits on patients with low FBG and high $\alpha$, whose conditions were more responsive to treatment. This targeted allocation strategy proved effective in mitigating progression for the most manageable cases, even under tight capacity constraints. In this context, the provider's objective aligns with identifying patients whose glycemic control can be most effectively improved, i.e., those with high treatment manageability.

At lower congestion levels, where $\frac{|\mathcal{I}_t|}{C}$ is low and capacity constraints are less binding, the EA Descending FBG heuristic performs comparatively better. Its localized, greedy search is able to exploit residual flexibility without requiring strong dual coordination. The contrast among these methods also highlights a tension between fairness and optimization: the EA Lagrangian and EA Ascending FBG heuristics achieve higher systemic performance by concentrating resources on patients with the greatest marginal benefit, whereas the EA Descending FBG variant distributes appointments more evenly, promoting broader at the expense of aggregate efficiency. In this sense, EA Lagrangian and EA Ascending FBG reflect optimization-driven notions of fairness, while EA Descending FBG embodies a more egalitarian allocation pattern.

\subsection{Experiments with Imperfect Information}
\label{sec:exp_w_imperfect_info}
The purpose of these experiments is to evaluate the performance of the EA algorithm in a setting without perfect information. In our setting, this means that the EA algorithm does not have access to the true patient parameters and estimates them based only on observed FBG measurements (i.e., when a patient is visited). In practice, the enrollment information of each patient would be available to the CHW intervention operator, since this information can be obtained through the mobile app component of the intervention.

\subsubsection{Setup.}

We modify the experimental setup used above because of the increased complexity of solving the MLE problem for each patient-period. We sample 10, 25, and 50 patients without replacement from the set of NanoHealth patients. For each patient, we obtain MLE estimated parameters fit using their entire history. We use these \emph{true MLE parameters} to simulate 7 days of patient enrollment and FBG levels, using a random visit policy (50\% chance of a visit each period). We seed our experiment using this seven day visit history, but we do not provide the FBG values for periods where a visit did not occur.

We use a time horizon of 30 periods. For each period, we first fit the MLE for each patient using their available history. We then solve the multi-patient problem for one period using one of five approaches: visit no one, visit everyone, EA descending FBG, EA value-to-go, and EA Whittle's index. From the solution, we obtain the visit decision for the next period and update the states according to the true MLE parameters. The patient history that is used to fit the MLE parameter obtains an update of the FBG only if a visit occurs. We omit the use of the Lagrangian relaxation for these experiments, since with frequent estimation of patient parameters, the relaxation problem would have to be resolved multiple times to obtain new Lagrangian multipliers which is computationally expensive. In practice, this may be less of a concern, since after managing a patient population for several periods, the provider could essentially solve a full information problem with the MLE estimates and could compute the Lagrangian less frequently.

For 10 patients, we consider capacity levels of 2, 4, 6, 8 visits per day, for 25 patients, we consider capacity levels of 5, 10, 15, 20 visits per day, and for 50 patients, we consider 10, 20, 30, 40 visits per day. We repeat each experiment 5 times.

\subsubsection{Results.}

\begin{figure}[ht]
\centering
\includegraphics[width=0.9\textwidth]{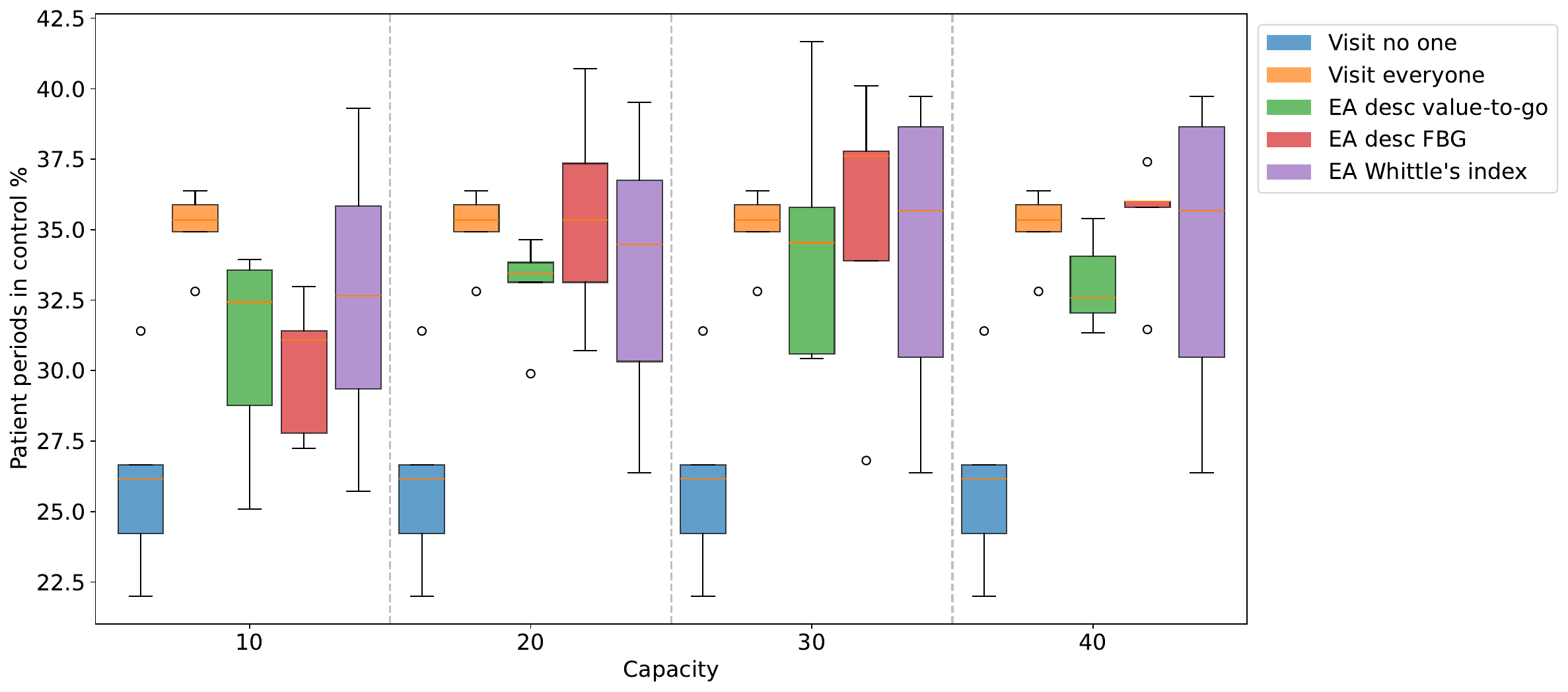}
\caption{Boxplots of the average number of patients in control across all periods for each repetition. \label{fig:CE_150}} 
\end{figure}

We display the results for 50 patients and refer the reader to Section~\ref{EC:imperfinfo} for the results with 10 and 25 patients. Figure~\ref{fig:CE_150} displays boxplots of the average number of patients in control across all periods for each repetition. The visit everyone and visit no one policies had average PPC values of 26.1\% and 35.1\%, respectively. For a 20\% capacity level (10 out of 50), the EA value-to-go, EA descending FBG, and EA Whittle's index had average PPC values of 30.8\%, 30.1\%, and 32.6\%, respectively. The EA descending FBG policy improved upon visit no one by 24.9\% and comprised 92.9\% of the visit everyone policy with 80\% fewer resources.

\begin{figure}[ht]
\centering
\includegraphics[width=0.9\textwidth]{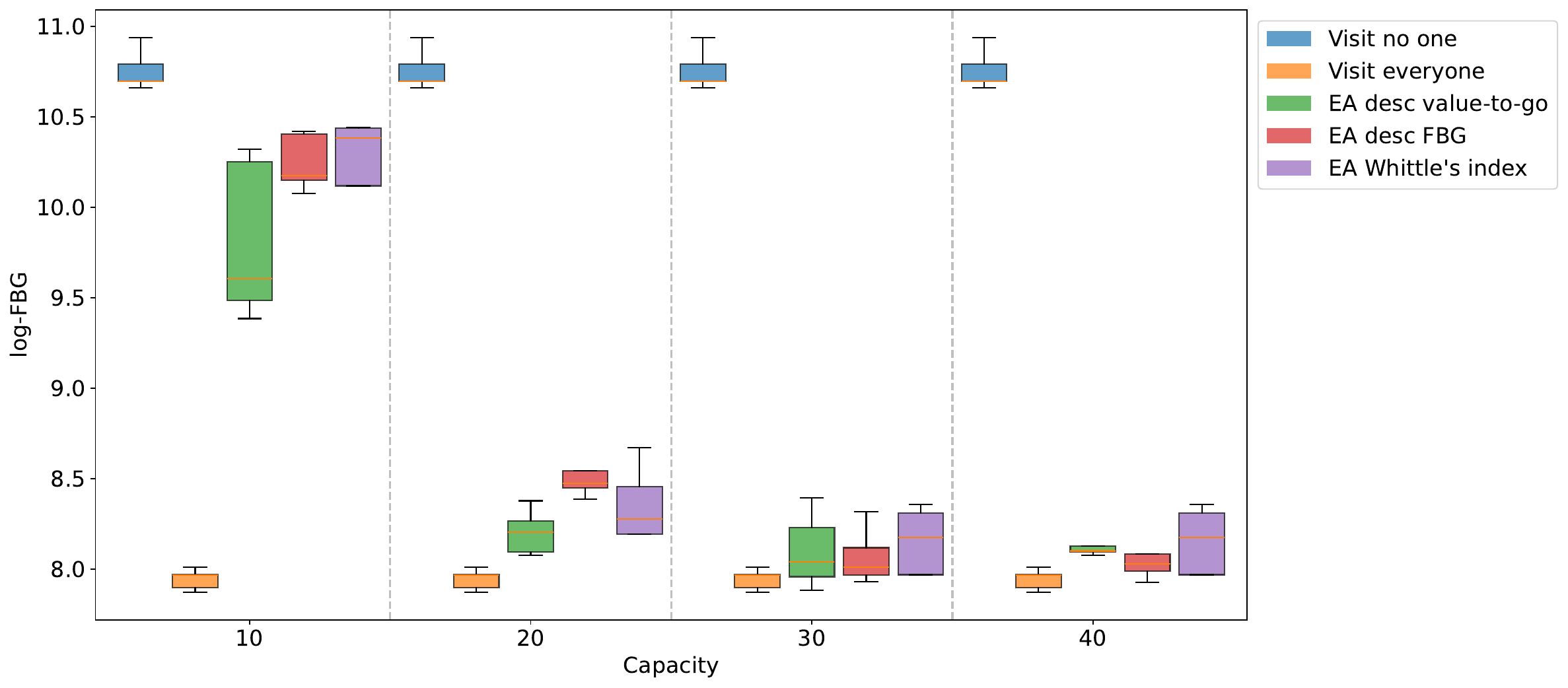}
\caption{Boxplots of the average log-FBG across all periods for each repetition. \label{fig:CE_FBG_150}} 
\end{figure}

Figure~\ref{fig:CE_FBG_150} displays boxplots of the average log-FBG across all periods for each repetition. We display log-FBG values because the actual FBG values are very large and patients would likely experience an acute event before reaching those levels. The visit everyone and visit no one policies had average log-FBG values of 10.8 and 7.9, respectively. For a 20\% capacity level (10 out of 50), the EA value-to-go, EA descending FBG, and EA Whittle's index had average FBG values of 9.8, 10.3, and 10.1, respectively. The EA descending FBG improved upon visit no one by 8.8\% (on log scale) and captured 34.7\% of improvement from visit no one to visit everyone with 80\% fewer resources (on log scale). 

\begin{figure}[ht]
\centering
\begin{subfigure}[ht]{0.45\textwidth}
\centering
\includegraphics[width=\textwidth]{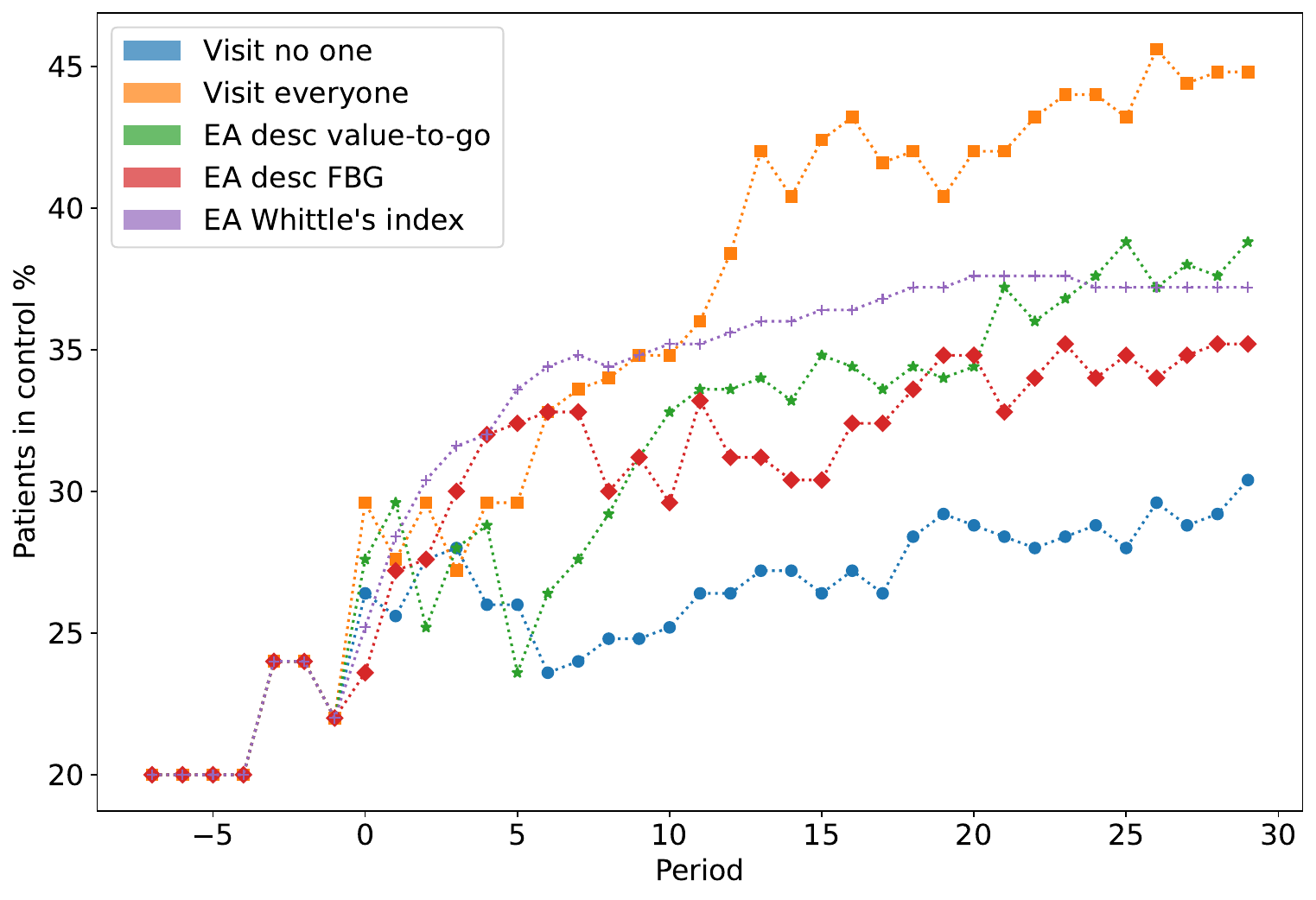}
\caption{\label{fig:CE_2a50}}
\end{subfigure}
\hfill
\begin{subfigure}[ht]{0.45\textwidth}
\centering
\includegraphics[width=\textwidth]{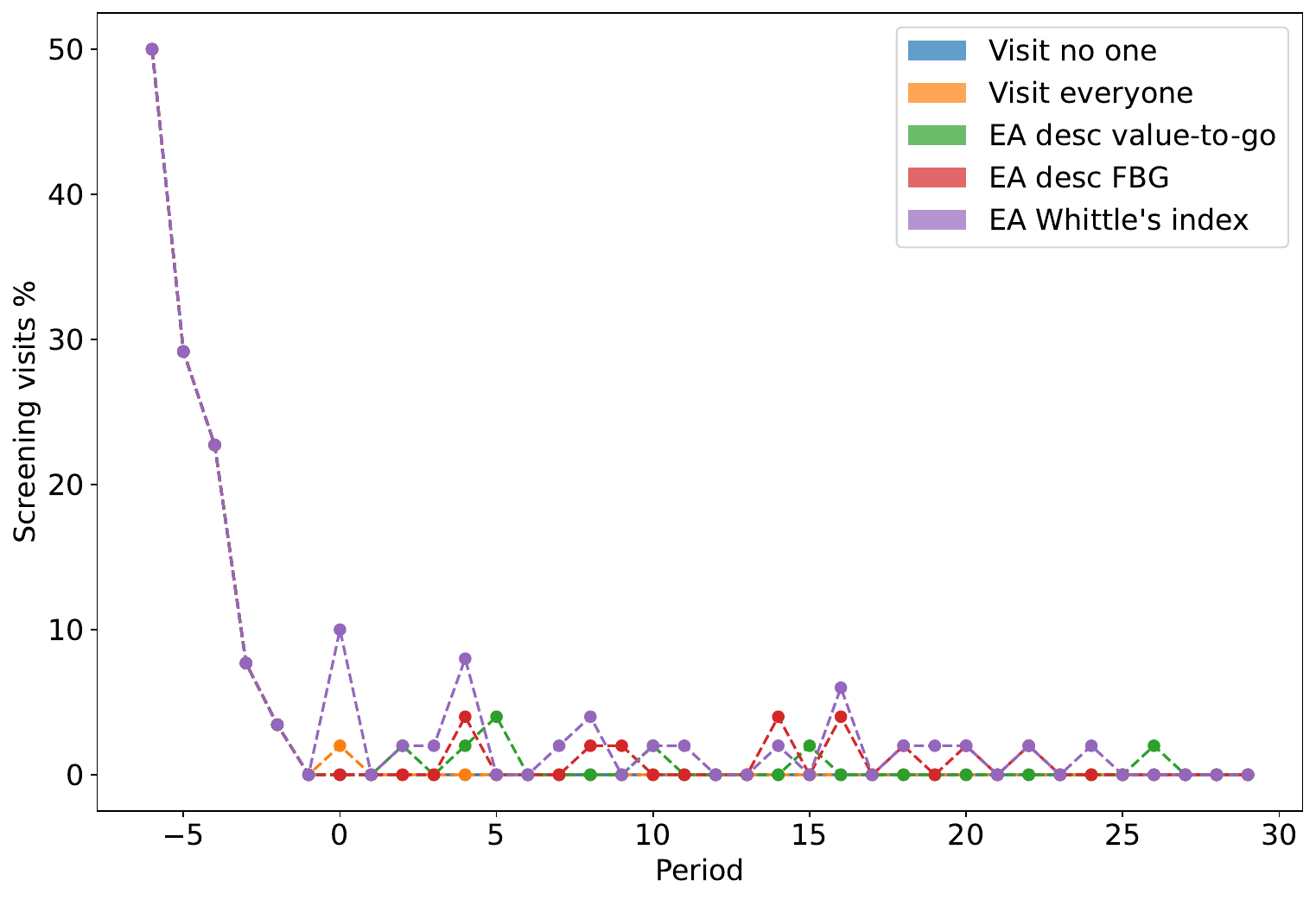} 
\caption{\label{fig:CE_2b50}}
\end{subfigure}

\begin{subfigure}[ht]{0.45\textwidth}
\centering
\includegraphics[width=\textwidth]{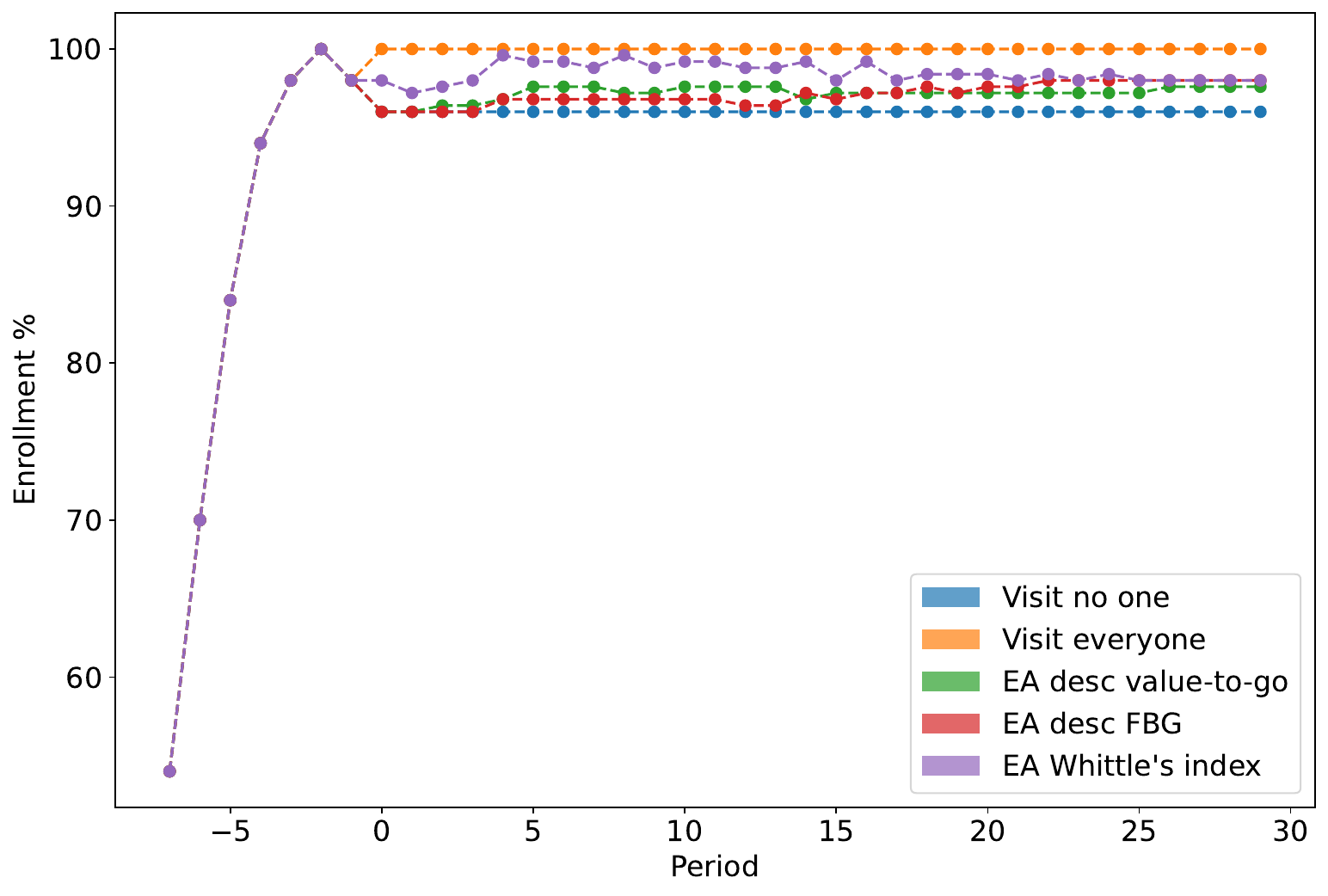} 
\caption{\label{fig:CE_2c50}}
\end{subfigure}\vspace{10pt}
\caption{(a) Line plots of the average patient periods in control as a function of the period, (b) Line plots of the average number of screening visits as a function of the period, and (c) Line plots of the average number of enrolled patients as a function of the period.\label{fig:CE_250}}
\end{figure}

Figure~\ref{fig:CE_250} displays line plots for PPC, enrollment, and visit composition as a function of period for a 20\% capacity level (10 out of 50). Figure~\ref{fig:CE_2a50} displays the proportion of patients in control each period for each policy. The x-axis displays the burn in period as negative periods; note that each policy is the same. There is a clear separation between visit no one and visit everyone. All EA based methods improve upon visit no one and perform similarly over time. Figure~\ref{fig:CE_2b50} displays the proportion of visits each period that were screening visits for each policy. Figure~\ref{fig:CE_2c50} displays the proportion of enrolled patients each period for each policy. The visit no one policy maintains an average enrollment of 96\%, while visit everyone maintains 100\% enrollment. All algorithms maintain or oscillate between 97\% and 99\% enrollment, which can also be seen by the oscillation in screening visits.

\subsection{Managerial Implications}
\label{sec:results-manag-implications}

Our work has relevant implications not only for NanoHealth's operations, but also for other CHW interventions for chronic diseases. We highlight the following key managerial implications: 

\begin{enumerate}

    \item \emph{Given the competing objectives of equity and efficiency, there is no `one-size-fits-all' solution to plan CHW interventions.} Through our analysis, we found that cohort composition, capacity level, and intervention goals all affect the choice of the best CHW visit plan. We presented results where simple benchmark heuristics had a comparable performance to the EA implementations (e.g., EA descending FBG and descending FBG performed equally for median and 90th percentile FBG for all three scenarios) and results where using the best benchmark heuristic requires 50\% greater capacity to achieve the same performance as compared to the best EA implementation (e.g. PPC values for Scenario 1, Scenario 2, and the NanoHealth cohort). Across all scenarios the best two methods were the EA Lagrangian (best for PPC in Scenario 1 and 2) and EA Descending FBG (best for PPC in NanoHealth cohort and best for FBG across all three scenarios and the nanohealth cohort). These methods use different approaches and demonstrate an important managerial trade-off between maximizing PPC and reducing FBG in the population. The EA Lagrangian method enrolls a lower percentage of patients as compared to the EA descending FBG method, which results in some patients being prioritized while others are ignored. The result is more patients in control at the cost of worse overall FBG metrics. On the other hand, the EA descending FBG method prioritizes individuals with high FBG thereby reducing the entire FBG distribution by not ignoring any patients at a cost of fewer patients in control. This trade-off highlights one of our key contributions: \emph{the development of a flexible framework that can be tailored by providers to meet their priorities, patient needs, and resource availability.}
    
    \item \emph{Incorporating patient motivational and health states is key to efficient resource use in CHW interventions, especially with heterogeneous patient cohorts.} We observe that the benchmark heuristics, that do not use any estimates of the motivational states, spend a significant portion of their CHW capacity on screening new patients rather than providing management visits to patients that are already enrolled in treatment (see Section~\ref{sec:nanohealth-visit-types}). This results in patients frequently being screened and then immediately dropping from the program, effectively wasting visit resources. On the other hand, EA is able to better target potential patients that are likely to remain in the program for a long time, likely due to the incorporation of motivational states and adverse factors. Moreover, the EA personalizes visit frequency for enrolled patients, avoiding visiting patients too often or not often enough, which can both lead to drop outs and poor glycemic control. At the cohort level, this leads the EA implementations to find an enrollment ``sweet spot" rather than simply growing the size of the program without consideration to presently enrolled patients, leading to large programs where many patients are ignored in favor of screening patients who may not benefit at all from enrollment.

    \item \emph{The performance of our approach is robust to settings with imperfect information.} Specifically, we evaluated a situation where the EA algorithm does not have access to the true patient parameters and estimates them based only on observed FBG measurements (i.e., when a patient is visited). These experiments more closely mimic reality and provide insight on how the model may perform without perfect information. We also demonstrate the feasibility of using machine learning to forecast patient-level parameters.
    
    \item \emph{Our framework is scalable and can be implemented on mobile tablets already used by CHW programs.} Many CHW programs currently have the infrastructure to implement the EA because they use mobile tablets for data collection and communication. Incorporating the EA approaches would also require using tablets for task list implementation. The EA relies on data typically collected by CHW programs during pilot phases or normal day-to-day operations. Moreover, our technical methods enable providers to identify patient types and to determine how to best personalize their intervention as opposed to providing a one size fits all treatment, all while maintaining a solution method that is computationally tractable.
\end{enumerate}

\subsection{Limitations and Extensions} 
\label{sed:limit_and_etend}
While we believe that we have presented a comprehensive modeling approach for the CHW scheduling problem, there are several limitations of our approach that must be acknowledged. First, while we chose to use a mechanistic system model for patient behavior, alternative behavioral models such as structural estimation  \citep{rust1994structural} or discrete choice models \citep{bierlaire1998discrete} could also potentially capture similar patient behavior. That said, our model was designed with the goal of being incorporated in an optimization framework, and performs well in predicting future enrollment behavior (see Appendix~{\ref{EC:incontrolprediction}}). Moreover, our model parameters comport with current understanding of behavior change in accordance with PBT \citep{rich2015theory}.

In terms of estimation, we note that there may be some bias in our dataset due to patient observations being censored by the CHW program capacities. However, we do not believe this is a major concern since currently, when CHW operations are constrained, program operators essentially allocate random visit schedules due to the lack of optimization tools at their disposal. However, this could be a concern once an optimization tool is implemented since it could mean that the patients are not sufficiently explored, leading to a violation of our sufficient excitation condition. This can be mitigated by initiating the intervention with a run-in period where visits are administered randomly (though within the capacity) to patients, and then using an adaptive estimation approach. Our results in Section~\ref{sec:exp_w_imperfect_info} reflect that this is an effective approach in such cases. A related concern is that our model assumes that screening and management visits require the same amount of effort from a CHW within a given period. While this may not strictly be the case, due to the limitations of the program, both visits would consume a sufficient amount of a CHW's time within a decision epoch that they can be treated as equivalent effort for the purposes of planning.

Additionally, our analysis and methods as presented assume an unchanging patient panel. While this assumption may hold for short timeframes, over the long run patient populations may change. First, it is important to note that the notion of set $\mathcal{I}_t$ is only assessed on a criteria based on the single-patient problem, and is thus not affected by the addition or removal of patients in the current period. However, methods such as the Lagrangian relaxation and Whittle's index, which aim to capture future resource utilization, may be affected by these changes. An approach from the queuing literature that has been proposed to address these problems is to explicitly model the participant arrival and departure processes into the dynamics of the model \citep{verloop2016asymptotically}. In the queuing literature, this generally involves a fluid model relaxation and the assumption that there are finite types of potential arrivals. While our existing formulation is for an uncountable infinite number of patient types, a model that aims to capture these dynamics over longer horizons could simplify these to our main patient type clusters generated in Appendix~\ref{EC:clusteringprediction} and proceed with a similar analysis. In these cases, if the resulting dynamic programming problem can be shown to have a unique fixed point solution to the Bellman equation, the Lagrangian and Whittle's index would be asymptotically optimal heuristics in the size of the patient population. Alternatively, our existing approaches could be modified such that the CHW capacity is stochastic at each time period. In this new formulation, we would compute a robust policy for CHW scheduling with the different realizations of the constraint corresponding to potential future changes in the patient population.

\section{Conclusion}
\label{sec:conclusion}


In this paper, we developed a modeling framework to optimize a resource-constrained CHW intervention for diabetes care in urban areas in LMICs. 
Our framework explicitly models the tradeoff between screening new patients and providing management visits to individuals who are already enrolled in treatment. We account for patients' motivational states, which affect their decisions to enroll or drop out of treatment and, therefore, the effectiveness of the intervention. We incorporate these decisions by modeling patients as utility-maximizing agents within a bi-level provider problem that we solve using approximate dynamic programming. Our scalable heuristics rely on theoretical results from both the single- and multi-patient problems. By performing several simulation experiments, we that our approach can maintain similar performance (in terms of patient periods in control) as compared to baseline methods with up to 50\% less capacity. For the NanoHealth cohort, our approach is able to reduce fasting blood glucose by up to 25\% with the same capacity as baseline methods. Finally, we conduct a set of experiments to demonstrate that the performance of our approach does not degrade with imperfect information.



\ACKNOWLEDGMENT{We thank NanoHealth for generously providing the operational data that made this research possible. We are particularly grateful to Manish Ranjan for his assistance in understanding the data and the organizational context of the program. Authors would also like to thank the support of an NLM training grant to the Computation and Informatics in Biology and Medicine Training Program (NLM 5T15LM007359), the University of Wisconsin-Madison Global Health Institute, who partially funded this project through a Seed Grant, and the American Family Funding Initiative for funding this research.}

\bibliographystyle{informs2014} 
\bibliography{bibliography.bib} 

\noindent\rule{\linewidth}{0.4pt}

\AUTHORBIO{Katherine Adams is a Postdoctoral Fellow in the Department of Operations and Analytics at the University of Texas at San Antonio. Her research leverages optimization and predictive analytics to enhance healthcare operations, with a dual focus on strategic workforce planning and operational efficiency. Her work spans applications in global health systems, community-based care delivery, and clinical settings. Her contributions to the field earned her the Bonder Scholarship for Applied Operations Research in Health Services as a PhD student.}

\AUTHORBIO{Justin Boutilier is an Assistant Professor at the Telfer School of Management within the University of Ottawa and a Research Chair in AI and Mental Health at the Institut du Savoir Montfort. His research focuses on developing and applying actionable analytics to solve health and humanitarian problems, primarily focusing on the interface between predictive and prescriptive analytics. His work has been recognized by the Institute for Operations Research and the Management Sciences (INFORMS); he has won the Pierskalla Best Paper Award, was a finalist for the Doing Good with Good OR prize, and received the Bonder Scholarship as a PhD student.}

\AUTHORBIO{Yonatan Mintz is an Assistant Professor at the Department of Industrial and Systems Engineering at the University of Wisconsin-Madison. His work sits at the intersection of machine learning, automated decision-making, and human-sensitive contexts, with a particular focus on making AI systems more equitable and effective in high-stakes fields like healthcare. His technical expertise covers reinforcement learning, stochastic control, non-convex optimization, and machine learning theory.}

\AUTHORBIO{Sarang Deo is a Professor of Operations Management at the Indian School of Business. His research focuses on understanding the impact of operational interventions on population health outcomes with emphasis on low- and middle-income countries. His current areas of work include antimicrobial resistance, developing surveillance systems, optimal deployment of frontline workers, and design and evaluation of digital health technologies including AI. Sarang works closely with global health funding and implementation agencies like the Gates Foundation, Clinton Health Access Initiative and PATH. He is currently on the WHO Strategic and Technical Advisory Group on Tuberculosis and serves on several state and central government committees on healthcare in India.}


\ECSwitch
\ECHead{Electronic Companion}

\section{Proofs of Propositions in the Text}\label{app:proofs}

\proof{Proof of Proposition \ref{prop:benefit}:}
Unless otherwise stated, all inequalities and equations hold with probability one. We can write the benefit function for each component of the utility function separately. To write this function for $B^b_{i,t}$, we first substitute the dynamic equations for $b_{i,t+1},s_{i,t+1}$ into the utility function components to obtain the following explicit representation in terms of the patient states at time $t$:
\begin{align*}
    U_b(b_{i,t}, z_{i,t}) &= -(b_{i,t} + p_i - \mu_i z_{i,t} - \alpha_i y_{i,t} z_{i,t} + \xi_{i,t}) \\
    U_s(s_{i,t}, \theta_{i,t}, z_{i,t}) &= -\theta_{i,t}(z_{i,t}(\gamma(s_{i,t}-s_{i,0}) + s_{i,0}) + \beta_i y_{i,t} z_{i,t})
\end{align*}

Let $B_{i,t} = B^b_{i,t} + B^s_{i,t}$, where the first component corresponds to the net benefit associated with $b_{i,t}$ and second component corresponds to the net benefit associated with $s_{i,t}$.  Then through a similar definition of $B_{i,t}$ by doing the appropriate subtractions we can derive the following forms.
%
\begin{align*}
    B^b_{i,t} &= \mu_i + \alpha_i y_{i,t} \\
    B^s_{i,t} &=
    -\theta_{i,t} (\gamma( s_{i,t} -  s_{i,0}) + s_{i,0}) - \theta_{i,t} \beta_i y_{i,t}
\end{align*}
Combining these expressions gives the first result of the proposition. 

To obtain the next result note that if $B_{i,t} \geq 0$ this implies that $U_i(b_{i,t+1}, s_{i,t+1}, \theta_{i,t}, \xi_{i,t}, 1) \geq U_i(b_{i,t+1}, s_{i,t+1}, \theta_{i,t}, \xi_{i,t}, 0)$ and thus the utility function is maximized when $z_{i,t} = 1$. Likewise, when $B_{i,t} < 0$ this means  $U_i(b_{i,t+1}, s_{i,t+1}, \theta_{i,t}, \xi_{i,t}, 1) < U_i(b_{i,t+1}, s_{i,t+1}, \theta_{i,t}, \xi_{i,t}, 0)$ and therefore $z_{i,t} = 0$. However, $z_{i,t}$ can only be equal to 1 if $y_{i,t}$ or $z_{i,t-1}$ are equal to 1. This gives the second result of the proposition.
%
\halmos
\endproof

\proof{Proof of Proposition \ref{prop:mle_form}:} 
We seek to use a MLE approach to estimate all unknown parameters:
\begin{equation}
  ( \hat{\bb},\hat{\bs},\hat{\btheta},\hat{\beta}, \hat{\alpha}, \hat{\mu}, \hat{p}, \hat{\lambda}, \hat{\gamma}, \hat{\rho}) \in \argmin_{\bb,\bs,\btheta,\beta, \alpha, \mu, p, \lambda, \gamma, \rho}  p(\{\bar{b}_t\}_{t\in K}|\bb,\bs,\btheta,\beta, \alpha, \mu, p, \lambda, \gamma, \rho, \bz, \by, \bxi) \label{eq:mle_packed}
\end{equation}

Using the dynamic equations described in Section~\ref{sec:patient_model}, we can decompose the likelihood function in \eqref{eq:mle_packed} as follows:
\begin{multline}
    p(\{\tilde{b}_t\}_{t\in K}|\bb,\bs,\btheta,\beta, \alpha, \mu, p, \lambda, \gamma, \rho, \bz, \by, \bxi) = \prod_{t\in K} p(\tilde{b}_t|b_t) \prod_{t\in T} p(b_t| b_{t-1}, p, \mu, z_t, \alpha, y_t,\xi_t) \\ p(s_t|s_{t-1}, s_0, \beta, y_t, z_t, \gamma)  p(\theta_t|\theta_{t-1},\theta_0, \lambda, y_t,z_t, \rho)p(z_t|\mu,\theta_t,s_t,s_0,\alpha,\beta,\gamma,y_t)p(\xi_t) \label{eq:likelihood}
\end{multline}
Note that while the product terms related to $\tilde{b}_t$ are only taken over time periods $\mathcal{K}\subset \mathcal{T}$,  the other terms are taken over all time indices. This is due to the definition of the one step dynamics, and allows us to estimate the values of these parameters during periods for which we do not have direct observations. 

Next, we take the log of both sides of the likelihood function in Equation \eqref{eq:likelihood} to obtain the log-likelihood:
\begin{align*}
     \log p(\{\tilde{b}_t\}_{t\in \mathcal{K}}|\bb,\bs,\btheta,\beta, \alpha, \mu, p, \bz, \by, \bxi) = & \sum_{t\in \mathcal{K}} \log p(\tilde{b}_t|b_t) + \sum_{t\in \mathcal{T}} \log p(b_t| b_{t-1}, p, \mu, z_t, \alpha, y_t, \xi_t) \\ &+ \log p(s_t|s_{t-1}, s_0, \beta, y_t, z_t,\gamma) + \log p(\theta_t|\theta_{t-1},\theta_0, \lambda, y_t,z_t,\rho) \\ &+ \log p(z_t|\mu,\theta_t,s_t,s_0,\alpha,\gamma,\beta,y_t) + \log p(\xi_t). \label{eq:log_likelihood}
\end{align*}
Observe that $p(\tilde{b}_t|b_t) = f_\epsilon (\tilde{b}_{t} - b_t)$  and that $p(\xi_t) = f_{\xi}(\xi_t)$ by definition of the disturbance model and note that because the remaining terms correspond to deterministic system dynamics, they are degenerate distributions. Thus, the MLE can be expressed as the following constrained optimization problem:
\begin{subequations}
\begin{align}
    \underset{\bb,\bs,\btheta,\beta, \alpha, \mu, p, \lambda, \rho, \gamma,\bxi}{\mathrm{minimize}} \: & \sum_{t \in \mathcal{K}} \log f_\epsilon(\tilde{b}_t - b_t) + \sum_{t\in \mathcal{T}} \log f_\xi(\xi_t) \\
    \mathrm{subject\,to}\quad &  b_{t+1} = b_t + p - \mu z_{t} -\alpha y_{t} z_{t} + \xi_t, \quad  \forall  t \in \mathcal{T},  \label{FBGup}\\ 
    & s_{t+1} = \gamma z_{t} (s_t - s_0) + z_t s_0 + \beta y_{t} z_{t}, \quad \forall t \in \mathcal{T}, \\
    & \theta_{t+1} = \rho (\theta_{t} - \theta_{0}) + \theta_{0} - \lambda y_{t} z_{t}, \quad \forall t \in \mathcal{T},\\
   & z_{i,t} = (z_{i,t-1} + y_{i,t} - z_{i,t-1}y_{i,t})\mathbbm{1}[B_{i,t} \geq 0] \quad \forall t \in \mathcal{T}, \label{eq:z_const_mle} \\
    & \hat{b}_{t},s_{t},\theta_t \geq 0, \quad \forall t \in \mathcal{T}, \\
    & \beta, \alpha, \mu, p, \lambda, \rho, \gamma \geq 0.
\end{align}
\end{subequations}
Constraint \eqref{eq:z_const_mle} is derived from Proposition \ref{prop:benefit} and the remaining constraints come from the model dynamics. To obtain the final formulation, constraint \eqref{eq:z_const_mle} is reformulated using disjunctive constraints \citep{wolsey1999integer}.
\halmos
\endproof

\proof{Proof of Lemma~\ref{lemma-1}:} Since $z_{t-1}\in\{0,1\}$, we simply evaluate $z_{t}$ for $z_{t-1}=0$ and $z_{t-1}=1$:
$$z_{t}(0,y_{t}) = y_{t} \cdot \mathbbm{1}(B_{t}(y_{t}) \geq 0) \leq \mathbbm{1}(B_{t}(y_{t}) \geq 0) = z_{t}(1,y_{t}).\halmos$$
\endproof

\proof{Proof of Lemma~\ref{lemma-2}:}
Since $V_N$ is only a function of the FBG value, $b_N$, and $b_N=b_{N-1}+p-\mu z_{N-1}-\alpha y_{N-1} z_{N-1} + \xi_{N-1}$, we have that for any value of $\xi_{N-1}$:
\begin{align*}
    V_N(b_N,s_N,\theta_N,z_{N-1})=
    \begin{cases}
        1, & \text{if } b_{N-1} \leq \delta - p + \mu z_{N-1}(y_{N-1}) + \alpha y_{N-1} z_{N-1}(y_{N-1}) - \xi_{N-1}, \\
        0, & \text{if } b_{N-1} > \delta - p + \mu z_{N-1}(y_{N-1}) + \alpha y_{N-1} z_{N-1}(y_{N-1}) - \xi_{N-1}.
    \end{cases}
\end{align*}
Since $z_{N-1}$ is a binary variable, we evaluate $V_N$ for $z_{N-1}=0$ and $z_{N-1}=1$:
\begin{align*}
   V_N(b_N,s_N,\theta_N,0)  &=
    \begin{cases}
        1, & \text{if } b_{N-1} \leq \delta-p - \xi_{N-1}\\
        0, & \text{if } \delta-p  - \xi_{N-1}< b_{N-1} \leq \delta-p+\mu+\alpha y_{N-1} - \xi_{N-1}\\
        0, & \text{if } b_{N-1} > \delta-p+\mu+\alpha y_{N-1} - \xi_{N-1}
    \end{cases}\\
    V_N(b_N,s_N,\theta_N,1) &=
    \begin{cases}
        1, & \text{if } b_{N-1} \leq \delta-p - \xi_{N-1}\\
        1, & \text{if } \delta-p - \xi_{N-1}< b_{N-1} \leq \delta-p+\mu+\alpha y_{N-1} - \xi_{N-1}\\
        0, & \text{if } b_{N-1} > \delta-p+\mu+\alpha y_{N-1} - \xi_{N-1}
    \end{cases}
\end{align*}
Note that $\delta - p - \xi_{N-1} \leq \delta - p + \mu + \alpha y_{N-1} - \xi_{N-1}$ because $\mu \geq 0$ and $\alpha \geq 0$. Since $V_N(b_N,s_N,\theta_N,0) \leq V_N(b_N,s_N,\theta_N,1)$ for all values of $b_{N-1}$ and $\xi_{N-1}$, the proof is complete. \halmos
\endproof

\proof{Proof of Theorem~\ref{thm-1}:} 

We proceed by induction. 
For the base case, we need to show that $V_{N-1}$ is nondecreasing in $z_{N-2}$, i.e., $V_{N-1}(b_{N-1},s_{N-1},\theta_{N-1},0) \leq V_{N-1}(b_{N-1},s_{N-1},\theta_{N-1},1)$ for any value of $\xi_{N-2}$, which is equivalent to:
\begin{equation}
\max_{y_{N-1}\in\{0,1\}} \{V_N(z_{N-2}=0)\} \leq \max_{y_{N-1}\in\{0,1\}} \{V_N(z_{N-2}=1)\}.
\end{equation}
Here we use the notation that $V_N(z_{N-2}=z) = V_N(b_N(z),s_N(z), \theta_N(z), z_{N-1}(z))$, where $(b_N(z),s_N(z), \theta_N(z), z_{N-1}(z)) $ correspond to the states in period $N$ given the states in period $N-1$ if the patient had an enrollment status of $z$.  Observe that $V_N$ does not depend directly $z_{N-2}$, but is only affected by it through $z_{N-1}$. 
%
%
Since $V_N$ is nondecreasing in $z_{N-1}$ for all values of $\xi_{N-1}$ (Lemma~\ref{lemma-2}) and $z_{N-1}$ is nondecreasing in $z_{N-2}$ (Lemma~\ref{lemma-1}), it follows
that $V_N$ is nondecreasing in $z_{N-2}$ since the composition of nondecreasing functions is also nondecreasing for all values of $\xi_{N-1}$. Moreover, since the dependence between $z_{N-1}$ and $z_{N-2}$ is only through $B_{N-1}$ this relationship is functionally independent of $\xi_{N-2}$ meaning the above composition holds for any realization of $\xi_{N-2}$. 


For the inductive hypothesis, using our new notational convention, we assume that $V_{t+1}$ is nondecreasing in $z_t$, that is, $V_{t+1}(z_t=0)\leq V_{t+1}(z_t=1)$ with probability 1.
Finally, we show that $V_t$ is nondecreasing in $z_{t-1}$, which follows the same logic as the base case. First, observe that $V_t(z_{t-1}=0) \leq V_t(z_{t-1}=1)$ is equivalent to
$$\max_{y_{t}\in\{0,1\}} \mathbb{E}_\xi \, \{V_{t+1}(z_{t-1}=0)\} \leq \max_{y_{t}\in\{0,1\}} \mathbb{E}_\xi \,\{V_{t+1}(z_{t-1}=1)\}.$$
Similar to the base case, $V_{t+1}$ is a composition function of $z_{t-1}$ through $z_t$, which is a function of $z_{t-1}$. Since $V_{t+1}$ is nondecreasing in $z_{t}$ (inductive hypothesis), $z_{t}$ is nondecreasing in $z_{t-1}$ (Lemma~\ref{lemma-1}), and that expectations over nondecreasing functions preserve their monotonicity, it follows
that $\mathbb{E}_\xi V_{t+1}$ is nondecreasing in $z_{t-1}$. And, since the maximum of monotonically nondecreasing functions is a monotonically nondecreasing function, we have that $V_{t}$ is nondecreasing in $z_{t-1}$. \halmos
\endproof

\proof{Proof of Theorem~\ref{thm:suff_nec_cond}:}

It is sufficient to show that $\mathbb{E}_\xi V_t(b_t,s_t,\theta_t,z_{t-1})$ is strictly increasing in $y_{t-1}$ for case $(i)$, $\mathbb{E}_\xi V_t(b_t,s_t,\theta_t,z_{t-1})$ is strictly decreasing in $y_{t-1}$ for case $(ii)$, and $\mathbb{E}_\xi V_t(b_t,s_t,\theta_t,z_{t-1})$ is constant in $y_{t-1}$ for case $(iii)$. Since we have already shown in Theorem~\ref{thm-1} that $\mathbb{E}_\xi V_t$ is nondecreasing in $z_{t-1}$,
we only need to show when $z_{t}(z_{t-1}, y_{t}) = (z_{t-1}+y_{t}-z_{t-1}y_{t}) \cdot \mathbbm{1}(B_{t}(y_{t}) \geq 0)$ is strictly increasing, strictly decreasing, and constant in $y_t$ (which would imply the same relationship between $z_{t-1}$ and $y_{t-1}$). We proceed by enumerating two cases corresponding to $z_{t-1}=0$ and $z_{t-1}=1$. 

For the case when $z_{t-1}=1$ (i.e., the patient is enrolled), $z_t(1,0) > z_t(1,1)$ (i.e., strictly decreasing) if and only if $B_{t}(0) \geq 0$ and $B_{t}(1) < 0$, $z_t(1,0) < z_t(1,1)$ if and only if $B_{t}(0) < 0$ and $B_{t}(1) \geq 0$, and $z_t(1,0) = z_t(1,1)$ if and only if $B_{t}(0) <0$ and $B_{t}(1) < 0$ or $B_{t}(0) \geq 0$ and $B_{t}(1)\geq 0$.

For the case when $z_{t-1}=0$ (i.e., the patient is not enrolled), $z_t(0,0)$ is never strictly greater than $z_t(0,1)$ because $z_t(0,0) = 0$, $z_t(0,0) < z_t(0,1)$ if and only if $B_{t}(1) \geq 0$, and $z_t(0,0) = z_t(0,1)$ if and only if $B_{t}(1) < 0$. \halmos

\endproof

\proof{Proof of Lemma~\ref{lemma-6}:}
There are two key results we need to establish to prove this lemma: first that the value function at time $N-1$ is additively decomposable, and second that the only patients that should be visited are those in the set $\mathcal{I}_{N-1}$. 

We will begin by showing the first result about additive decomposability. Define the helper function $J_{N-1}: \mathbb{R}_+^{|\mathcal{P}| \times 3} \times \{0,1\}^{|\mathcal{P}|} \mapsto \mathbb{R}$ as the value function of the the following optimization problem:
\begin{equation}
    J_{N-1}(\bb_{N-1},\bs_{N-1},\btheta_{N-1},\bz_{N-2}) = \max_{\by_{N-1} \in \{0,1\}^{|\mathcal{P}|}, \by_{N-1}^\top \mathbbm{1} \leq C }  \mathbb{E}_\xi \mathcal{V}_{N} (h(\bb_{N-1},\bs_{N-1},\btheta_{N-1},\bz_{N-2},\by_{N-1})) \label{eq:helper_jnmin}
\end{equation}

Recall from \eqref{eq:joint_dp_N} that $\mathcal{V}_N(\bb_N,\bs_N,\btheta_N,\bz_{N-1}) = \sum_{i\in \mathcal{P}} \mathbbm{1}(b_{i,N} \leq \delta)$. Substituting into \eqref{eq:helper_jnmin} allows us to perform the following reformulation:
\begin{align}
& \eqref{eq:helper_jnmin} =  \max_{\by_{N-1} \in \{0,1\}^{|\mathcal{P}|}, \by_{N-1}^\top \mathbbm{1} \leq C } \sum_{i\in \mathcal{P}} \mathbb{E}_\xi \mathbbm{1}(b_{i,N}(\bb_{N-1},\bs_{N-1},\btheta_{N-1},\bz_{N-2},\by_{N-1}) \leq \delta)\\
& =\max_{\by_{N-1} \in \{0,1\}^{|\mathcal{P}|}, \by_{N-1}^\top \mathbbm{1} \leq C } \sum_{i\in \mathcal{P}} \mathbb{P}_\xi(b_{i,N}(\bb_{N-1},\bs_{N-1},\btheta_{N-1},\bz_{N-2},\by_{N-1}) \leq \delta) \\
& =\max_{\by_{N-1} \in \{0,1\}^{|\mathcal{P}|}, \by_{N-1}^\top \mathbbm{1} \leq C } \sum_{i\in \mathcal{P}} \mathbb{P}_\xi(b_{i,N}(b_{i,N-1},s_{i,N-1},\theta_{i,N-1},z_{i,N-2},y_{i,N-1}) \leq \delta) \label{eq:distrib_prob}
\end{align}

Where $b_{i,N}(\cdot)$ can be thought of as the first entry in the output of $h$ for the given input state tuple. Note that to obtain the second equality, we note that each individual patient's blood glucose only depends on their particular states and whether or not they were visited and not on the states of other patients. Define $V_{i,N}^j := \mathbb{P}_\xi(b_{i,N}(b_{i,N-1},s_{i,N-1},\theta_{i,N-1},z_{i,N-2},j) \leq \delta)$ for $j \in \{0,1\}$.  Since each individual $y_{i,N-1}$ is binary this allows us to perform the following reformulation:
\begin{align}
    & \eqref{eq:distrib_prob} =\max_{\by_{N-1} \in \{0,1\}^{|\mathcal{P}|}, \by_{N-1}^\top \mathbbm{1} \leq C } \sum_{i\in \mathcal{P}} V_{i,N}^1 \cdot y_{i,N} + V_{i,N}^0 \cdot (1-y_{i,N}) \\
    & = \sum_{i\in \mathcal{P}} V_{i,N}^0 + \max_{\by_{N-1} \in \{0,1\}^{|\mathcal{P}|}, \by_{N-1}^\top \mathbbm{1} \leq C } \sum_{i\in \mathcal{P}} (V_{i,N}^1 - V_{i,N}^0) \cdot y_{i,N} \\
    & = \sum_{i\in \mathcal{P}} V_{i,N}^0 + \max_{\by_{N-1} \in [0,1]^{|\mathcal{P}|}, \by_{N-1}^\top \mathbbm{1} \leq C } \sum_{i\in \mathcal{P}} (V_{i,N}^1 - V_{i,N}^0) \cdot y_{i,N}
\end{align}

Observe that the final equality states that the linear programming (LP) relaxation of the single step problem provides the same solution as the integer programming solution. This comes from the fact that the constraint set is totally unimodular, and since $C$ is always an integer value the resulting LP solution will also be integral \citep{wolsey1999integer}. Note that since this is a linear program, strong duality holds for its Lagrangian dual. Therefore there exists a constant $\lambda_{N-1}^* \geq 0$ such that: 
\begin{equation}
    \max_{\by_{N-1}^{|\mathcal{P}|} \in [0,1]^{|\mathcal{P}|}, \by_{N-1}^\top \mathbbm{1} \leq C} \sum_{i \in \mathcal{P}} (V_{i,N}^1 - V_{i,N}^0)\cdot y_{i,N} = \max_{\by_{N-1}^{|\mathcal{P}|} \in [0,1]^{|\mathcal{P}|}} \sum_{i \in \mathcal{P}} (V_{i,N}^1 - V_{i,N}^0)\cdot y_{i,N} + \lambda_{N-1}^*\bigg(C - \sum_{i\in \mathcal{P}}y_{i,N}\bigg)
\end{equation}

If we let $\by^*_{N-1} = \argmax \Big\{ \sum_{i \in \mathcal{P}} (V_{i,N}^1 - V_{i,N}^0)\cdot y_{i,N}: \by_{N-1}^{|\mathcal{P}|} \in [0,1]^{|\mathcal{P}|}, \by_{N-1}^\top \mathbbm{1} \leq C \Big\}$ then we see:
\begin{align}
    &J_{N-1}(\bb_{N-1},\bs_{N-1},\btheta_{N-1},\bz_{N-2}) = \sum_{i \in \mathcal{P}} V_{i,N}^0 + \sum_{i\in \mathcal{P}} (V_{i,N}^1 - V_{i,N}^0) \cdot y_{i,N}^* + \lambda^*_{N-1}\bigg(C - \sum_{i\in \mathcal{P}}y_{i,N}^*\bigg), \\
    & = \sum_{i \in \mathcal{P}} V_{i,N}^0 + (V_{i,N}^1 - V_{i,N}^0- \lambda_{N-1}^*) \cdot y_{i,N}^* + \frac{C}{|\mathcal{P}|}\lambda^*_{N-1} .
\end{align}
Letting $J_{i,N-1}(\bb_{N-1},\bs_{N-1},\btheta_{N-1},\bz_{N-2}) = V_{i,N}^0 + (V_{i,N}^1 - V_{i,N}^0- \lambda_t^*) \cdot y_{i,N}^* + \frac{C}{|\mathcal{P}|}\lambda^*_t $ then we have $J_{N-1}(\bb_{N-1},\bs_{N-1},\btheta_{N-1},\bz_{N-2}) = \sum_{i\in \mathcal{P}} J_{i,N-1}(\bb_{N-1},\bs_{N-1},\btheta_{N-1},\bz_{N-2})$. Recall \eqref{eq:joint_dp_N}, we can rewrite it as:
\begin{align}
    & \mathcal{V}_{N-1}(\bb_{N-1},\bs_{N-1},\btheta_{N-1},\bz_{N-2}) = \sum_{i\in \mathcal{P}} \mathbbm{1}(b_{i,N} \leq \delta) + J_{N-1}(\bb_{N-1},\bs_{N-1},\btheta_{N-1},\bz_{N-2}) \\ 
    & = \sum_{i \in \mathcal{P}} \mathbbm{1}(b_{i,N} \leq \delta) + J_{i,N-1}(\bb_{N-1},\bs_{N-1},\btheta_{N-1},\bz_{N-2})
\end{align}

Letting $\mathcal{V}_{i,N-1}(\bb_{N-1},\bs_{N-1},\btheta_{N-1},\bz_{N-2}) = \mathbbm{1}(b_{i,N} \leq \delta) + J_{i,N-1}(\bb_{N-1},\bs_{N-1},\btheta_{N-1},\bz_{N-2})$ completes the portion of the proof and shows additive decomposability.


Next we show the claim that we would only ever visit patients in set $\mathcal{I}_{N-1}$. Note that, in the single period problem, visiting a patient who is not in set $\mathcal{I}_{N-1}$ would result in a patient who would not benefit in the short term from the visit taking resources instead of one who would by Theorem \ref{thm-1}. Thus we must only consider cases based on the size of set $\mathcal{I}_{N-1}$. If $|\mathcal{I}_{N-1}|\leq C$, then the proof follows directly from the definition of $\mathcal{I}_t$, Theorem~\ref{thm-1}, and Corollary~\ref{cor-1} because the problem is equivalent to $|\mathcal{I}_{N-1}|$ single patient problems. To prove the claim for the case where $|\mathcal{I}_{N-1}| > C$, we proceed by contradiction. Without loss of generality, assume $\exists j \in \mathcal{S}_{N-1}, \ell \notin \mathcal{S}_{N-1}$ such that
\begin{align*}
    V_{\ell,N}^1 - V_{\ell,N}^0 & > V_{j,N}^1 - V_{j,N}^0.
\end{align*}
Let $\mathcal{S}'_{N-1}$ be the set obtained by swapping the elements $j$ and $\ell$, that is, $\mathcal{S}'_{N-1} = \mathcal{S}_{N-1} \cup \{\ell\} \setminus \{j\} $.
Let us consider the relationship between the value function of visiting patients in set $\mathcal{S}'_{N-1}$ instead of $\mathcal{S}_{N-1}$.
\begin{align}
&\sum_{i \notin \mathcal{S}_{N-1}} V_{i,N}^0 + \sum_{i \in \mathcal{S}_{N-1}} V_{i,N}^1 \\
&= \sum_{i \notin \mathcal{S}_{N-1}\setminus\{\ell\}} V_{i,N}^0 + \sum_{i \in \mathcal{S}_{N-1}\setminus\{j\}} V_{i,N}^1 + V_{j,N}^1 + V_{\ell, N}^0 \\
&< \sum_{i \notin \mathcal{S}_{N-1}\setminus\{\ell\}} V_{i,N}^0 + \sum_{i \in \mathcal{S}_{N-1}\setminus\{j\}} V_{i,N}^1 + V_{j,N}^0 + V_{\ell, N}^1 \\
&= \sum_{i \notin \mathcal{S}'_{N-1}} V_{i,N}^0 + \sum_{i \in \mathcal{S}'_{N-1}} V_{i,N}^1 
\end{align}
However, $\mathcal{S}_{N-1}$ is defined as the set of patients where $y_{i,N-1}=1$ that maximizes the sum of value functions for all patients. Therefore, we have reached a contradiction and conclude the proof.
\halmos
\endproof

\proof{Proof of Proposition \ref{prop:lagrange_struct}:}
We will proceed to prove all three results by induction.

\emph{Proof of~Proposition~\ref{prop:lagrange_struct}.1:} We will proceed by induction on time index $t$. For the base case, note that for any $\blambda$ by \eqref{eq:lagrange_joint_dp_N} we have that $\mathcal{V}_N^{\blambda} = \mathcal{V}_N$, where the omission of the arguments implies the equality holds for all states. Clearly then 1 holds at time $N$. Next, let our induction hypothesis be that for some $t<N$, $\mathcal{V}_{t+1} \leq \mathcal{V}^{\blambda}_{t+1}$. Then for any $\bb,\bs,\btheta,\bz \in \mathbb{R}_+^{3\times |\mathcal{P}|}\times\{0,1\}^{|\mathcal{P}|}$:
\begin{align}
    & \mathcal{V}_t(\bb,\bs,\btheta,\bz) = \sum_{i \in \mathcal{P}} \mathbbm{1}(b_{i} \leq \delta) + \max_{\by_t \in \{0,1\}^{|\mathcal{P}|},\by_t^\top\mathbbm{1}\leq C}  \mathbb{E}_{\bxi} \Bigg\{\mathcal{V}_{t+1}\big(h(\bb,\bs,\btheta,\bz,\by_t)\big) \Bigg\}, \\
    & \leq \sum_{i \in \mathcal{P}} \mathbbm{1}(b_{i} \leq \delta) + \max_{\by_t \in \{0,1\}^{|\mathcal{P}|},,\by_t^\top\mathbbm{1}\leq C}  \mathbb{E}_{\bxi} \Bigg\{\mathcal{V}^{\blambda}_{t+1}\big(h(\bb,\bs,\btheta,\bz,\by_t)\big) + \lambda_t \bigg(C - \sum_{i\in \mathcal{P}} y_{i,t}\bigg) \Bigg\}\\
    & \leq \sum_{i \in \mathcal{P}} \mathbbm{1}(b_{i} \leq \delta) + \max_{\by_t \in \{0,1\}^{|\mathcal{P}|}}  \mathbb{E}_{\bxi} \Bigg\{\mathcal{V}^{\blambda}_{t+1}\big(h(\bb,\bs,\btheta,\bz,\by_t)\big) + \lambda_t \bigg(C - \sum_{i\in \mathcal{P}} y_{i,t}\bigg) \Bigg\} = \mathcal{V}_t^{\blambda}(\bb,\bs,\btheta,\bz)
\end{align}
Where the first equation follows from \eqref{eq:joint_dp_N}, the next inequality follows by the induction hypothesis and noting that because $\blambda \geq 0$ the quantity $\lambda_t \bigg(C - \sum_{i\in \mathcal{P}} y_{i,t}\bigg) \geq 0$ on the feasible region, the following inequality follows since it is a relaxation, and the last equality holds by \eqref{eq:lagrange_joint_dp_N}. This thus completes the proof for Property 1.

\emph{Proof of Proposition~\ref{prop:lagrange_struct}.2:} Again we proceed by induction. For the base case note that at time $N$ $\mathcal{V}_N^{\blambda}$ is clearly additively decomposable and satisfies \eqref{eq:langrange_singele_patient}. Then for the induction hypothesis, suppose that for some $t < N$, $\mathcal{V}_{t+1}^{\blambda}$ is additively decomposable and satisfies \eqref{eq:langrange_singele_patient}. Then we can proceed by using \eqref{eq:lagrange_joint_dp_N}:
\begin{align}
    & \mathcal{V}_t^{\blambda}(\bb,\bs,\btheta,\bz) = \sum_{i \in \mathcal{P}} \mathbbm{1}(b_{i} \leq \delta) + \max_{\by_t \in \{0,1\}^{|\mathcal{P}|}}  \mathbb{E}_{\bxi} \Bigg\{\mathcal{V}^{\blambda}_{t+1}\big(h(\bb,\bs,\btheta,\bz,\by_t)\big) + \lambda_t \bigg(C - \sum_{i\in \mathcal{P}} y_{i,t}\bigg) \Bigg\} \\
    & =  \sum_{i \in \mathcal{P}} \mathbbm{1}(b_{i} \leq \delta) + \max_{\by_t \in \{0,1\}^{|\mathcal{P}|}}  \mathbb{E}_{\bxi} \Bigg\{C \sum_{k = t+1}^N\lambda_k + \sum_{i \in \mathcal{P}} \mathcal{V}_{i,t+1}^{\blambda}(h_i(b_i,s_i,\theta_i,z_i,y_{i,t})) + \lambda_t \bigg(C - \sum_{i\in \mathcal{P}} y_{i,t}\bigg) \Bigg\} \\
    & =  C \sum_{k = t}^N\lambda_k + \max_{\by_t \in \{0,1\}^{|\mathcal{P}|}}  \mathbb{E}_{\bxi}\Bigg\{ \sum_{i \in \mathcal{P}} \mathbbm{1}(b_{i} \leq \delta) +  \sum_{i \in \mathcal{P}} \mathcal{V}_{i,t+1}^{\blambda}(h_i(b_i,s_i,\theta_i,z_i,y_{i,t})) - \lambda_t \sum_{i\in \mathcal{P}} y_{i,t} \Bigg\} \\
    & = C \sum_{k = t}^N\lambda_k + \sum_{i \in \mathcal{P}} \mathcal{V}_{i,t}^{\blambda}(b_i,s_i,\theta_i,z_i)
\end{align}
Where the final equality follows by definition of $\mathcal{V}_{i,t}^{\blambda}$, thus completing the proof of of Property 2.

\emph{Proof of Proposition~\ref{prop:lagrange_struct}.3:} We again proceed by induction. Note that $\mathcal{V}_N^{\blambda}$ is not dependent on $\lambda$ and is thus clearly convex in $\blambda$. Next, by induction hypothesis, suppose that for some $t <N$ we have that $\mathcal{V}_{t+1}^{\blambda}$ is convex in $\blambda$. Then by \eqref{eq:lagrange_joint_dp_N}, note that $\mathcal{V}_t^{\blambda}$ is formed by expectation, and point-wise maximization of convex functions of $\blambda$, both operations that preserve convexity. Therefore, $\mathcal{V}_t^{\blambda}$ is also convex thus completing the proof. \halmos
\endproof

\proof{Proof of Proposition \ref{prop:lagrange_single_struct}:} Using \eqref{eq:langrange_singele_patient} and making the substitution for $V_{i,t}^{\blambda,j}$ we can express $\mathcal{V}_{i,t}^{\blambda}$ as follows:
\begin{align}
    &\mathcal{V}_{i,t}^{\blambda}(b_{i,t},s_{i,t},\theta_{i,t},z_{i,t-1}) = \mathbbm{1}(b_{i,t}\leq \delta) + \max_{y_{i,t} \in \{0,1\}} \mathbb{E}_\xi \mathcal{V}^{\blambda}_{i,t+1}\big(h(b_{i,t},s_{i,t},\theta_{i,t},z_{i,t-1},y_{i,t})\big) - \lambda_t y_{i,t}, \\
    &= \mathbbm{1}(b_{i,t}\leq \delta) + \max_{y_{i,t}\in\{0,1\}} \Big\{V_{i,t}^{\blambda,1}y_ {i,t} + V_{i,t}^{\blambda,0}(1-y_ {i,t}) - \lambda_ty_{i,t}\Big\} \\
    &= \mathbbm{1}(b_{i,t}\leq \delta) + V_{i,t}^{\blambda,0} + \max_{y_{i,t}\in\{0,1\}} \Big\{(V_{i,t}^{\blambda,1} - V_{i,t}^{\blambda,0} - \lambda_t )y_ {i,t} \Big\}
\end{align}

The result directly follows from solving the final optimization problem above for $y_{i,t}$. \halmos 
\endproof

\proof{Proof of Proposition \ref{prop:lagrange_relax_gap}} This proof proceeds by induction on $t$. First consider time $N$, and note that clearly $\mathcal{V}_N^{\blambda^*(\cdot)}(\bb,\bs,\btheta,\bz) - \mathcal{V}_N(\bb,\bs,\btheta,\bz) = 0$ for all possible state values. For our base case let us consider time index $N-1$. Then observe:
\begin{align}
& \mathcal{V}_{N-1}^{\blambda^*(\cdot)}(\bb,\bs,\btheta,\bz) - \mathcal{V}_{N-1}(\bb,\bs,\btheta,\bz) \nonumber\\
& = \min_{\lambda_{N-1} \geq 0}\max_{\by\in \{0,1\}^{|\mathcal{P}|}} \Bigg\{\mathbb{E}_{\bxi}\mathcal{V}^{\blambda^*(\cdot)}_{N}\big(h(\bb,\bs,\btheta,\bz,\by)\big) + \lambda_{N-1} \bigg(C - \sum_{i \in \mathcal{P}}y_{i}\bigg) \Bigg\} - \max_{\substack{\by\in \{0,1\}^{|\mathcal{P}|},\\ \by^\top\mathbbm{1} \leq C}} \mathbb{E}_{\bxi}\mathcal{V}_{N}\big(h(\bb,\bs,\btheta,\bz,\by)\big) \\
& = \min_{\lambda_{N-1} \geq 0}\max_{\by\in \{0,1\}^{|\mathcal{P}|}} \Bigg\{\mathbb{E}_{\bxi}\mathcal{V}_{N}\big(h(\bb,\bs,\btheta,\bz,\by)\big) + \lambda_{N-1} \bigg(C - \sum_{i \in \mathcal{P}}y_{i}\bigg) \Bigg\} - \max_{\substack{\by\in \{0,1\}^{|\mathcal{P}|},\\ \by^\top\mathbbm{1} \leq C}} \mathbb{E}_{\bxi}\mathcal{V}_{N}\big(h(\bb,\bs,\btheta,\bz,\by)\big) \leq 2
\end{align}

Where the first equality follows from our previous observation and the fact that ${\blambda}$ is state dependent, and final inequality comes from Proposition 5.26 in \cite{bertsekas1982constrained}. This proposition applies since our action space is discrete, and the constraint set is made of a single convex function. Since $N - (N-1) = 1$, this follows our desired result. Now we proceed with the induction hypothesis. Suppose that for some $t < N-1$ we have that $\mathcal{V}_{t+1}^{\blambda^*(\cdot)}(\bb,\bs,\btheta,\bz) - \mathcal{V}_{t+1}(\bb,\bs,\btheta,z) \leq 2(N-t-1)$ for all possible state. Then we can proceed as follows:
\begin{align}
    &\mathcal{V}_{t}^{\blambda^*(\cdot)}(\bb,\bs,\btheta,\bz) - \mathcal{V}_{t}(\bb,\bs,\btheta,\bz) \nonumber\\
    &  = \min_{\lambda_t \geq 0}\max_{\by\in \{0,1\}^{|\mathcal{P}|}} \Bigg\{\mathbb{E}_{\bxi}\mathcal{V}^{\blambda^*(\cdot)}_{t+1}\big(h(\bb,\bs,\btheta,\bz,\by)\big) + \lambda_t \bigg(C - \sum_{i \in \mathcal{P}}y_{i}\bigg) \Bigg\} - \max_{\by\in \{0,1\}^{|\mathcal{P}|}, \by^\top\mathbbm{1} \leq C} \mathbb{E}_{\bxi}\mathcal{V}_{N}\big(h(\bb,\bs,\btheta,\bz,\by)\big) \\
    & \leq 2(N-t-1) + \min_{\lambda_t \geq 0}\max_{\by\in \{0,1\}^{|\mathcal{P}|}} \Bigg\{\mathbb{E}_{\bxi}\mathcal{V}_{t+1}\big(h(\bb,\bs,\btheta,\bz,\by)\big) + \lambda_t \bigg(C - \sum_{i \in \mathcal{P}}y_{i}\bigg) \Bigg\} \nonumber\\
    &\phantom{\leq} - \max_{\by\in \{0,1\}^{|\mathcal{P}|}, \by^\top\mathbbm{1} \leq C} \mathbb{E}_{\bxi}\mathcal{V}_{N}\big(h(\bb,\bs,\btheta,\bz,\by)\big) \\
    & \leq 2(N-t)
\end{align}

Where first equality follows by the same reasoning as the base case, the second inequality follows from the induction hypothesis, and the final inequality again follows from Proposition 5.26 in \cite{bertsekas1982constrained}. Thus proving our desired result. \halmos
\endproof

\proof{Proof of Proposition \ref{prop:whittles_struct}:} Results 1 and 2 follow by a near identical argument to Results 1 and 2 in Proposition \ref{prop:lagrange_struct}, thus we will present only a proof for Result 3. Clearly $\mathcal{V}_{i,t}^{\omega}(b_{i,t},s_{i,t},\theta_{i,t},z_{i,t-1}) \in [\frac{-\omega(N-t)}{N}, N-t ] $ from \eqref{eq:whittles_single_patient}. Thus it just remains to show that it is convex and monotonically non-increasing in $\omega$. This can be done by induction. Note that $\mathcal{V}_{i,N}^{\omega}$ is not dependent on $\omega$ and thus trivially satisfies these conditions. Thus let us consider time index $N-1$.
\begin{align}
  &\mathcal{V}_{i,N-1}^{\omega}(b_{i,N-1},s_{i,N-1},\theta_{i,N-1},z_{i,N-2}) = \nonumber\\
  &= \mathbbm{1}(b_{i,N-1}\leq \delta) + \max_{y_{i,N-1} \in \{0,1\}} \mathbb{E}_\xi \mathcal{V}^{\omega}_{i,N}\big(h_i(b_{i,N-1},s_{i,N-1},\theta_{i,N-1},z_{i,N-2},y_{i,N-1})\big) -\frac{\omega}{N} y_{i,N-1} \\
   &= \mathbbm{1}(b_{i,N-1}\leq \delta) + \max_{y_{i,N-1} \in \{0,1\}} \mathbb{P}_\xi(b_{i,N}(b_{i,N-1},s_{i,N-1},\theta_{i,N-1},z_{i,N-2},y_{i,N-1}) \leq \delta \big) - \frac{\omega}{N} y_{i,N-1}
\end{align}

Solving for $y_{i,N-1}$ above shows that:
\begin{align}
    &\mathcal{V}_{i,N-1}^{\omega}(b_{i,N-1},s_{i,N-1},\theta_{i,N-1},z_{i,N-2})= \nonumber\\ 
    &= \mathbbm{1}(b_{i,N-1}\leq \delta) + (\mathbb{P}_\xi(b_{i,N}(b_{i,N-1},s_{i,N-1},\theta_{i,N-1},z_{i,N-2},1) \leq \delta) \nonumber\\
    &\phantom{=} - \mathbb{P}_\xi(b_{i,N}(b_{i,N-1},s_{i,N-1},\theta_{i,N-1},z_{i,N-2},0) \leq \delta) - \omega)^+ + \mathbb{P}_\xi(b_{i,N}(b_{i,N-1},s_{i,N-1},\theta_{i,N-1},z_{i,N-2},0) \leq \delta).
\end{align}

Thus it is clearly convex and non-increasing $\omega$. Next we can proceed by induction. Suppose by induction hypothesis that for some $t < N-1$, $\mathcal{V}_{i,t+1}^{\omega}(b_{i,t+1},s_{i,t+1},\theta_{i,t+1},z_{i,t})$ is convex and non-increasing.  Then notice that $\mathcal{V}_{i,t}^{\omega}(b_{i,t},s_{i,t},\theta_{i,t},z_{i,t-1})$ is formed by functional composition with a convex increasing function, expectation, and point-wise minimization of convex non-increasing functions. Thus by Lemma \ref{lemma:technical} and properties of convex functions and monotonic functions, it is clear that $\mathcal{V}_{i,t}^{\omega}(b_{i,t},s_{i,t},\theta_{i,t},z_{i,t-1})$ is convex and non-increasing. \halmos
\endproof
\section{Technical Lemma}
\begin{lemma}
\label{lemma:technical}
    Let $f:\mathbb{R} \mapsto \mathbb{R}$ and $g: \mathbb{R} \mapsto \mathbb{R}$ both be nonincreasing functions. Then $h:\mathbb{R} \mapsto \mathbb{R}$ where $\forall x\in \mathbb{R}: \ h(x) = \min \{ f(x),g(x)\}$ is also non-increasing.
\end{lemma}

\proof{Proof: } Let $x,y \in \mathbb{R}$ such that $x \geq y$. Then by definition $f(x) \leq f(y)$ and $g(x) \leq g(y)$. Thus clearly $f(y) \geq \min\{f(x),g(x)\} $ and $g(y) \geq \min\{f(x),g(x)\} $. Therefore $h(y) = \min\{f(y),g(y)\} \geq \min\{f(x),g(x)\} = h(x) $ showing that $h$ is non-increasing as desired. \halmos 
\endproof

\section{Patient parameter estimation}\label{EC_patparamest}

To solve the MLE problem in Proposition~\ref{prop:mle_form}, we modeled the FBG disturbance $\epsilon_{i,t}$ using a Laplace distribution and used a grid search procedure for $s_0$, $\beta$, $\rho$, and $\gamma$ ($s_0$, $\beta$ only for expanded scenarios), allowing us to cast Problem~\eqref{eq:mle_prob} as a mixed integer linear program. Specifically, the choice of a Laplace distribution allowed us to linearize the objective function with an absolute value by introducing auxiliary variables \citep{wolsey1999integer}, while the grid search fixed the variables leading to multilinear terms each time the optimization problem was solved, resulting in a mixed integer linear problem. we performed grid search on two parameters ($s_0$ and $\beta$), where we searched over $S=\{0,1,2,3\}$ for $s_0$ and $B=\{0,1,2,3\}$ for $\beta$. We define the following variables for the grid search procedure. Let $x^*$ denote the best optimal solution found through grid search and let $w^*$ denote corresponding optimal value. We let $\Bar{x}$ be a variable that stores the current best optimal solution, let $\Bar{w}$ be a variable that stores the current best optimal value, let $w^{\textnormal{UB}}$ denote the upper bound on $w^*$ to initialize the algorithm, and let $w$ represent the optimal value to a particular grid search instance obtained by solving the parameter estimation problem stated in Proposition~\ref{prop:mle_form} (denoted below as optProb). 

\begin{algorithm2e}[H]
    \DontPrintSemicolon
    \KwData{$(\by, \bz, \bar{\bb}, S, B)$}
    \KwResult{$w^*$, $x^*$ = ($\hat{\bb}^*$, $\bs^*$, $\btheta^*$, $p^*$, $\mu^*$, $\alpha^*$, $\theta_0^*$, $\lambda^*$, $s_0^*$, $\beta^*$)}
    \Begin{
        $\Bar{w}= w^{\textnormal{UB}}$
        
        $\Bar{x} = \emptyset$
        
        searchGrid = list($S \times B$)
    
        \For{$i \in |S \times B|$}{
            $s_0^i$, $\beta^i$ = searchGrid[$i$]
            
            ($w$, $\hat{\bb}$, $\bs$, $\btheta$, $p$, $\mu$, $\alpha$, $\theta_0$, $\lambda$) = optProb($\by$, $\bz$, $\bar{\bb}$, $s_0^i$, $\beta^i$)
            
            \If{$w < \Bar{w}$}{
                $\Bar{w} = w$ 
                
                $\Bar{x}$ = ($\hat{\bb}$, $\bs$, $\btheta$, $p$, $\mu$, $\alpha$, $\theta_0$, $\lambda$, $s_0^i$, $\beta^i$)
            }      
        }
        
        $w^* = \Bar{w}$
        
        $x^* = \Bar{x}$
    }
    \caption{Coarse grid search for parameter estimation. \label{alg:grid-search}}
\end{algorithm2e}

\section{NanoHealth patient parameter estimation results}\label{EC_patparamresults}

We solved the problem stated in Proposition~\ref{prop:mle_form} using patient visit history data from NanoHealth (described in Section~\ref{sec:data-description}). Then, we used $k$-means clustering to partition NanoHealth's patient cohort into distinct clusters based on seven features: FBG progression ($p$), effect of enrollment ($\mu$), effect of a visit ($\alpha$), baseline perception of adverse factors ($\theta_0$), decrease in perception of adverse factors per visit ($\lambda$), baseline in adverse factors state ($s_0$), and increase in adverse factors per visit ($\beta$). We used an elbow plot to identify the ideal number of clusters: four. Table~\ref{tab:clusters} displays the proportion of patients in each cluster and the mean patient parameter values for each cluster (i.e., the centroid). 

\begin{table}[!h]
  \centering 
  \begin{tabular}{>{\centering\arraybackslash}p{8cm}|>{\centering\arraybackslash}p{1.5cm}>{\centering\arraybackslash}p{1.5cm}>{\centering\arraybackslash}p{1.5cm}>{\centering\arraybackslash}p{1.5cm}}
    \toprule
    \multirow{2}{*}{Patient parameter} & \multicolumn{4}{c}{Cluster}\\\cline{3-4} 
     & 0 &  1 &  2 & 3 \\ \midrule
    Disease progression ($p$) & 0.091 & 6.994 & 0.039 & 6.990  \\
    Enrollment effect ($\mu$) & 0.006 & 6.899 & 0.009 &  0.011  \\
    Visit effect ($\alpha$) & 0.109 & 0.125 & 0.050 & 6.997 \\
    Baseline perception of adverse factors ($\theta_0$) & 0.001 & 0 & 0.002 & 0 \\
    Reduction in perception of adverse factors ($\lambda$) & 0 & 0 & 0.001 & 0 \\
    Baseline adverse factors ($s_0$) & 0.072 & 0 & 0.060 & 0  \\
    Increase in adverse factors ($\beta$) & 0 & 0.011 & 1.040 & 0 \\ \midrule
    \% of patients & 47.9\% & 23.8\% & 26.5\% & 1.9\% \\ \midrule
    No. of patients ($N=378$) & 181 & 90 & 100 & 7\\ \bottomrule
  \end{tabular}
  \caption{Cluster centroids and the percentage of patients in each cluster for NanoHealth's cohort}
  \label{tab:clusters}
\end{table}

\section{Enrollment algorithm details}\label{EC_EAdeets}

Algorithm~\ref{alg:multi-patient-heuristic} provides a detailed description of the enrollment algorithm.

\begin{algorithm2e}[H]
    \DontPrintSemicolon
    \KwData{$(P, C, N, f)$: Set of patients with known initial states and parameters, capacity, length of planning horizon, heuristic sorting function.}
    \KwResult{$\hat{S}$: Visit schedule specifying which patients should be visited in each period.}
    \Begin{
        \For{$t \in \{1,\dots, N\}$}{
            \For{$i \in \{1, \dots, |\mathcal{P}|\}$}{
                \If{$i$ \textnormal{meets conditions in Corollary~\ref{cor-1}}}{
                    $\mathcal{I}_t \longleftarrow \mathcal{I}_t \cup \{i\}$ 
                }
            }
            \If{$|\mathcal{I}_t| \leq C$}{
                $\hat{\mathcal{S}}_t \longleftarrow \mathcal{I}_t$
            }
            \Else{
                $\Bar{\mathcal{I}}_t \longleftarrow f(\mathcal{I}_t)$ \Comment{Sort set $\mathcal{I}_t$ based on chosen heuristic to obtain the sequence $\Bar{\mathcal{I}}_t$}
                
                $\hat{\mathcal{S}}_t \longleftarrow \{(a_n)_{n=1}^C | a_n \in \Bar{\mathcal{I}}_t\}$  \Comment{Select first $C$ patients from $\Bar{\mathcal{I}}_t$ to be visited in period $t$}
            }
            \For{$i \in \{1, \dots, |\mathcal{P}|\}$}{
                \If{$i \in \hat{\mathcal{S}}_t$}{
                    $y_{i,t}=1$
                }
                \Else{
                    $y_{i,t}=0$
                }
                
                $\:\:\: z_{i,t} = (z_{i,t-1} + y_{i,t} - z_{i,t-1} y_{i,t}) \mathbbm{1}[B_{i,t} \geq 0]$\\
                $b_{i,t+1} = b_{i,t} + p_i - \mu_i z_{i,t} - \alpha_i y_{i,t} z_{i,t}$ \\
                $s_{i,t+1} = z_{i,t}(\gamma (s_{i,t}-s_{i,0}) + s_{i,0})+\beta_i y_{i,t} z_{i,t}$ \\
                $\theta_{i,t+1} = \rho (\theta_{i,t} - \theta_{i,0}) + \theta_{i,0} - \lambda_i y_{i,t} z_{i,t}$
            }
        }
    }
    \caption{Enrollment algorithm \label{alg:multi-patient-heuristic}}
\end{algorithm2e}

\section{Full information numerical experiments}

\subsection{Background and setup}

\begin{figure}
    \centering
    \includegraphics[width=0.6\linewidth]{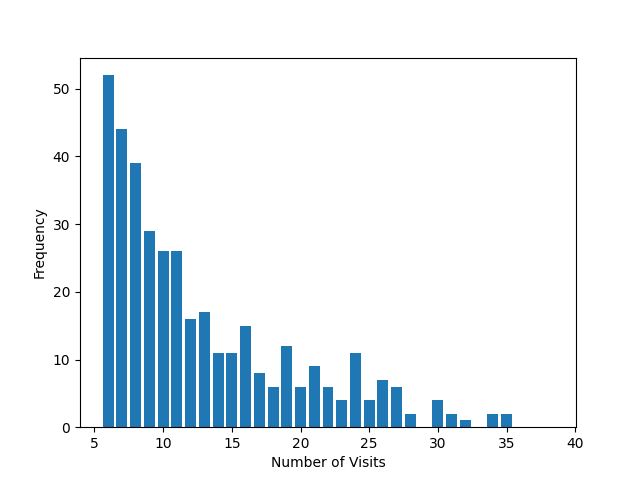}
    \caption{Distribution of number of visits per patient in the dataset provided by NanoHealth filtered by patients who had at least 6 visits.}
    \label{fig:nanohealth-data-visit-distribution}
\end{figure}

Figure~\ref{fig:nanohealth-data-visit-distribution} shows the visit distribution for the dataset provided by NanoHealth, Table~\ref{table:nanohealth-summary-data} provides summary statistics for the NanoHealth cohort, Table~\ref{table:patient_observations} provides summary statistics for visit information from NanoHealth's patient cohort, Table~\ref{table:nanohealth-summary-data-missing} presents the missingness for each feature, and 
Table~\ref{table:clusters-and-scenarios} provides summary statistics for five patient groups and corresponding scenarios.

\begin{table}[ht]
\begin{center}
\begin{tabular}{l c r}
\hline
Characteristic && \\ \hline
Number of patients && 378 \\ 
Age in years, mean (sd) && 53.3 (11.0) \\ 
Female, n (\%) && 219 (59.2) \\
Body mass index (kg/m$^2$), mean (sd) && 27.6 (5.9)\\
Waist circumference (inches), mean (sd) && 37.1 (4.3) \\
Heart rate (beats per minute), mean (sd) && 83.9 (12.3) \\ 
Diastolic blood Pressure (mmHg), mean (sd) && 90.2 (11.9) \\
Systolic blood Pressure (mmHg), mean (sd) && 138.2 (21.6) \\
Tobacco smoker, n (\%) && 29 (7.8)\\
Initial FBG (mg/dL), mean (sd) && 175.1 (71.9) \\
Final FBG (mg/dL), mean (sd) && 156.8 (61.6) \\
Number of management visits, mean (sd) && 13.1 (7.6) \\
Visits per month, mean (sd) && 0.6 (0.4) \\
\hline
\end{tabular}
\end{center}
\caption{Summary statistics from NanoHealth's patient cohort. See Table ~\ref{table:nanohealth-summary-data-missing} for the number of missing observations for each characteristic. \label{table:nanohealth-summary-data}}
\end{table}

\begin{table}[ht]
    \centering
    \begin{tabular}{l c}\hline
        Number of visits &  \\\hline
        Total across all patients & 5322\\
        Minimum per patient & 6 \\
        Maximum per patient & 39\\
        Average per patient & 14.1\\
        Standard deviation & 7.6\\\hline
    \end{tabular}
    \caption{Number of observations from NanoHealth's cohort.}
    \label{table:patient_observations}
\end{table}

\begin{table}[ht]
\begin{center}
\begin{tabular}{l c r}
\hline
Characteristic && Number of missing observations\\ \hline
Age in years&& 8 (2.1\%) \\ 
Female && 8 (2.1\%)\\
Body mass index && 133 (35.2\%) \\
Waist circumference  && 147 (38.9\%) \\
Heart rate  && 119 (31.5\%) \\ 
Diastolic blood Pressure  && 8 (2.1\%) \\
Systolic blood Pressure  && 18 (4.8\%) \\
Tobacco smoker && 8 (2.1\%)\\
Initial FBG  && 8 (2.1\%)\\
Final FBG  && 0 \\
Number of management visits && 0 \\
Visits per month && 0 \\
\hline
\end{tabular}
\end{center}
\caption{Missingness for NanoHealth's patient cohort. \label{table:nanohealth-summary-data-missing}}
\end{table}


\begin{table}
    \centering
    \begin{tabular}{>{\centering\arraybackslash}p{3cm}|>{\centering\arraybackslash}p{1.5cm}>{\centering\arraybackslash}p{1.5cm}>{\centering\arraybackslash}p{1.5cm}>{\centering\arraybackslash}p{1.5cm}>{\centering\arraybackslash}p{1.5cm}} 
        \hline
        Patient & \multicolumn{5}{c}{Group}\\\cline{4-4}
        parameter & A &  B &  C & D & E\\ \hline
        $p$ & 0.05 & 5 & 5 & 7.5 & 0.05 \\
        $\mu$ & 0.025 & 4 & 2 & 4 & 0.025 \\
         $\alpha$ & 0.1 & 2 & 4 & 2 & 0.35 \\
         $\theta_0$ & 0.7 & 0.7 & 0.7 & 0.7 & 2 \\
         $\lambda$ & 0.5 & 0.5 & 0.5 & 0.5 & 1.5 \\
         $s_0$ & 1 & 0.2 & 0.2 & 0.2 & 0.2 \\
         $\beta$ & 0.3 & 1.5 & 1.5 & 1.5 & 1.5 \\ \hline 
         \multicolumn{6}{c}{Simulation cohort composition} \\ \hline
             Scenario 1 & 20\% & 20\% & 20\% & 20\% & 20\% \\
             Scenario 2 & 0\% & 50\% & 0\% & 50\% & 0\% \\
             Scenario 3 & 0\% & 50\% & 0\% & 0\% & 50\% \\ \hline
    \end{tabular}
    \caption{The table displays the mean value for each parameter for each group and the percentage of patients in each group. The bottom panel displays the percentage of patients from each group in the corresponding artificial cohort for each scenario.
\label{table:clusters-and-scenarios}}
\end{table}

\subsection{MLE fit for NanoHealth}

\begin{figure}
    \centering
        \begin{subfigure}{0.32\textwidth}
        \includegraphics[width=\linewidth, trim={0.3cm 0 1.5cm 0}, clip]{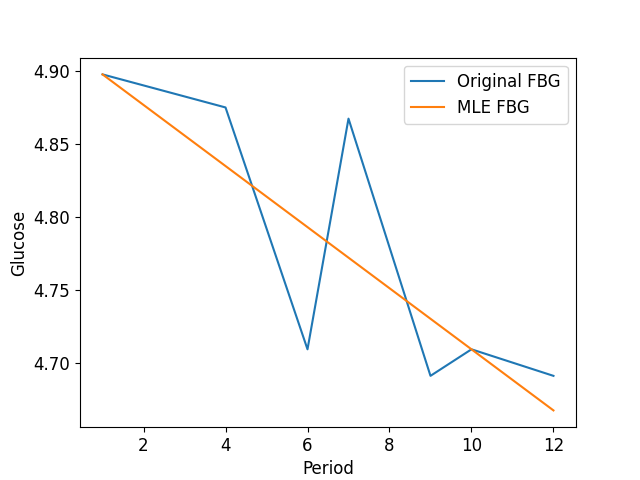}
        \caption{Patient A} \label{fig:mle-actual-estimated-line-a}
    \end{subfigure}
    \begin{subfigure}{0.32\textwidth}
        \includegraphics[width=\linewidth, trim={0.3cm 0 1.5cm 0}, clip]{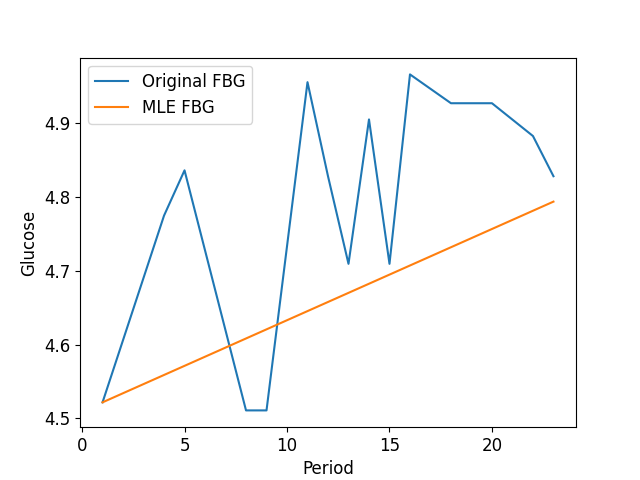}
        \caption{Patient B} \label{fig:mle-actual-estimated-line-b}
    \end{subfigure}
    \begin{subfigure}{0.32\textwidth}
        \includegraphics[width=\linewidth, trim={0.3cm 0 1.5cm 0}, clip]{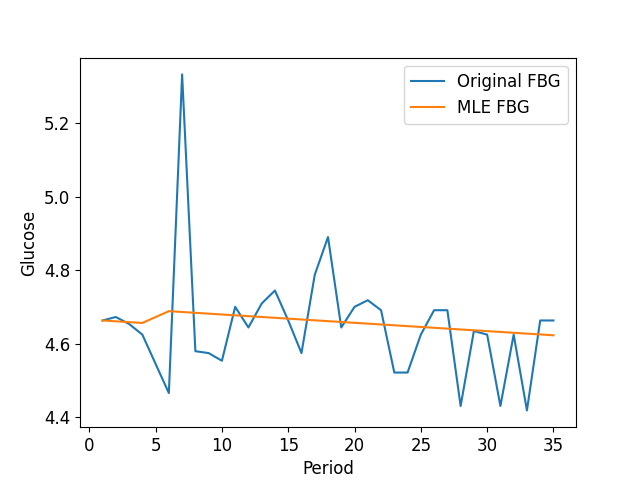}
        \caption{Patient C} \label{fig:mle-actual-estimated-line-c}
    \end{subfigure}
    \vspace{5pt}
    \caption{Line graphs showing actual and estimated FBG levels for different NanoHealth patients.}
    \label{fig:mle-actual-estimated-line}
\end{figure}

\begin{figure}
    \centering
    \vfill
        \begin{subfigure}{0.32\textwidth}
        \includegraphics[width=\linewidth, trim={0.2cm 0.5cm 0.2cm 0}, clip]{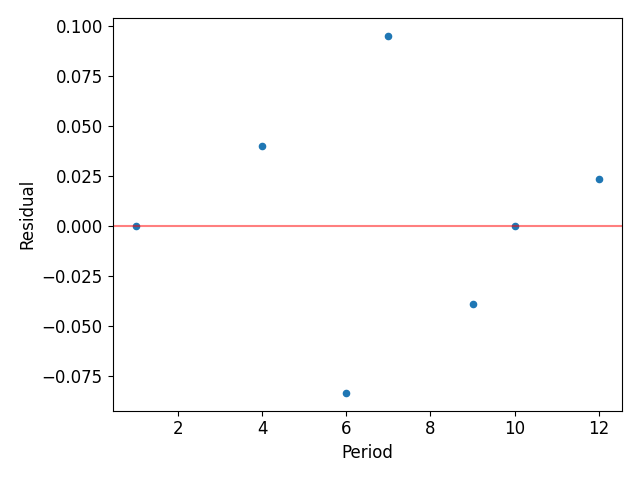}
        \caption{Patient A} \label{fig:mle-actual-estimated-scatter-a}
    \end{subfigure}
    \begin{subfigure}{0.32\textwidth}
        \includegraphics[width=\linewidth, trim={0.2cm 0 1.5cm 0}, clip]{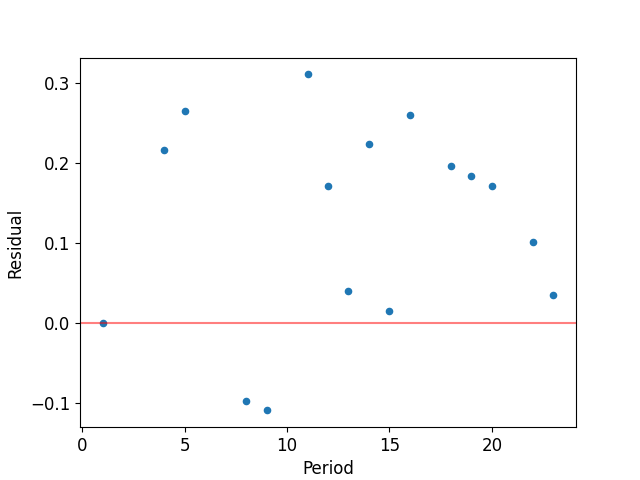}
        \caption{Patient B} \label{fig:mle-actual-estimated-scatter-b}
    \end{subfigure}
    \begin{subfigure}{0.32\textwidth}
        \includegraphics[width=\linewidth, trim={0.2cm 0 1.5cm 0}, clip]{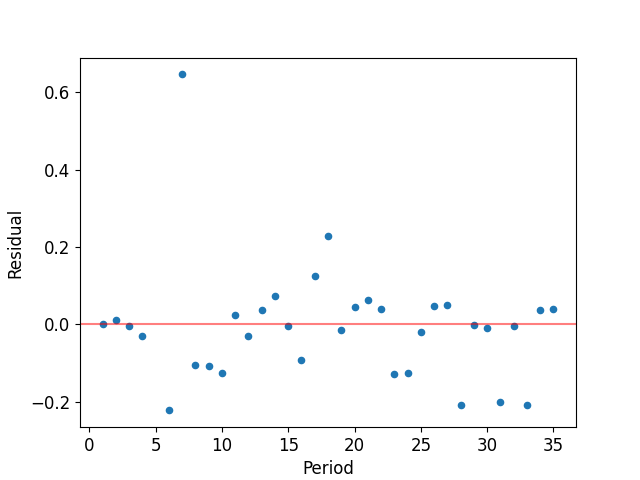}
        \caption{Patient C} \label{fig:mle-actual-estimated-scatter-c}
    \end{subfigure}
    \vspace{5pt}
    \caption{Scatter plots showing the residuals between actual and estimated FBG levels for different NanoHealth patients.}
    \label{fig:mle-actual-estimated-scatter}
\end{figure}

\begin{figure}
    \centering
    \includegraphics[width=0.6\linewidth]{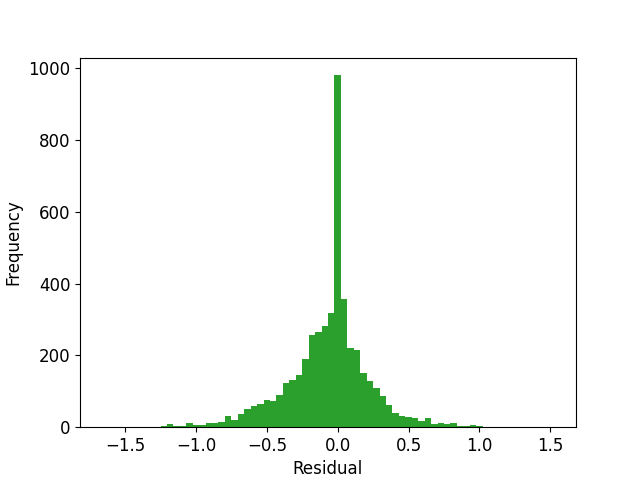}
    \caption{Histogram showing the residual distribution for the FBG of all periods and patients from NanoHealth.}
    \label{fig:residuals-hist}
\end{figure}

To demonstrate that our MLE approach appropriately models the FBG trajectories and that the disturbance terms are independent and identically distributed, we are including line graphs and residual plots created by comparing the actual and estimated FBG trajectories of a small subset of NanoHealth patients in Figures~\ref{fig:mle-actual-estimated-line} and \ref{fig:mle-actual-estimated-scatter}. We are also including a histogram with the residuals for all patients and periods in Figure~\ref{fig:residuals-hist}, which shows that the average residual is close to zero.

\subsection{Run times}

Table~\ref{table:computation-times-full-info} shows the run times in seconds for all cohorts/scenarios, capacities, and heuristics. All run times were averaged across 10 replications. Note that the Lagrangian does not include the initial time to obtain the multipliers by solving an LP, which took around 10-15 minutes for each scenario.
\begin{table}[ht]
\begin{center}
\begin{tabular}{p{3cm}|>{\centering\arraybackslash}p{1.5cm}|>{\centering\arraybackslash}p{2.4cm}>{\centering\arraybackslash}p{2.4cm}>{\centering\arraybackslash}p{2.4cm}>{\centering\arraybackslash}p{2.4cm}}
\hline
\multicolumn{1}{c|}{\multirow{2}{*}{Heuristic}} & \multirow{2}{*}{Scenario} & \multicolumn{4}{c}{Capacity (\%)}\\\cline{4-5}
 & & 20 & 40 & 60 & 80 \\ \hline
\multirow{4}{*}{Ascending FBG} & NH & 17.3 (0.4) & 17.6 (0.4) & 17.7 (0.6) & 17.6 (0.5) \\
& 1 & 38.2 (3.0) & 37.2 (1.4) & 34.2 (0.5) & 34.6 (0.5)  \\
& 2 & 17.3 (0.4) & 17.6 (0.4) & 17.7 (0.6) & 17.6 (0.5) \\
& 3 & 17.3 (0.4) & 17.6 (0.4) & 17.7 (0.6) & 17.6 (0.5)\\\hline
\multirow{4}{*}{Descending FBG} & NH & 17.4 (0.3) & 17.6 (0.7) & 17.5 (0.4) & 17.4 (0.5) \\
& 1 & 40.7 (8.4) & 38.3 (2.8) & 34.4 (0.7) & 34.5 (0.7)  \\
& 2 & 17.4 (0.3) & 17.6 (0.7) & 17.5 (0.4) & 17.4 (0.5) \\
& 3 & 17.4 (0.3) & 17.6 (0.7) & 17.5 (0.4) & 17.4 (0.5) \\\hline
\multirow{4}{*}{\begin{minipage}[t]{3cm}\raggedright
    EA ascending FBG
\end{minipage}} & NH & 20.3 (0.5) & 19.9 (0.4) & 17.6 (0.3) & 17.6 (0.3) \\
& 1 & 47.1 (4.0) & 48.4 (5.3) & 42.7 (0.6) & 43.1 (0.7)  \\
& 2 & 20.3 (0.5) & 19.9 (0.4) & 17.6 (0.3) & 17.6 (0.3) \\
& 3 & 20.3 (0.5) & 19.9 (0.4) & 17.6 (0.3) & 17.6 (0.3) \\\hline
\multirow{4}{*}{\begin{minipage}[t]{3cm}\raggedright
    EA descending FBG
\end{minipage}} & NH & 20.1 (0.5) & 19.8 (0.6) & 17.8 (0.4) & 18.1 (1.5) \\
& 1 & 48.6 (6.5) & 47.9 (5.0) & 42.7 (0.7) & 42.9 (0.4)  \\
& 2 & 20.1 (0.5) & 19.8 (0.6) & 17.8 (0.4) & 18.1 (1.5) \\
& 3 & 20.1 (0.5) & 19.8 (0.6) & 17.8 (0.4) & 18.1 (1.5) \\\hline
\multirow{4}{*}{\begin{minipage}[t]{3cm}\raggedright
    EA value-to-go
\end{minipage}} & NH & 20.2 (0.4) & 19.7 (0.3) & 17.8 (0.4) & 17.4 (0.2) \\
& 1 & 45.9 (2.2) & 47.6 (4.0) & 42.5 (0.2) & 42.9 (0.3) \\
& 2 & 20.2 (0.4) & 19.7 (0.3) & 17.8 (0.4) & 17.4 (0.2) \\
& 3 & 20.2 (0.4) & 19.7 (0.3) & 17.8 (0.3) & 17.4 (0.2)\\\hline
\multirow{4}{*}{\begin{minipage}[t]{3cm}\raggedright
    EA value-to-go/visits
\end{minipage}} & NH & 20.1 (0.6) & 19.7 (0.3) & 17.8 (0.6) & 17.8 (0.4) \\
& 1 & 46.4 (3.2) & 47.8 (4.6) & 43.0 (0.4) & 43.0 (0.6)  \\
& 2 & 20.1 (0.6) & 19.7 (0.3) & 17.8 (0.6) & 17.8 (0.4) \\
& 3 & 20.1 (0.6) & 19.7 (0.3) & 17.8 (0.6) & 17.8 (0.4) \\\hline
\multirow{4}{*}{\begin{minipage}[t]{3cm}\raggedright
    EA Lagrangian$^*$
\end{minipage}} & NH & 39.2 (1.2)  & 29.3 (1.1) & 1.9 (0.1) & 1.5 (0.1) \\
& 1 & 87.7 (1.8) & 87.9 (1.1) & 86.3 (1.5) & 87.8 (1.2) \\
& 2 & 42.8 (0.5) & 42.5 (0.3) & 42.2 (0.3) & 41.7 (0.2)\\
& 3 & 42.5 (0.4) & 42.0 (0.2) & 42.3 (0.4) & 42.2 (0.3) \\\hline
\multirow{4}{*}{\begin{minipage}[t]{3cm}\raggedright
    EA Whittle index
\end{minipage}} & NH & 561.0 (23.0) & 631.8 (37.6) & 169.4 (3.4) & 119.2 (2.2) \\
& 1 & 650.7 (49.8) & 1550.5 (185.3) & 1595.0 (28.9) & 1852.2 (33.3) \\
& 2 & 750.4 (22.5) & 878.3 (52.3) & 944.6 (30.0) & 1036.1 (30.9) \\
& 3 & 597.8 (17.2) & 663.5 (13.9) & 779.4 (58.5) & 872.4 (13.4) \\\hline
\end{tabular}
\end{center}
\caption{Table showing computation times in seconds averaged across 10 replications for several and heuristics and scenarios (NH: NanoHealth). \label{table:computation-times-full-info}}
\end{table}

\subsection{Visit distribution}

In addition to the graphs shown in the full information experimental results (Sections~\ref{sec:experiments-setup-nanohealth} and \ref{sec:scen-results}), here we provide histograms showing the number of visits per patient for different heuristics.
\begin{figure}[H]
    \centering
    \includegraphics[width=\linewidth]{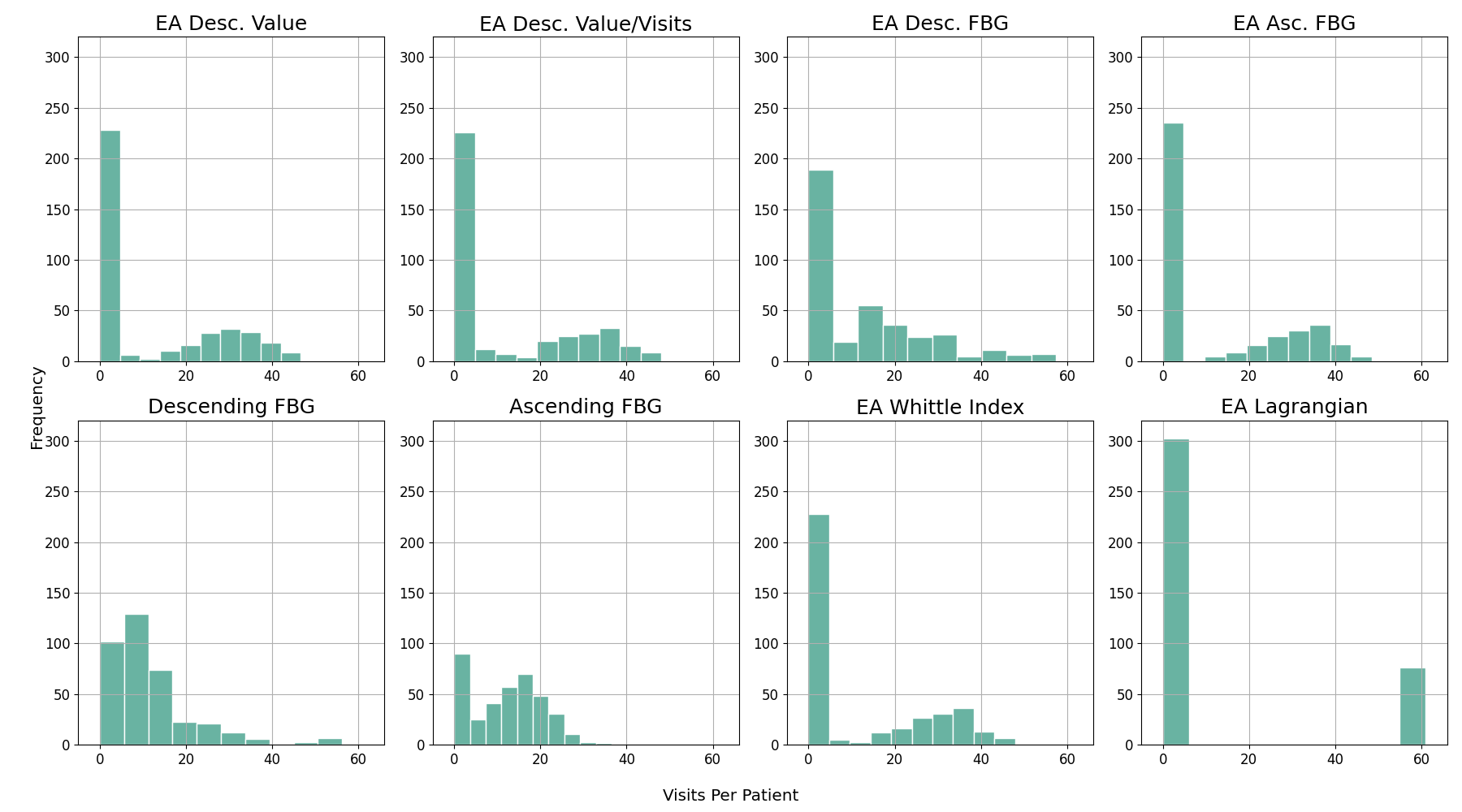}
    \caption{Distribution of the number of visits per patient for NanoHealth at 20\% capacity for several heuristics.}
    \label{fig:nanohealth-visit-dist}
\end{figure}
\begin{figure}[H]
    \centering
    \includegraphics[width=\linewidth]{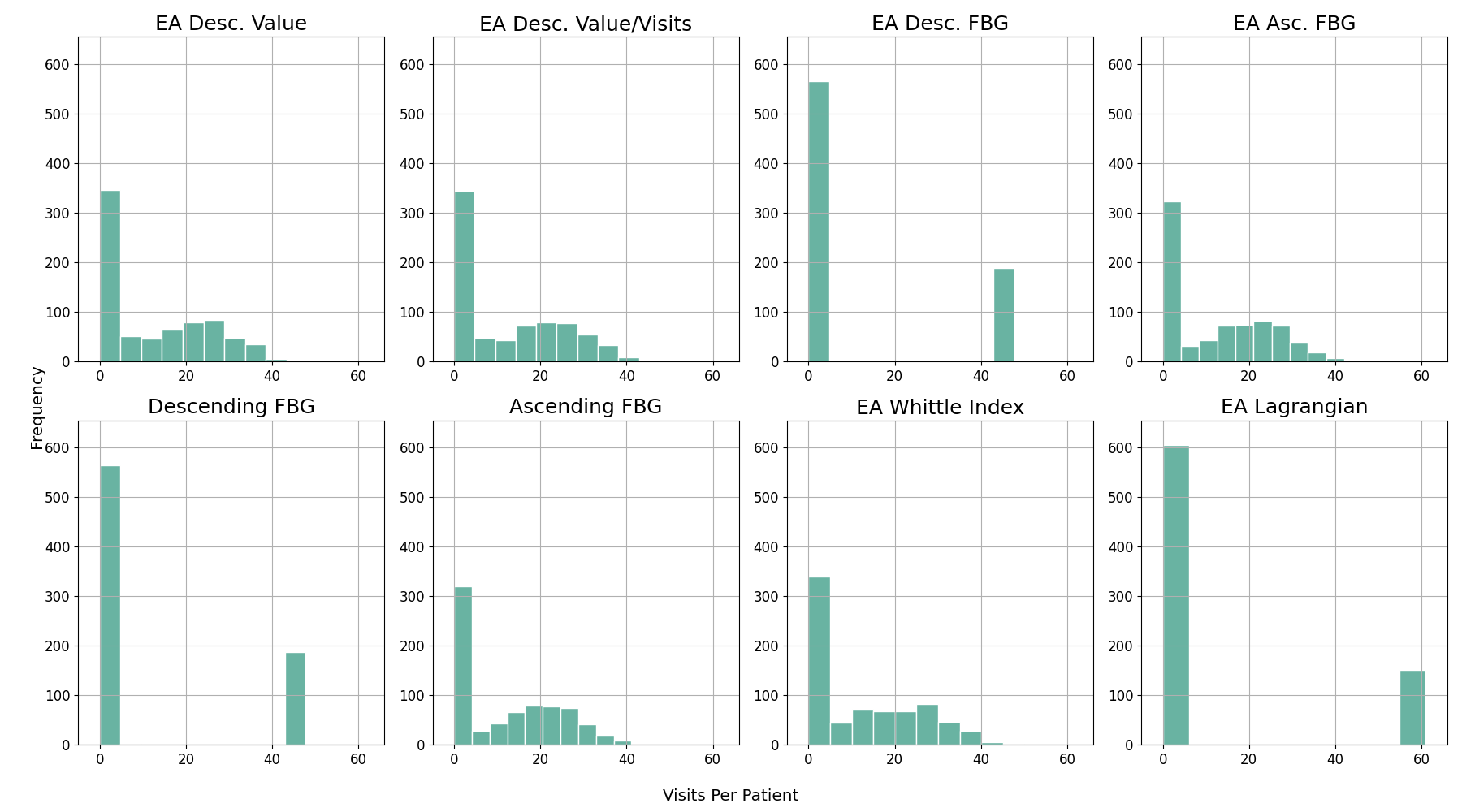}
    \caption{Distribution of the number of visits per patient for Scenario 1 at 20\% capacity for several heuristics.}
    \label{fig:scenario1-visit-dist-cap151}
\end{figure}
\begin{figure}[H]
    \centering
    \includegraphics[width=\linewidth]{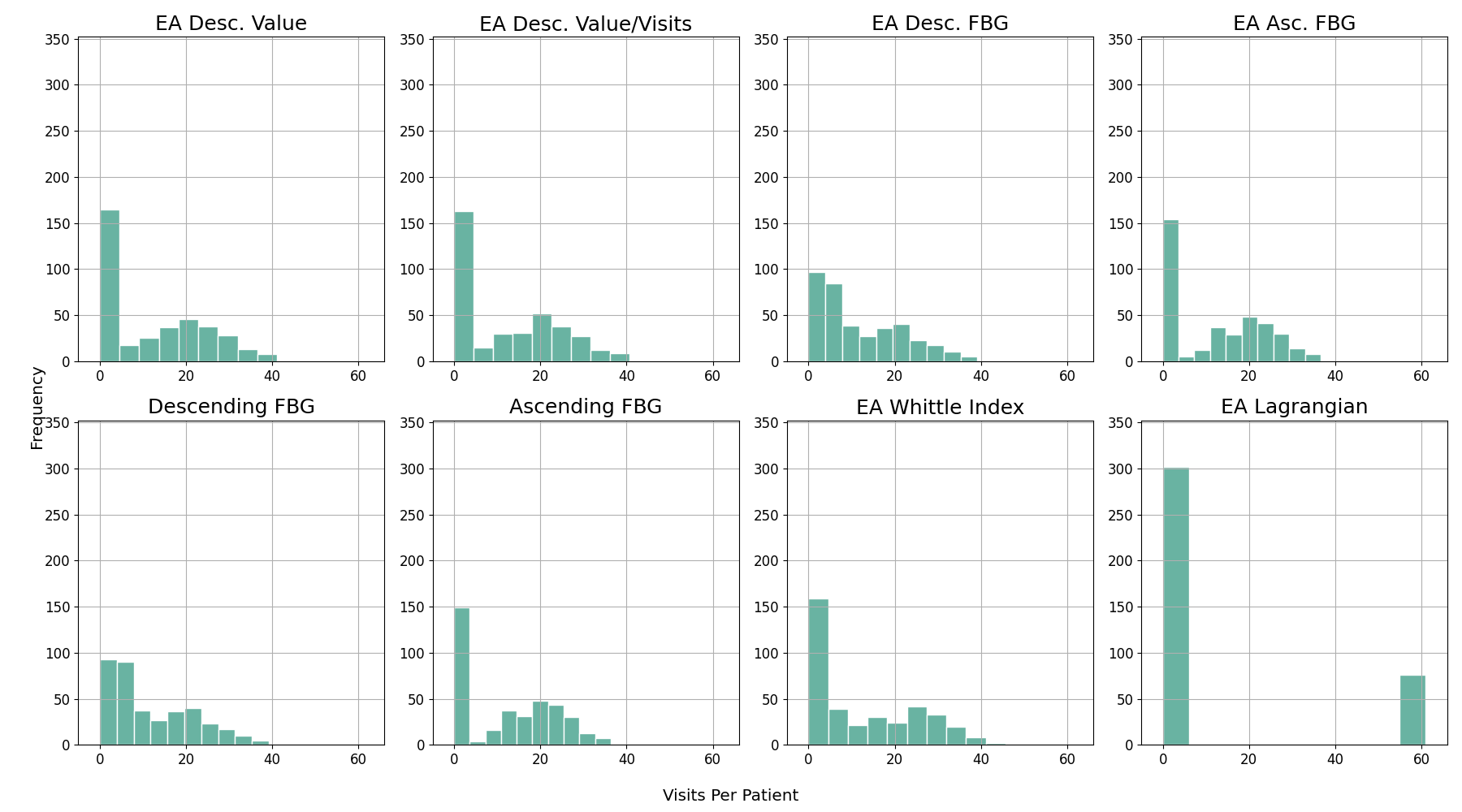}
    \caption{Distribution of the number of visits per patient for Scenario 2 at 20\% capacity for several heuristics.}
    \label{fig:scenario2-visit-dist-cap76}
\end{figure}
\begin{figure}[H]
    \centering
    \includegraphics[width=\linewidth]{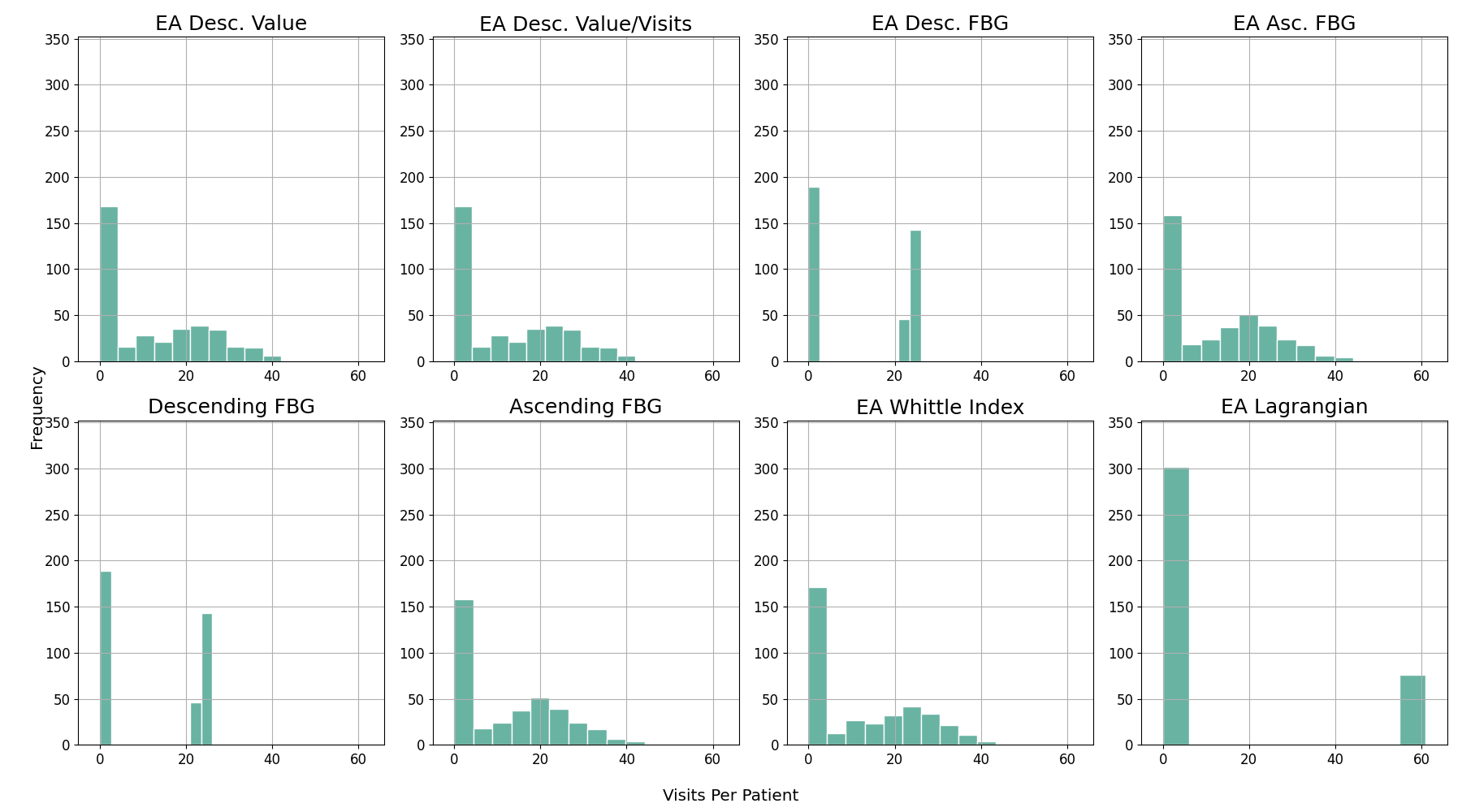}
    \caption{Distribution of the number of visits per patient for Scenario 3 at 20\% capacity for several heuristics.}
    \label{fig:scenario3-visit-dist-cap76}
\end{figure}
\begin{figure}[H]
    \centering
    \includegraphics[width=\linewidth]{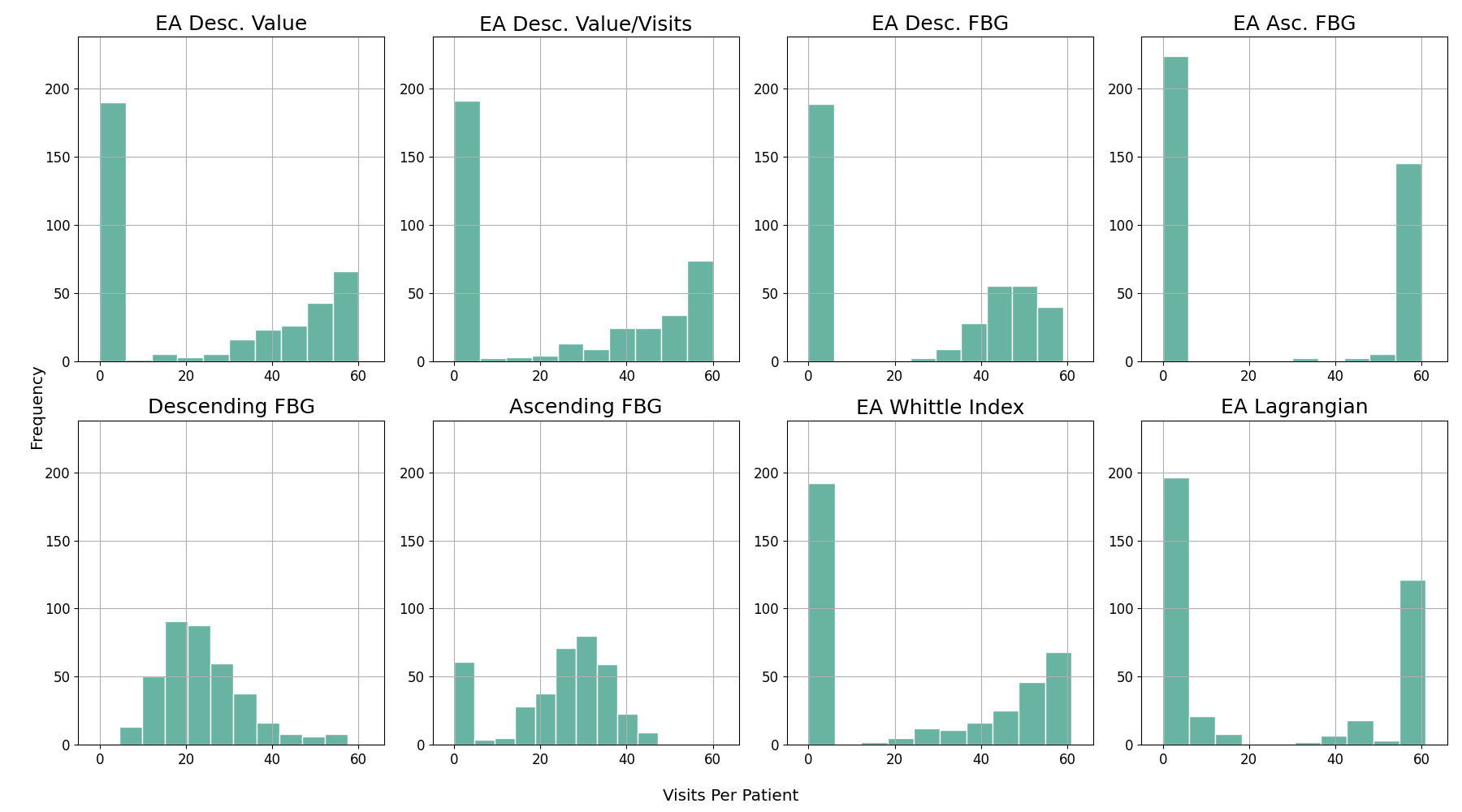}
    \caption{Distribution of the number of visits per patient for NanoHealth at 40\% capacity for several heuristics.}
    \label{fig:nanohealth-visit-dist-cap76}
\end{figure}
\begin{figure}[H]
    \centering
    \includegraphics[width=\linewidth]{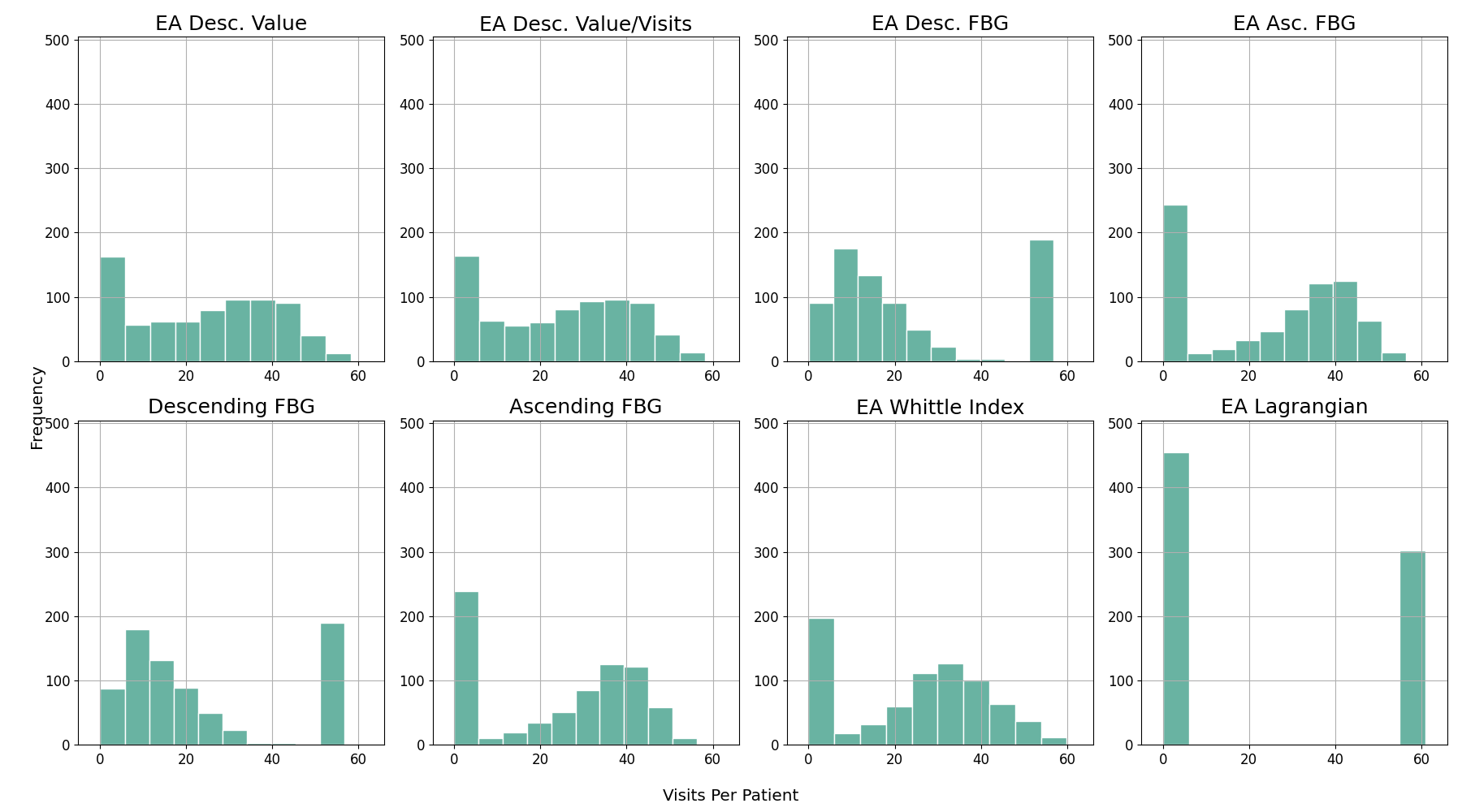}
    \caption{Distribution of the number of visits per patient for Scenario 1 at 40\% capacity for several heuristics.}
    \label{fig:scenario1-visit-dist-cap302}
\end{figure}
\begin{figure}[H]
    \centering
    \includegraphics[width=\linewidth]{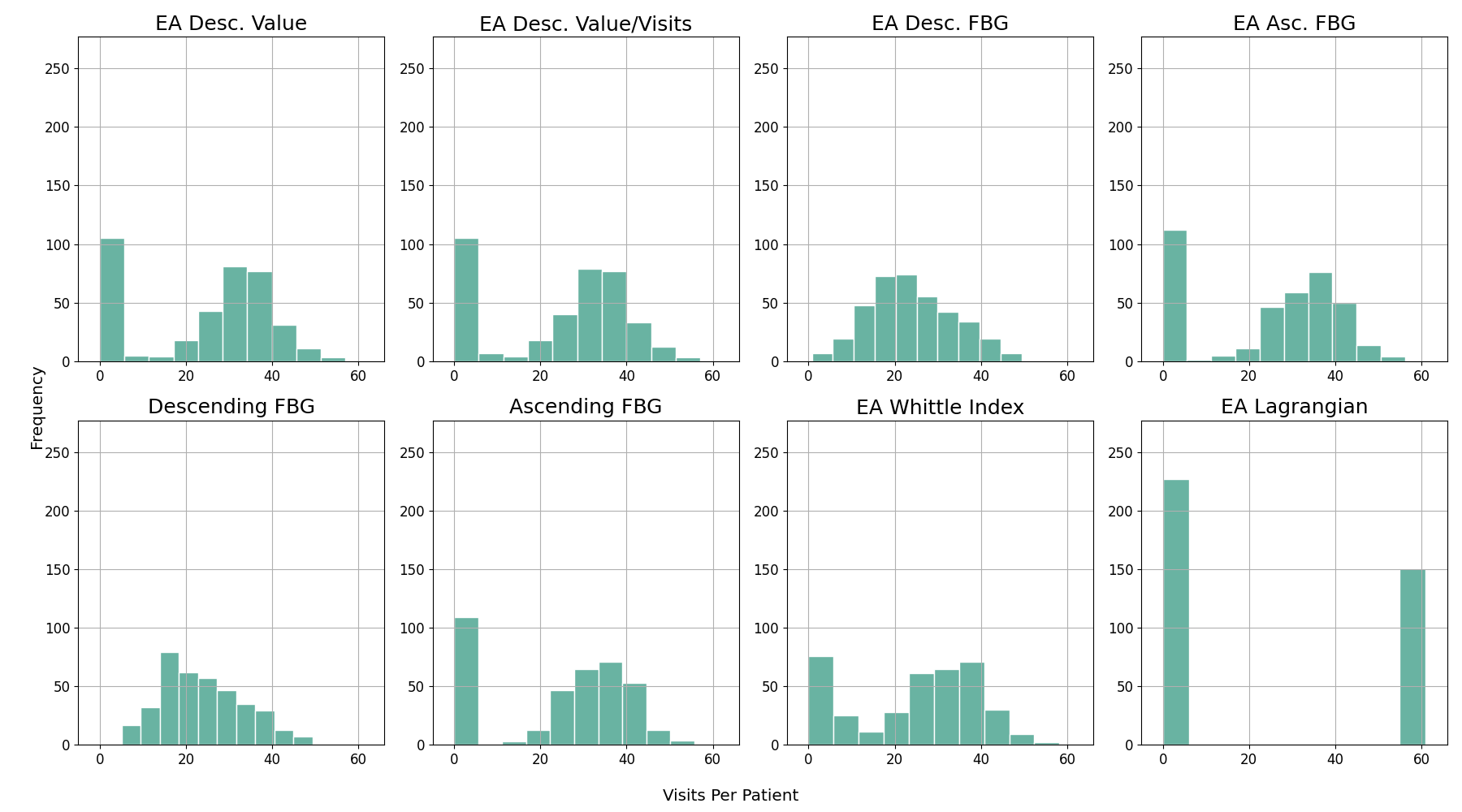}
    \caption{Distribution of the number of visits per patient for Scenario 2 at 40\% capacity for several heuristics.}
    \label{fig:scenario2-visit-dist-cap151}
\end{figure}
\begin{figure}[H]
    \centering
    \includegraphics[width=\linewidth]{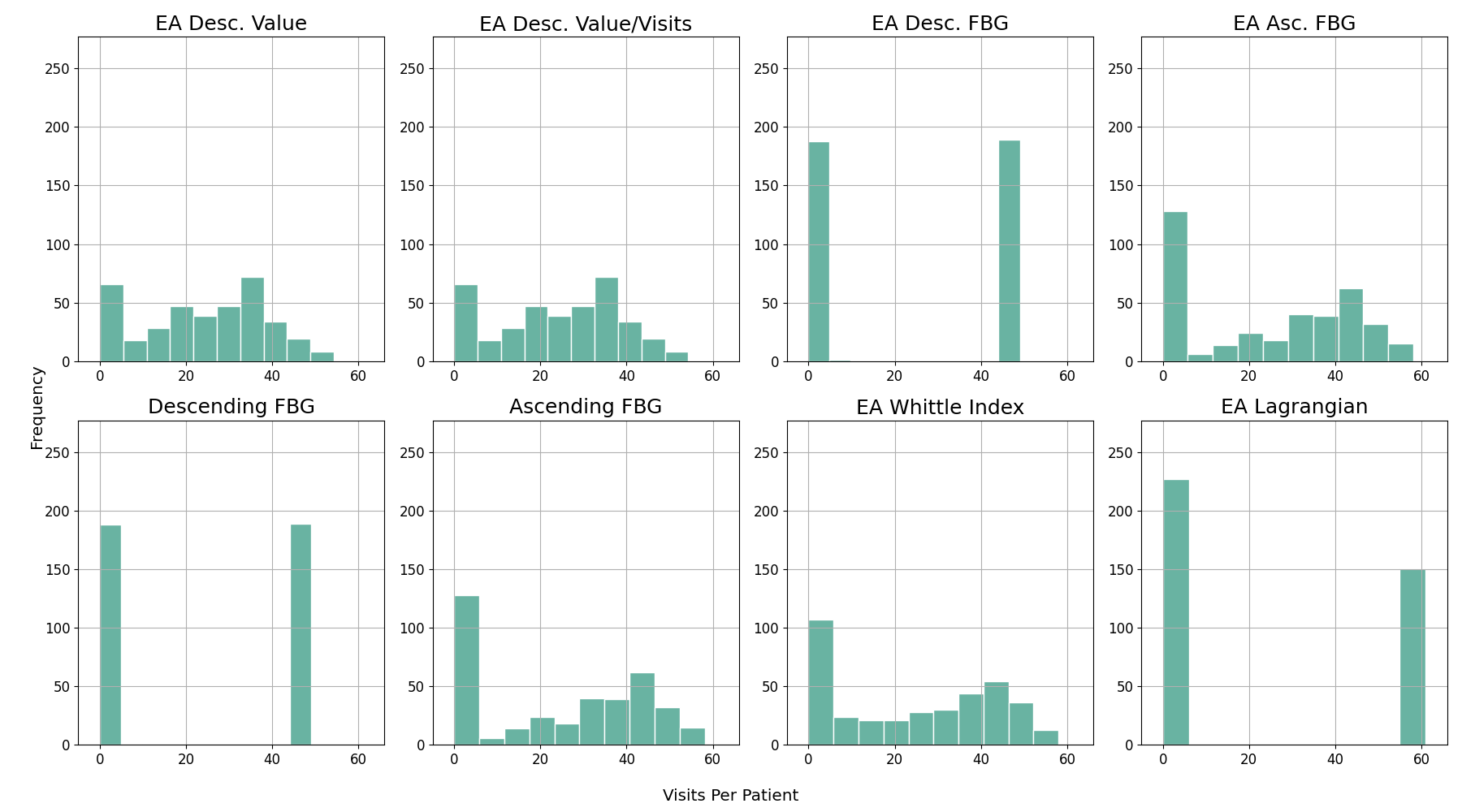}
    \caption{Distribution of the number of visits per patient for Scenario 3 at 40\% capacity for several heuristics.}
    \label{fig:scenario3-visit-dist-cap151}
\end{figure}
\begin{figure}[H]
    \centering
    \includegraphics[width=\linewidth]{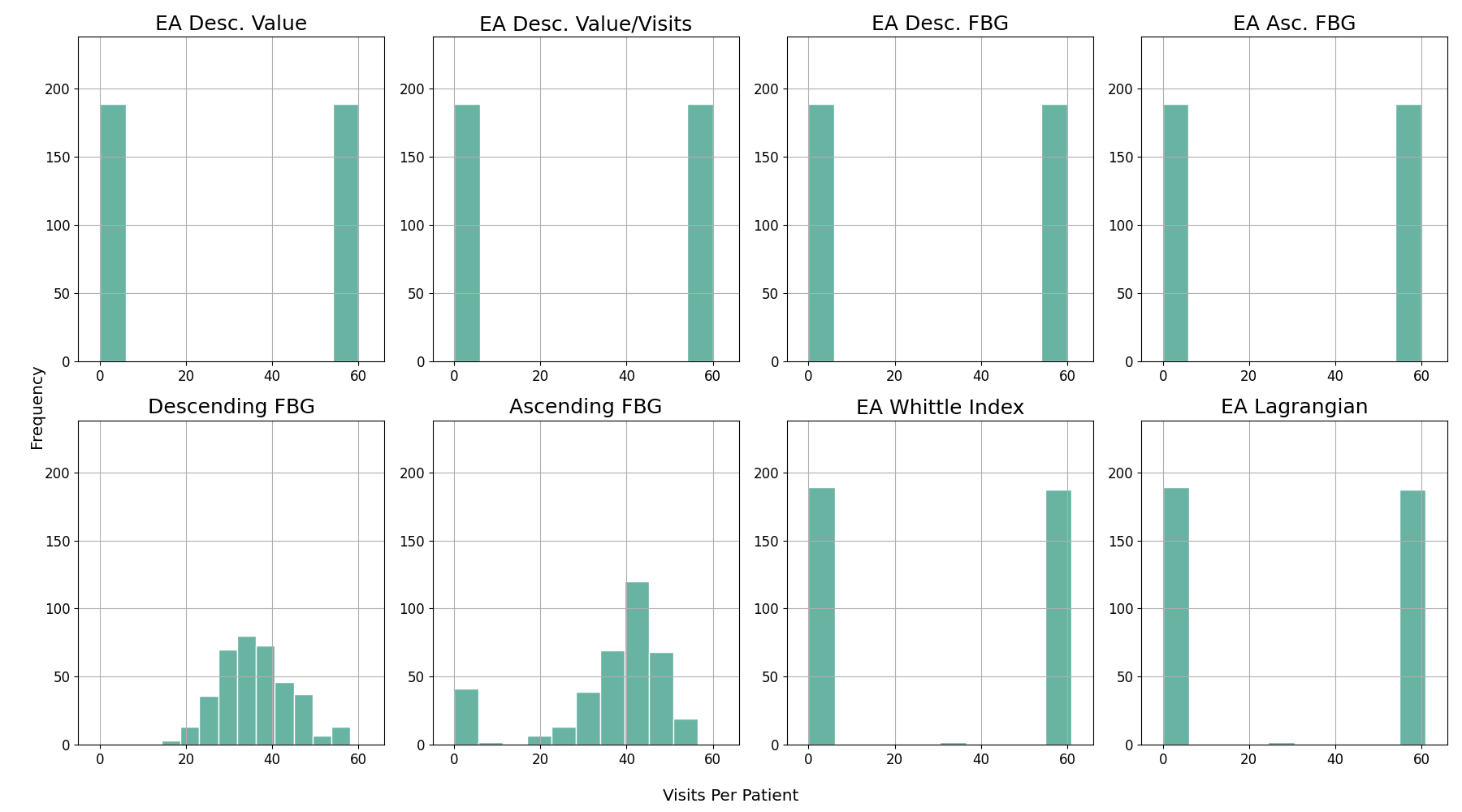}
    \caption{Distribution of the number of visits per patient for NanoHealth at 60\% capacity for several heuristics.}
    \label{fig:nanohealth-visit-dist-cap227}
\end{figure}
\begin{figure}[H]
    \centering
    \includegraphics[width=\linewidth]{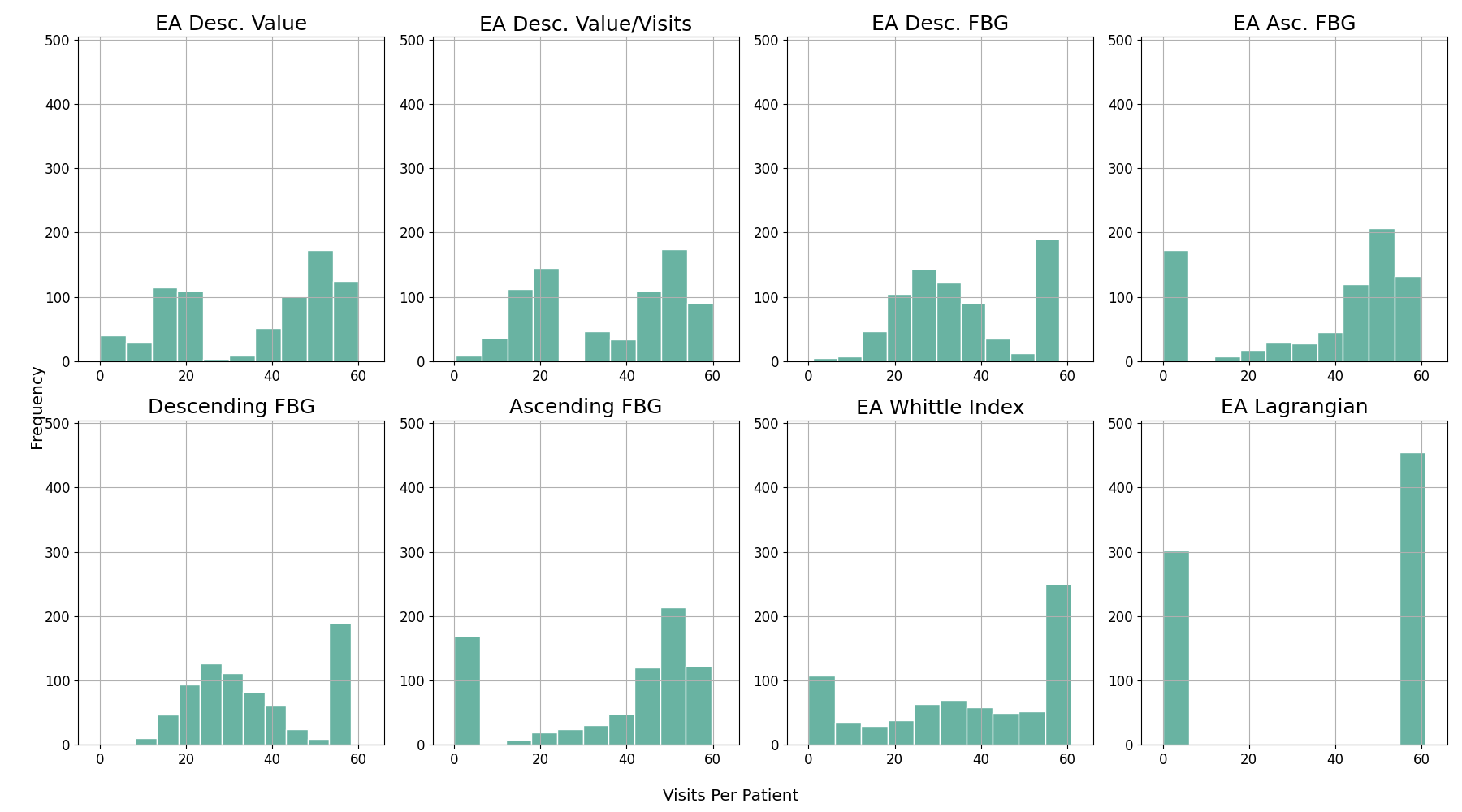}
    \caption{Distribution of the number of visits per patient for Scenario 1 at 60\% capacity for several heuristics.}
    \label{fig:scenario1-visit-dist-cap454}
\end{figure}
\begin{figure}[H]
    \centering
    \includegraphics[width=\linewidth]{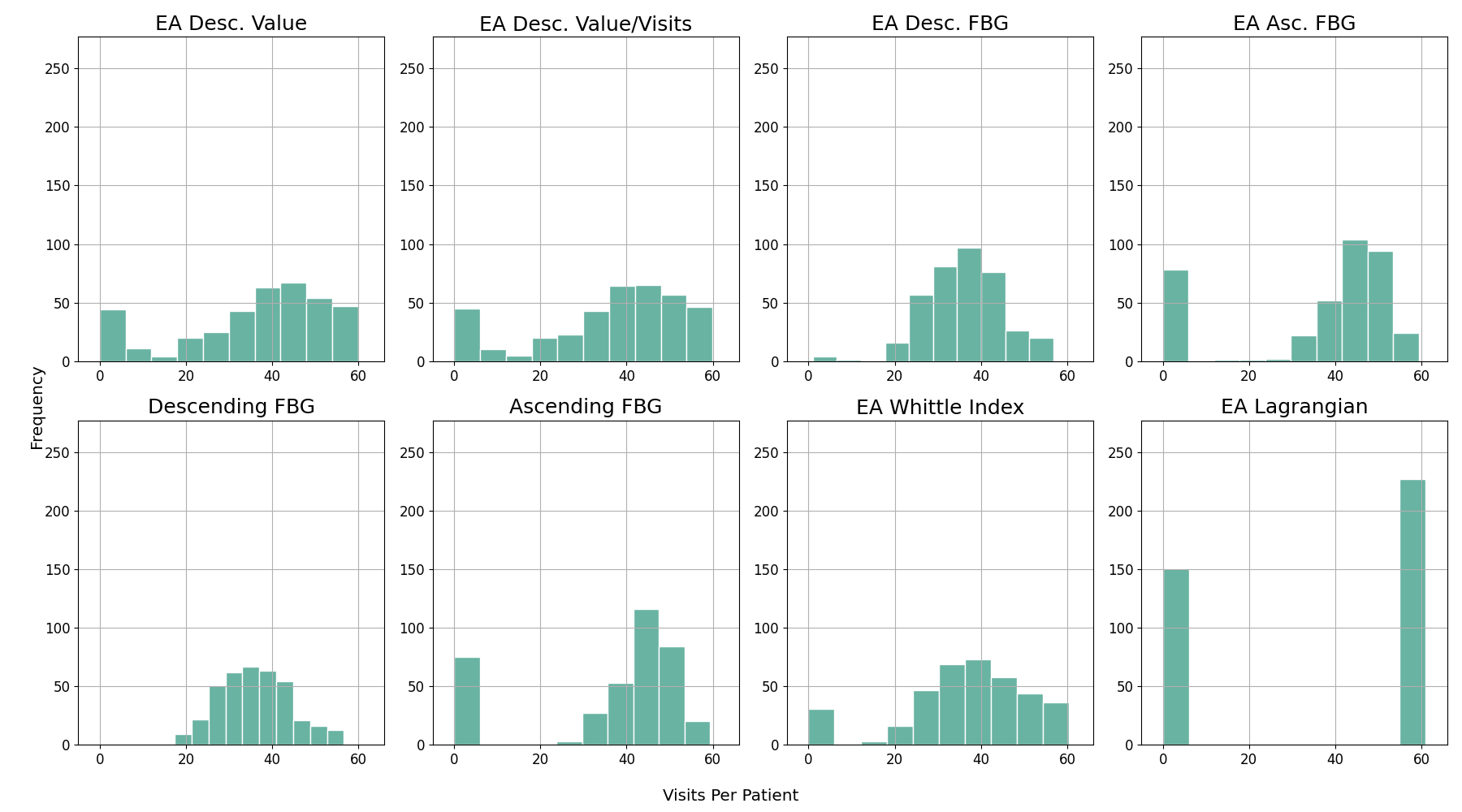}
    \caption{Distribution of the number of visits per patient for Scenario 2 at 60\% capacity for several heuristics.}
    \label{fig:scenario2-visit-dist-cap227}
\end{figure}
\begin{figure}[H]
    \centering
    \includegraphics[width=\linewidth]{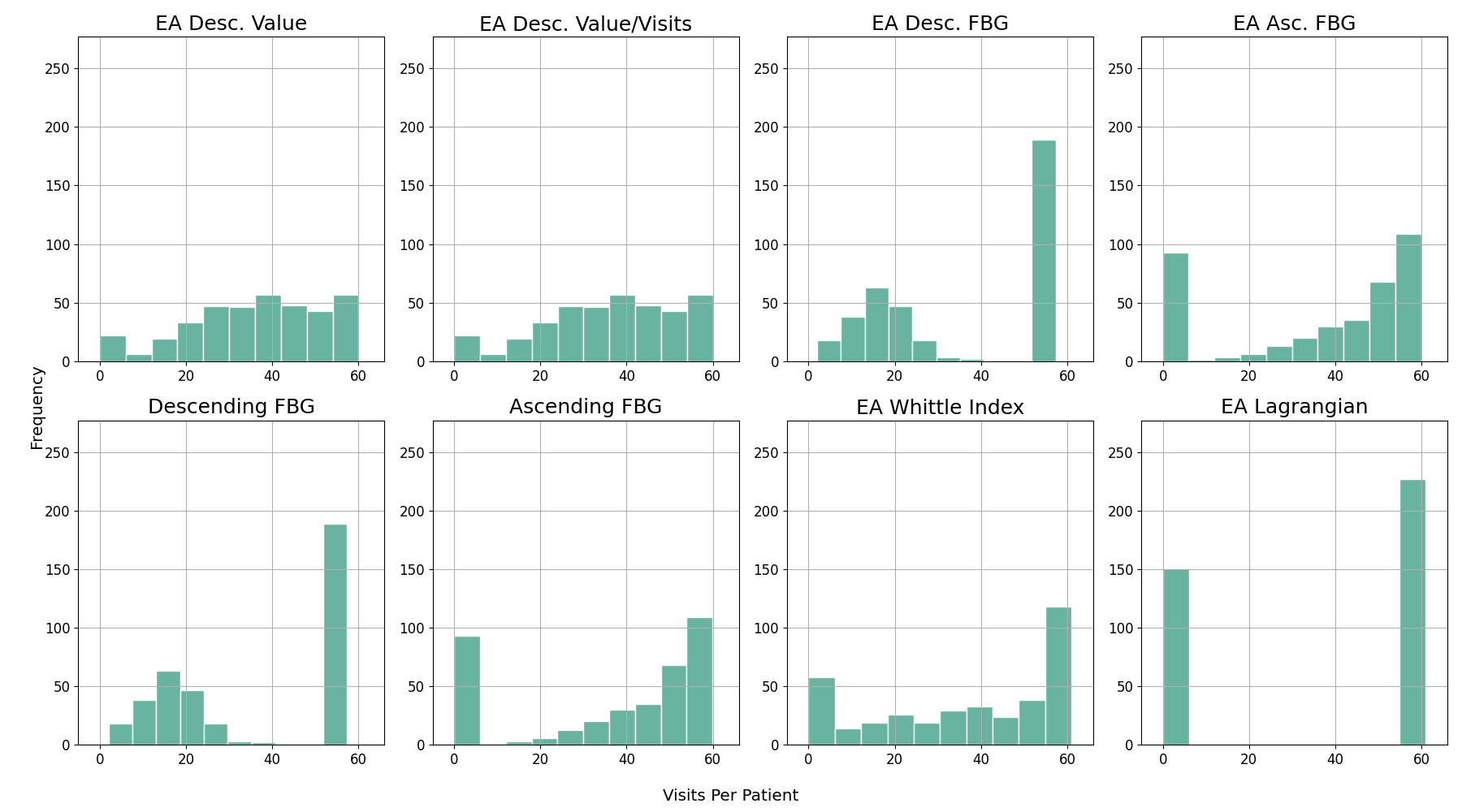}
    \caption{Distribution of the number of visits per patient for Scenario 3 at 60\% capacity for several heuristics.}
    \label{fig:scenario3-visit-dist-cap227}
\end{figure}
\begin{figure}[H]
    \centering
    \includegraphics[width=\linewidth]{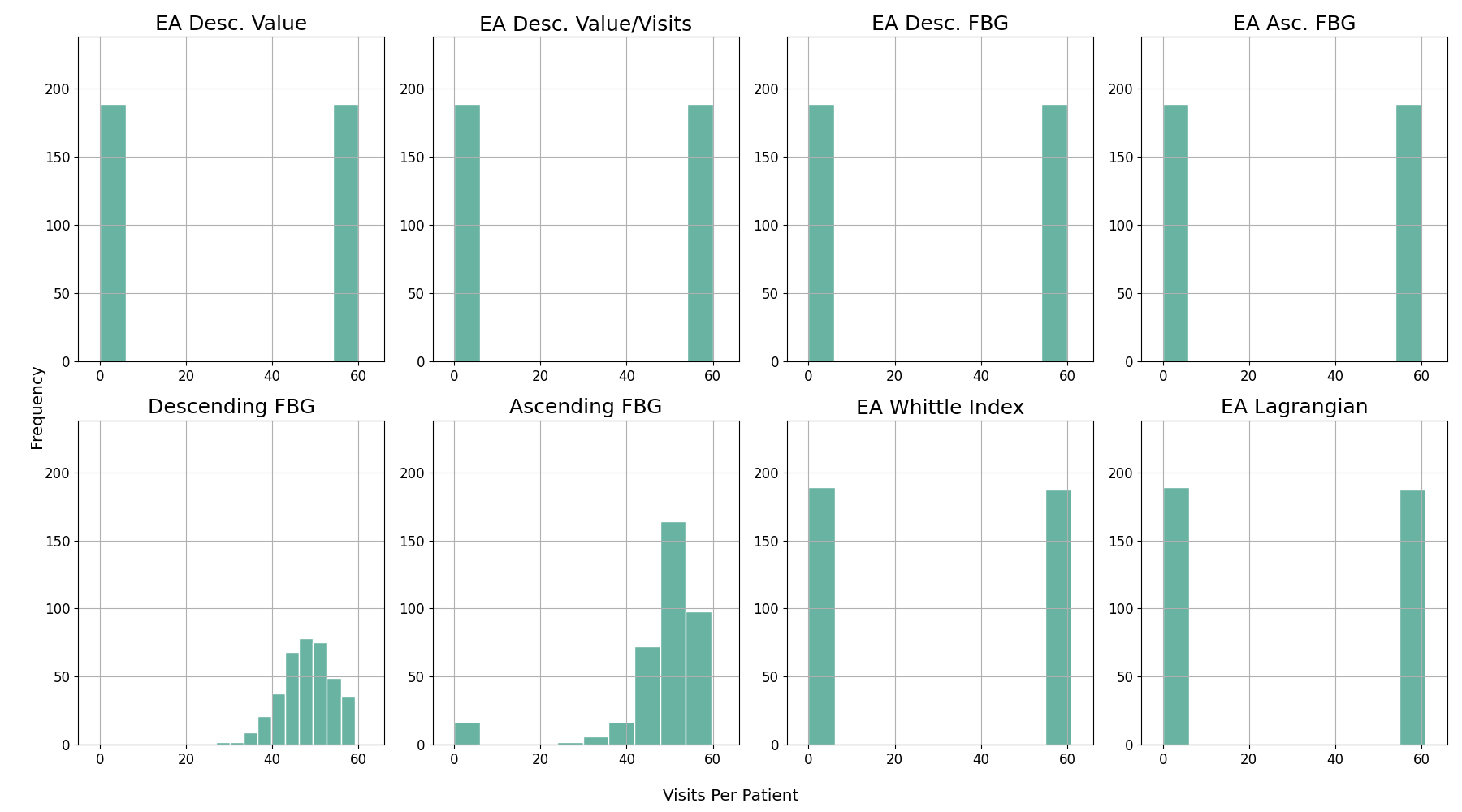}
    \caption{Distribution of the number of visits per patient for NanoHealth at 80\% capacity for several heuristics.}
    \label{fig:nanohealth-visit-dist-cap302}
\end{figure}
\begin{figure}[H]
    \centering
    \includegraphics[width=\linewidth]{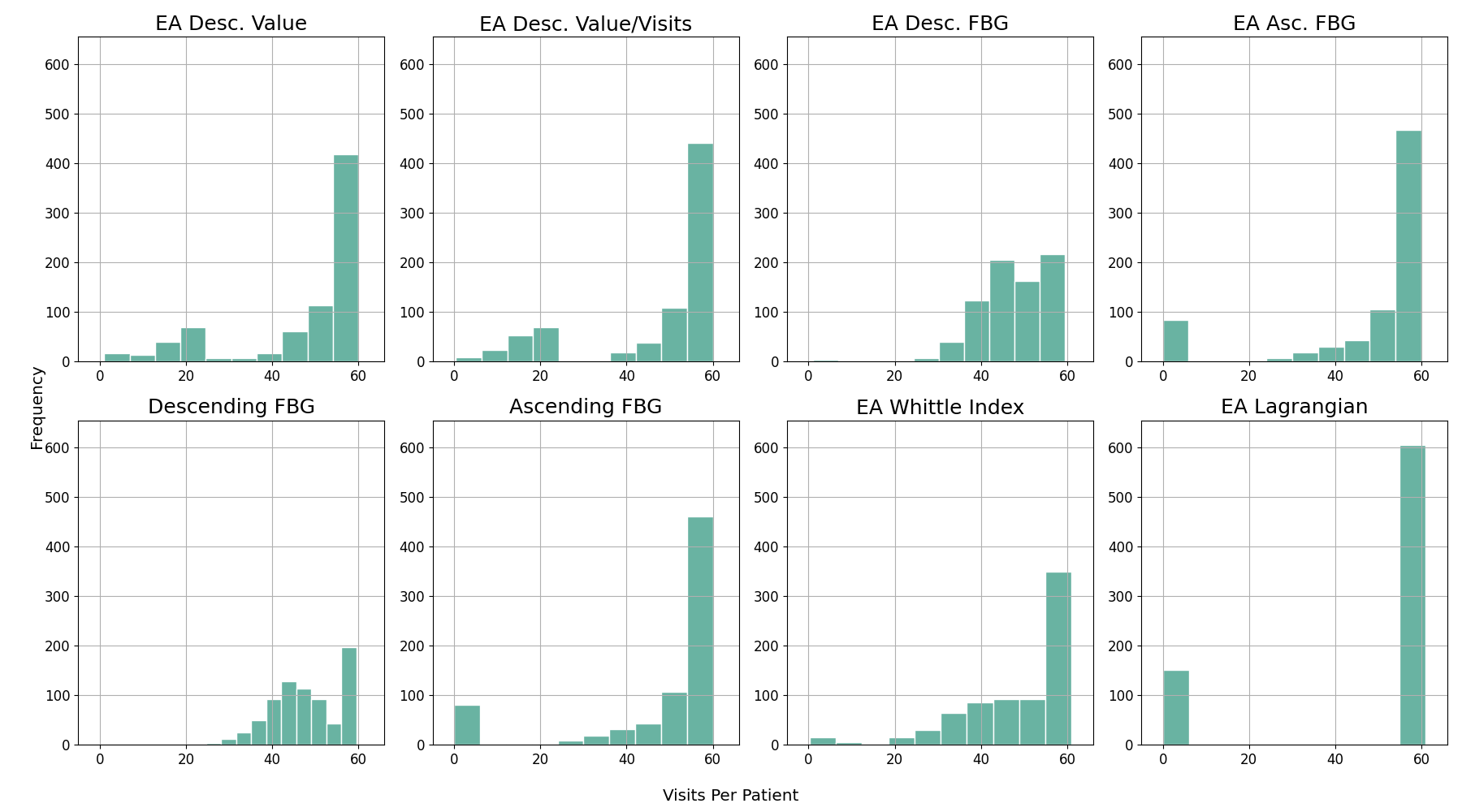}
    \caption{Distribution of the number of visits per patient for Scenario 1 at 80\% capacity for several heuristics.}
    \label{fig:scenario1-visit-dist-cap605}
\end{figure}
\begin{figure}[H]
    \centering
    \includegraphics[width=\linewidth]{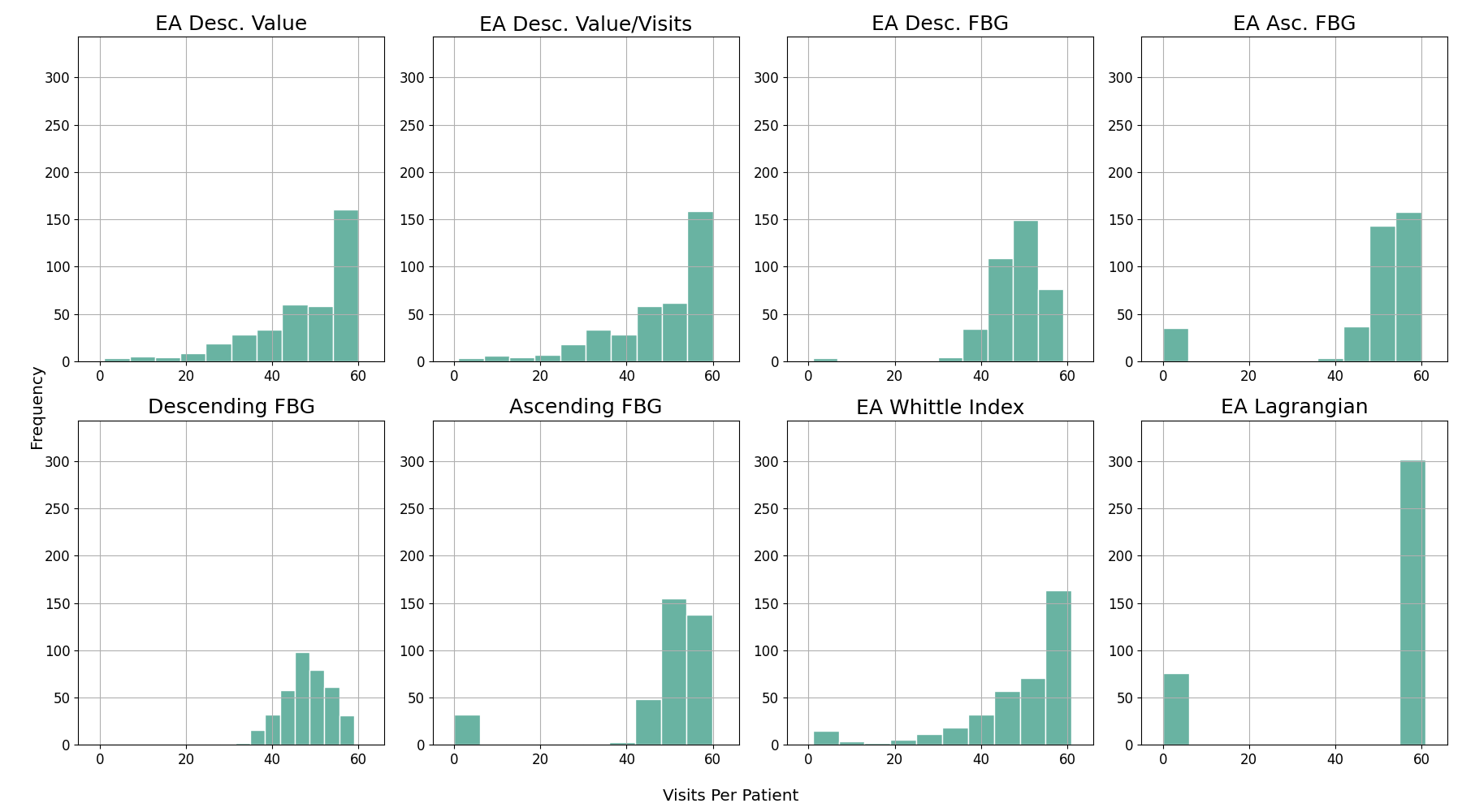}
    \caption{Distribution of the number of visits per patient for Scenario 2 at 80\% capacity for several heuristics.}
    \label{fig:scenario2-visit-dist-cap302}
\end{figure}
\begin{figure}[H]
    \centering
    \includegraphics[width=\linewidth]{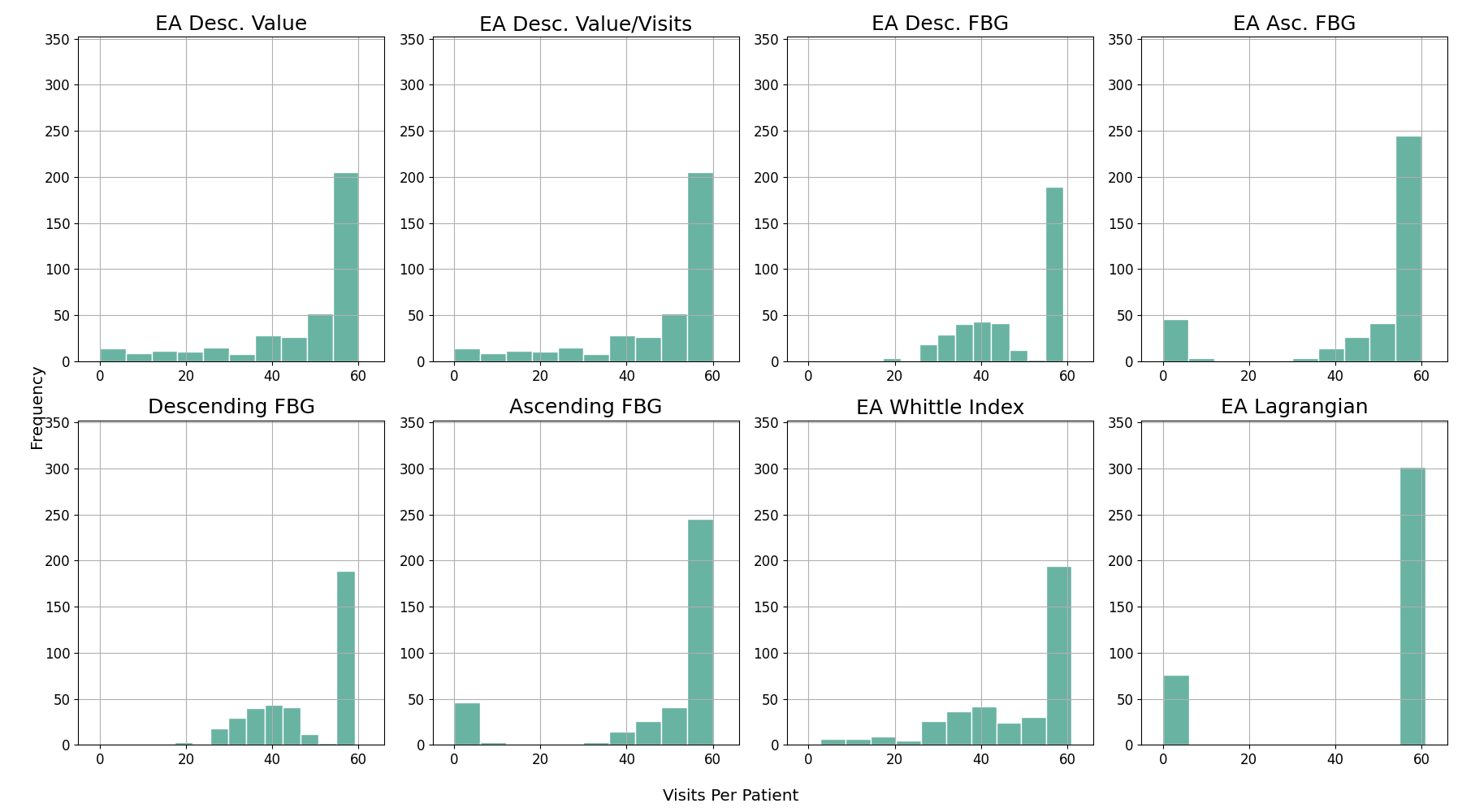}
    \caption{Distribution of the number of visits per patient for Scenario 3 at 80\% capacity for several heuristics.}
    \label{fig:scenario3-visit-dist-cap302}
\end{figure}

\section{Machine learning experiments}\label{EC:mlexperiments}

\subsection{In control prediction}\label{EC:incontrolprediction}

The goal of these experiments is to compare the prediction performance of the MLE problem~\eqref{eq:mle_prob} in Proposition~\ref{prop:mle_form} for estimating FBG \emph{in control} probabilities with estimates from standard machine learning models. We compare \eqref{eq:mle_prob} with two models: logistic regression and random forest. We provide the machine learning models with five features: whether or not the patient was visited, whether or not the patient is enrolled in treatment, the number of days since the last visit, the total number of visits so far, and the average of all previous FBG measurements. All features were scaled to the $[0,1]$ interval. We evaluate the performance using leave-one-out cross validation and we use both the ROC curve and the corresponding AUC metric to quantify performance. For each test set patient, we predict the probability of \emph{in control} for each visit in their history since these are the only periods where we have a corresponding FBG measurement that can be used as the target. Note that the length of the history is different for each patient. For each machine learning model, we conduct five-fold cross-validation using the training data to select the best hyperparameters. For logistic regression, we consider both L1 and L2 regularization and six values for the penalty parameter: $[0.001, 0.01, 0.1, 1, 10, 100]$. For random forest, we use 250 trees and consider four max depth parameter values: $[\textnormal{None}, 2, 5, 10]$.

The MLE estimates for \emph{in control} probabilities were obtained by building profile likelihood curves for each patient and visit. The profile likelihood curves were built by discretizing the full range of FBG values from past simulations into 50 values and re-solving \eqref{eq:mle_prob} for each new visit in their history (including the history up to each visit) conditioned on the FBG in the following period being equal to each one of the 50 values. We then normalized the likelihood values to resemble probability density functions and integrated them up to the value $\delta$ using Riemann sums, leading us to obtain \emph{in control} probability estimates for each patient for each visit.

\begin{figure}[ht]
\centering
\includegraphics[width=0.45\textwidth]{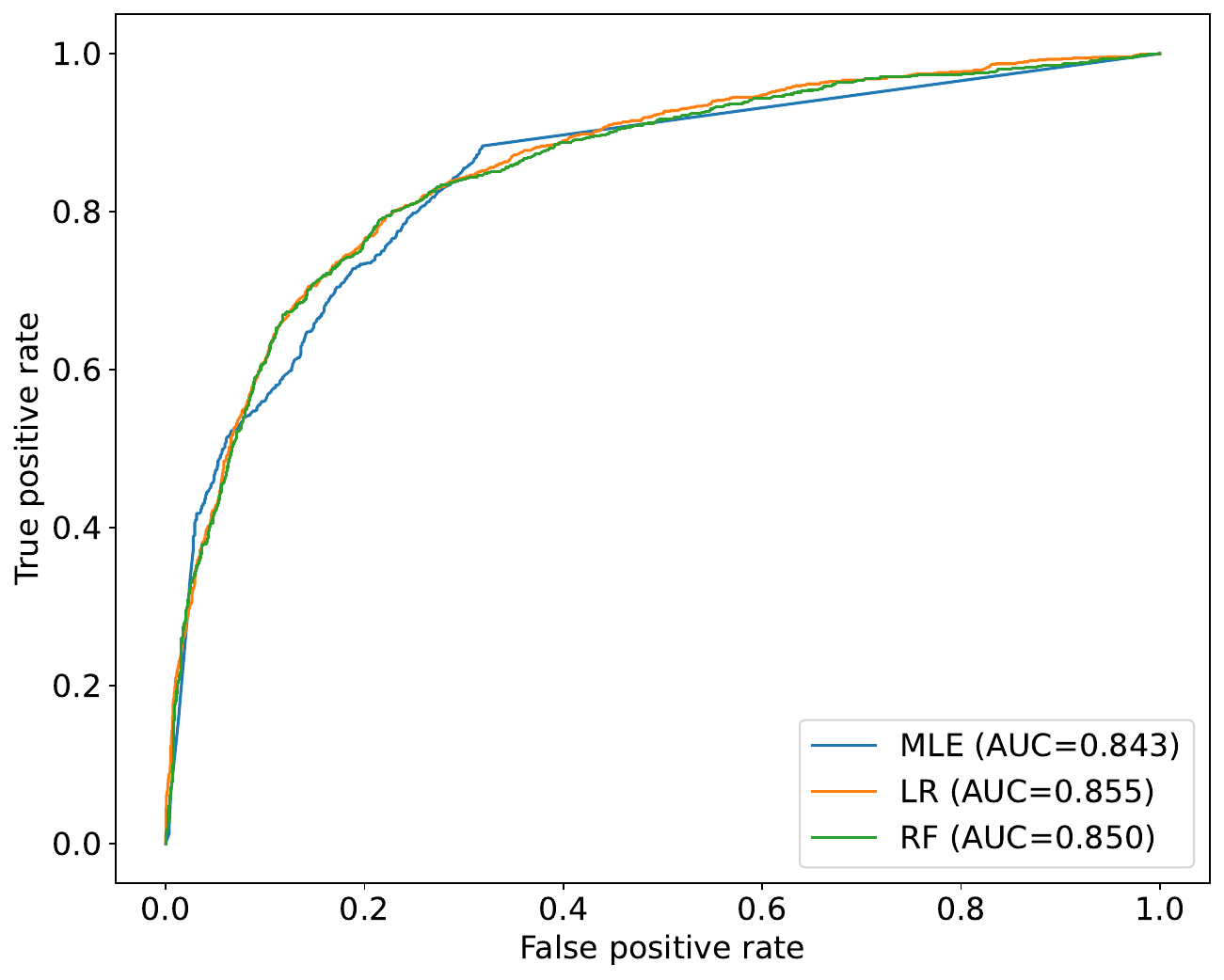}
\caption{Out-of-sample ROC curves and corresponding AUC values for logistic regression (LR), random forest (RF), and the MLE estimates. \label{fig:ML_incontrol}} 
\end{figure}

Figure~\ref{fig:ML_incontrol} displays the ROC curve for each model obtained through leave-one-out cross validation. We obtained these curves by grouping all out-of-sample patient-periods. The best performing model was the logistic regression with an AUC of 0.855. Both random forest and the MLE estimates performed similarly with AUC scores of 0.850 and 0.843, respectively.

\begin{figure}[ht]
\centering
\includegraphics[width=0.45\textwidth]{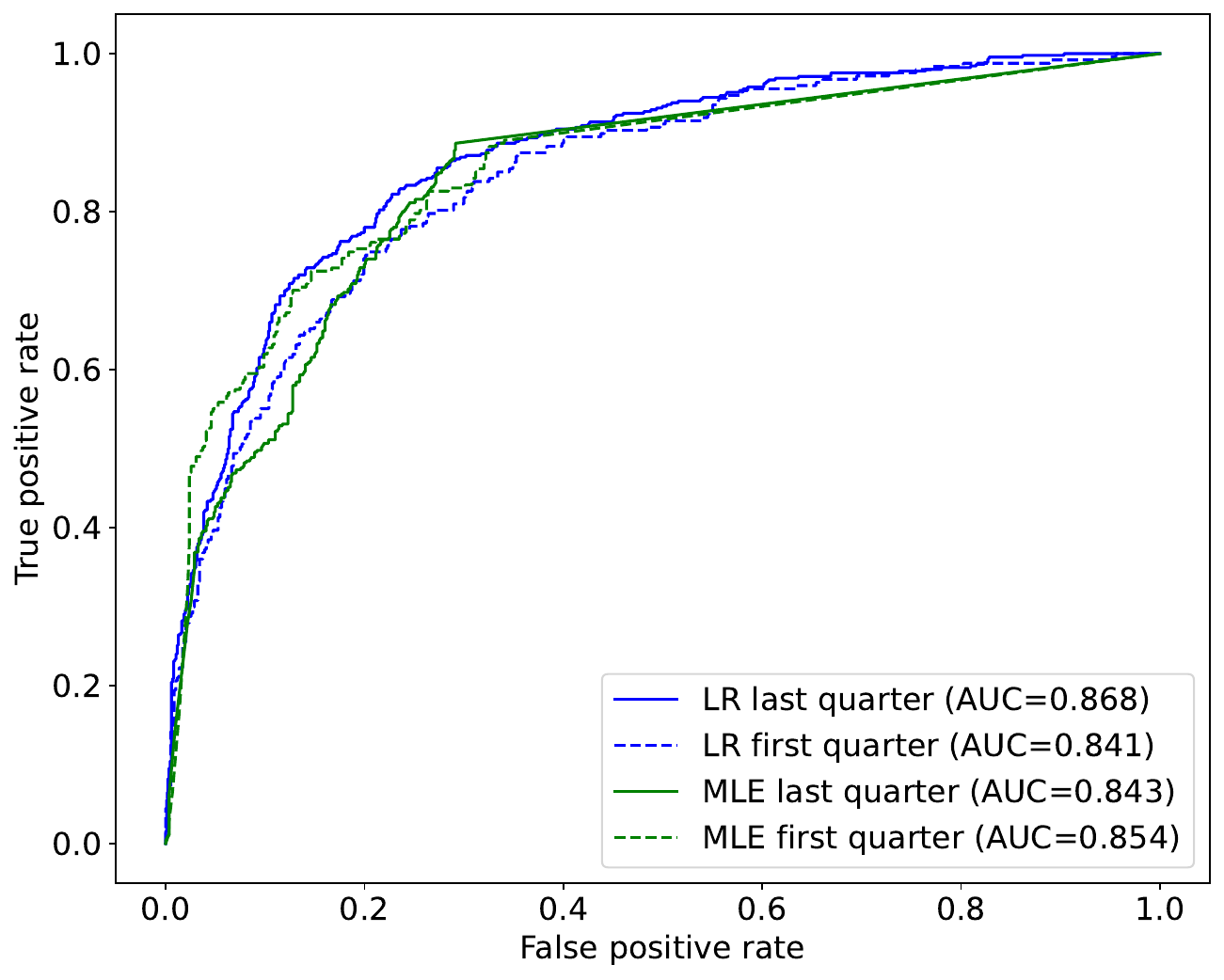}
\caption{Out-of-sample ROC curves and corresponding AUC values for logistic regression (LR) and the MLE estimates for the first and last quarter of each patients time horizon. \label{fig:ML_incontrol2}}
\end{figure}

Figure~\ref{fig:ML_incontrol2} displays the ROC curves for the first and last quarter of the time horizon for the logistic regression and MLE estimates. The performance of the logistic regression model improved slightly over time (from 0.841 to 0.868), likely due to an increase in information captured by the features. In contrast, the performance of the MLE estimates degraded slightly over time. from 0.854 to 0.843. This suggests that the MLE model may perform best in data-limited settings (such as the launch of a program) but that modern machine learning methods may add value once there is enough data.

Overall, we find that the MLE estimates perform similarly to modern machine learning methods. This is an encouraging result because it shows a comparable performance of an estimation procedure that can be directly integrated into an optimization framework for visit planning and that has theoretical guarantees on the identifiability and consistency of the estimated parameters.

\subsection{Clustering and cluster prediction} \label{EC:clusteringprediction}

The purpose of these experiments is to demonstrate that is may be possible to obtain initial estimates for patient parameters using machine learning. To do this, we first cluster the NanoHealth patients according to the parameters returned by the MLE. Then we assign each cluster a class label and develop machine learning models to predict those class labels using basic information commonly obtained during a screening visit.

\subsubsection{Clustering.}

\begin{figure}[ht]
\centering
\includegraphics[width=0.45\textwidth]{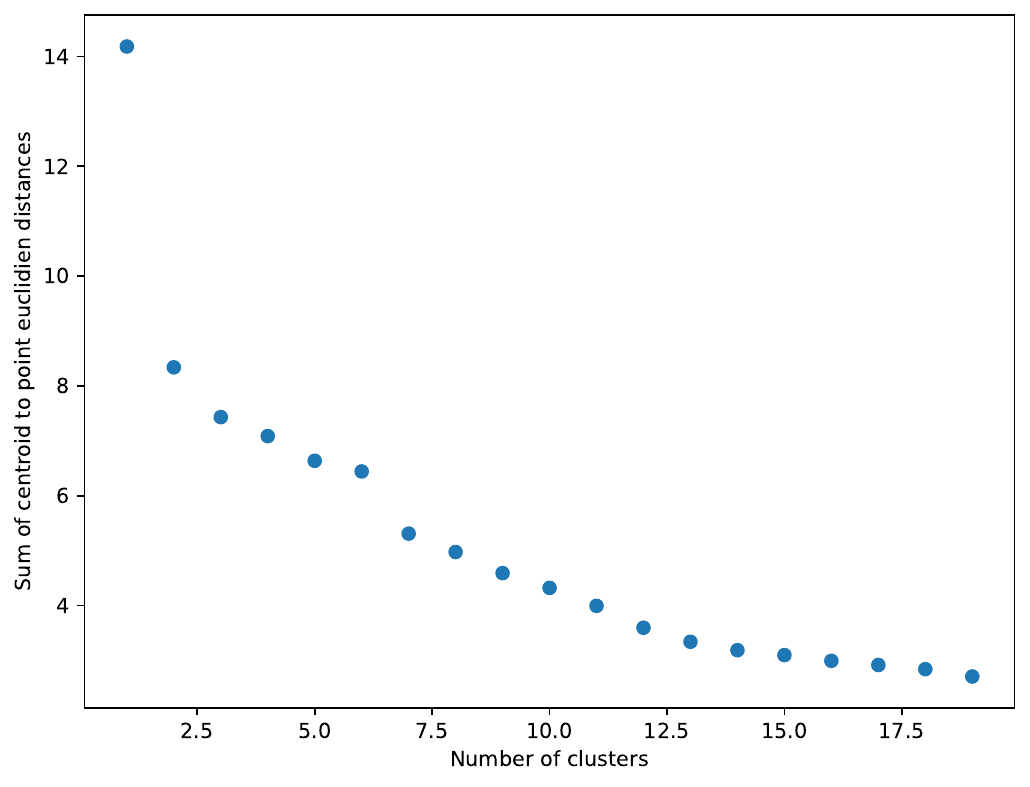}
\caption{Elbow plot, showing the sum of the distances to the centroids against the total number of clusters. \label{fig:ML_cluster}} 
\end{figure}

We cluster patients according to the 11 parameters returned by the MLE procedure: ($p,\mu,\alpha,\theta,\lambda,s_0,\beta,\gamma,\rho,b,z$). All features were scaled to the $[0,1]$ interval. We conducted K-means clustering with Euclidean distance as the metric. We considered cluster sizes from 1 to 20, and computed the total distance between each observation and it's assigned centroid for each cluster size. Figure~\ref{fig:ML_cluster} displays the total distance as a function of the number of clusters. Based on this figure, we chose five clusters corresponding to five patient types (as a trade-off between the number of groups and prediction accuracy). The cluster centroids for five clusters and the number of patients in each cluster are shown in Table~\ref{tab:ML_centroid}. The cluster centroids for seven clusters and the number of patients in each cluster are shown in Table~\ref{tab:ML_centroid2}.

\renewcommand{\arraystretch}{1.3}
\begin{table}
\caption{Cluster centroids for five clusters and the number of patients in each cluster.}
\label{tab:ML_centroid}
\centering
\resizebox{\textwidth}{!}{
\begin{tabular}{ccccccccccccr}
\hline
Class & Size & $p$ & $\mu$ & $\alpha$ & $\theta$ & $\lambda$ & $s_0$ & $\beta$ & $\gamma$ & $\rho$ & $b$ & $z$ \\
\hline
1 & 16 & 0.0002 & 0.0039 & 0.0006 & 0.1559 & 0.1323 & 0.0208 & 0.3333 & 0.8884 & 0.2848 & 0.4787 & 0.0000 \\
2 & 90 & 0.9992 & 0.9856 & 0.0178 & 0.0000 & -0.0000 & -0.0000 & 0.0037 & 0.0000 & 0.0478 & 0.4623 & 0.0000 \\
3 & 115 & 0.0108 & 0.0012 & 0.0147 & 0.0273 & 0.0182 & 0.0261 & 0.0928 & 0.0000 & 0.0483 & 0.6575 & 0.0000 \\
4 & 7 & 0.9986 & 0.0015 & 0.9996 & 0.0000 & 0.0000 & 0.0000 & 0.0000 & 0.0000 & 0.0000 & 0.4825 & 0.0000 \\
5 & 150 & 0.0111 & 0.0005 & 0.0123 & 0.0065 & 0.0016 & 0.0200 & 0.1244 & 0.0000 & 0.0076 & 0.3113 & 0.0000 \\
\hline
\end{tabular}
}
\end{table}

\renewcommand{\arraystretch}{1.3}
\begin{table}
\caption{Cluster centroids for seven clusters and the number of patients in each cluster.}
\label{tab:ML_centroid2}
\centering
\resizebox{\textwidth}{!}{
\begin{tabular}{ccccccccccccr}
\hline
Class & Size & $p$ & $\mu$ & $\alpha$ & $\theta$ & $\lambda$ & $s_0$ & $\beta$ & $\gamma$ & $\rho$ & $b$ & $z$ \\
\hline
1 & 101 & 0.0129 & 0.0005 & 0.0141 & 0.0045 & 0.0003 & 0.0132 & -0.0000 & 0.0000 & 0.0075 & 0.3095 & 0.0000 \\
2 & 90 & 0.9992 & 0.9856 & 0.0178 & 0.0000 & -0.0000 & -0.0000 & 0.0037 & 0.0000 & 0.0478 & 0.4623 & 0.0000 \\
3 & 67 & 0.0076 & 0.0006 & 0.0091 & 0.0000 & -0.0000 & -0.0000 & 0.3333 & 0.0000 & 0.0000 & 0.3883 & 0.0000 \\
4 & 7 & 0.9986 & 0.0015 & 0.9996 & 0.0000 & 0.0000 & 0.0000 & 0.0000 & 0.0000 & 0.0000 & 0.4825 & 0.0000 \\
5 & 14 & 0.0039 & 0.0009 & 0.0095 & 0.2608 & 0.1647 & 0.3095 & 0.2143 & 0.0000 & 0.4241 & 0.5628 & 0.0000 \\
6 & 16 & 0.0002 & 0.0039 & 0.0006 & 0.1559 & 0.1323 & 0.0208 & 0.3333 & 0.8884 & 0.2848 & 0.4787 & 0.0000 \\
7 & 83 & 0.0126 & 0.0013 & 0.0165 & 0.0000 & -0.0000 & 0.0040 & 0.0482 & 0.0000 & 0.0000 & 0.6886 & 0.0000 \\
\hline
\end{tabular}
}
\end{table}

\subsubsection{Cluster prediction.}

Next, we build machine learning models to predict the cluster class/label based on five simple features that are collected during the screening visit: age, diastolic blood pressure, gender, BMI, and FBG. All features were scaled to the $[0,1]$ interval. We evaluate the performance using leave-one-out cross validation and we use all vs. rest ROC curves and various corresponding AUC metrics to quantify performance. For each test set patient, we predict the probability \emph{that they belong in each cluster}. For each machine learning model, we conduct five-fold cross-validation using the training data to select the best hyperparameters. For logistic regression, we consider both L1 and L2 regularization and six values for the penalty parameter: $[0.001, 0.01, 0.1, 1, 10, 100]$. For random forest, we use 250 trees and consider four max depth parameter values: $[\textnormal{None}, 2, 5, 10]$.

\begin{figure}[ht]
\centering
\includegraphics[width=0.9\textwidth]{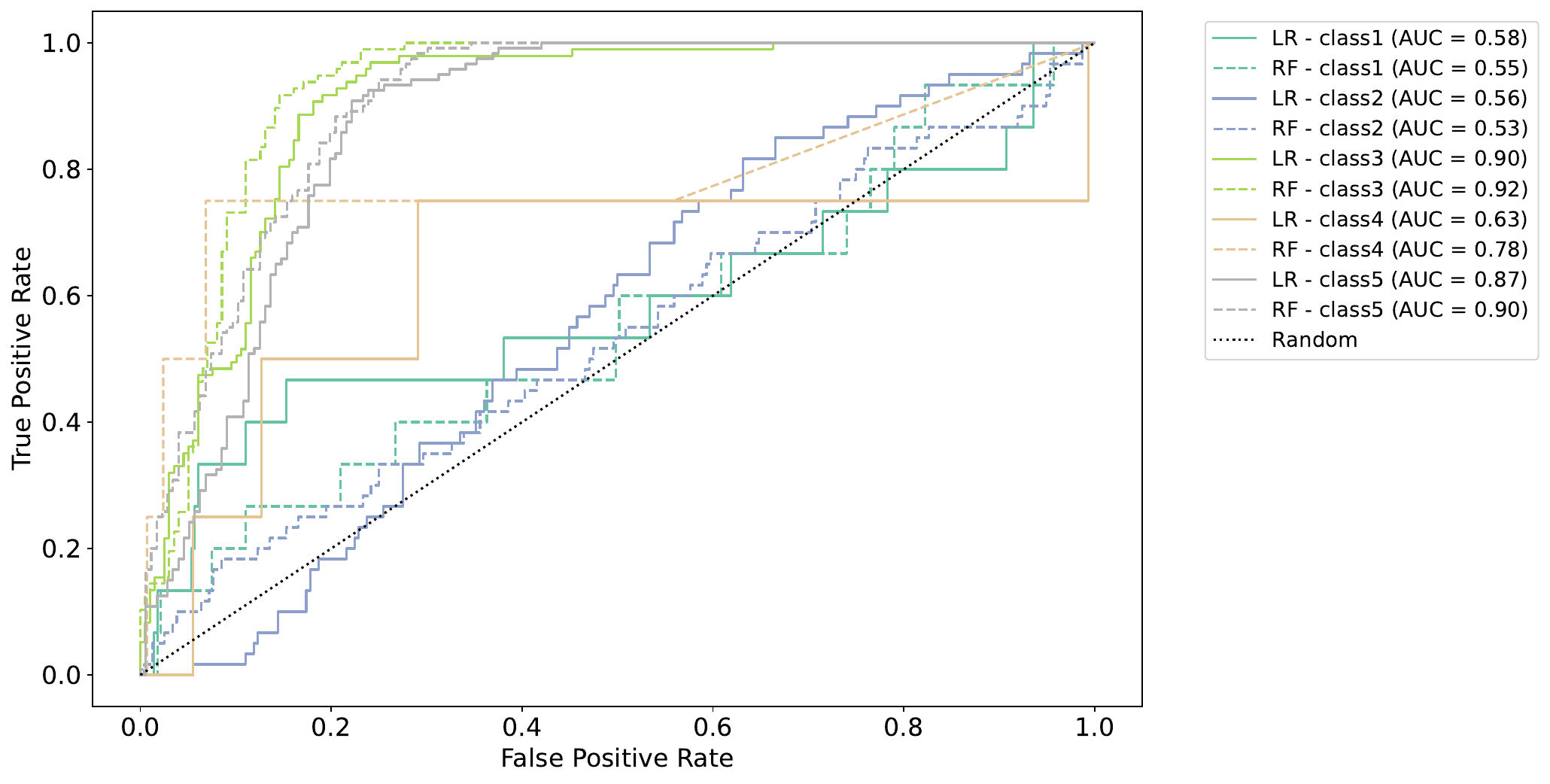}
\caption{Out-of-sample ROC curves and corresponding AUC values for logistic regression (LR) and random forest (RF) methods for predicting patient clusters. \label{fig:ML_clusterPRED}} 
\end{figure}

Figure~\ref{fig:ML_clusterPRED} displays the one vs. rest ROC curves for both models and all five classes. Logistic regression and random forest both performed well with overall weighted-averaged AUC scores of 0.801 and 0.814, respectively. Both models performed best for class 3 (AUC of 0.92 for RF) followed closely by class 5 (AUC of 0.90 for RF) , which account for 70\% of all patients. Both models performed marginally better than guessing (AUC values between 0.53 and 0.58) for class 1 and 2, indicating that these patient types were difficult to identify from the five screening features.

\begin{figure}[ht]
\centering
\includegraphics[width=0.9\textwidth]{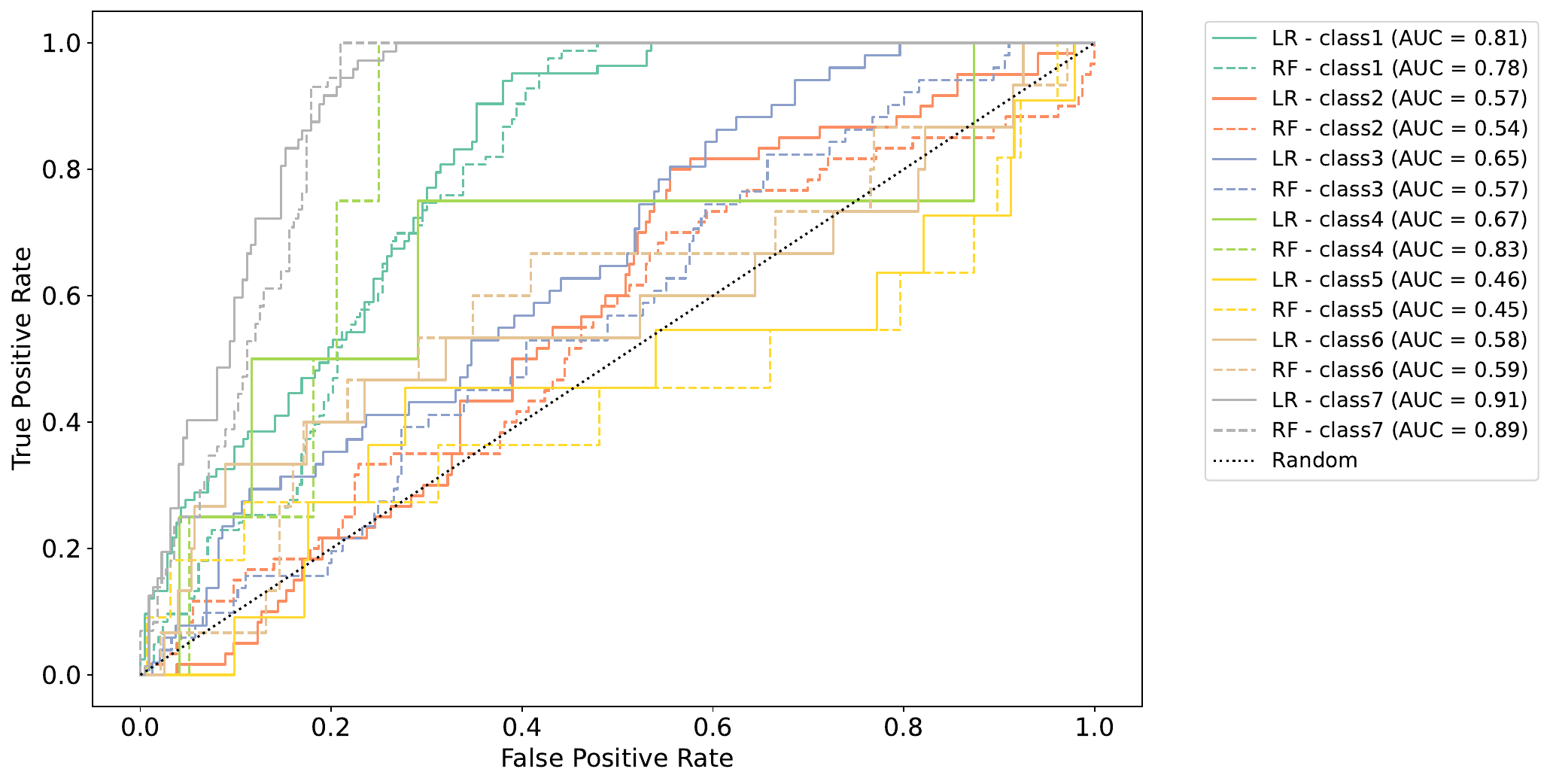}
\caption{Out-of-sample ROC curves and corresponding AUC values for logistic regression (LR), random forest (RF), and the MLE estimates. \label{fig:ML_clusterPRED2}} 
\end{figure}

Figure~\ref{fig:ML_clusterPRED2} displays the one vs. rest ROC curves for both models and all seven classes. Logistic regression and random forest both performed well with overall weighted-averaged AUC scores of 0.730 and 0.702, respectively. Both models performed best for class 7 (AUC of 0.91 for LR) followed by class 1 (AUC of 0.81 for LR), which account for 49\% of all patients. Both models performed slightly better than guessing (AUC values between 0.54 and 0.67) for class 2, 3, 4, and 6 and worse than guessing for class 5, indicating that these patient types were difficult to identify from the five screening features.

Overall, these results demonstrate that it is possible to cluster patients according to their MLE parameters and predict to which cluster a new patient belong (using basic features collecting at screening). This approach could be used to initialize patient parameters and then the MLE estimates can be updated once more data is available for that particular individual.

\section{Experiments with imperfect information}\label{EC:imperfinfo}

\subsection{Results for 10 patients}

\subsubsection{Overall performance.}

\begin{figure}[ht]
\centering
\includegraphics[width=0.9\textwidth]{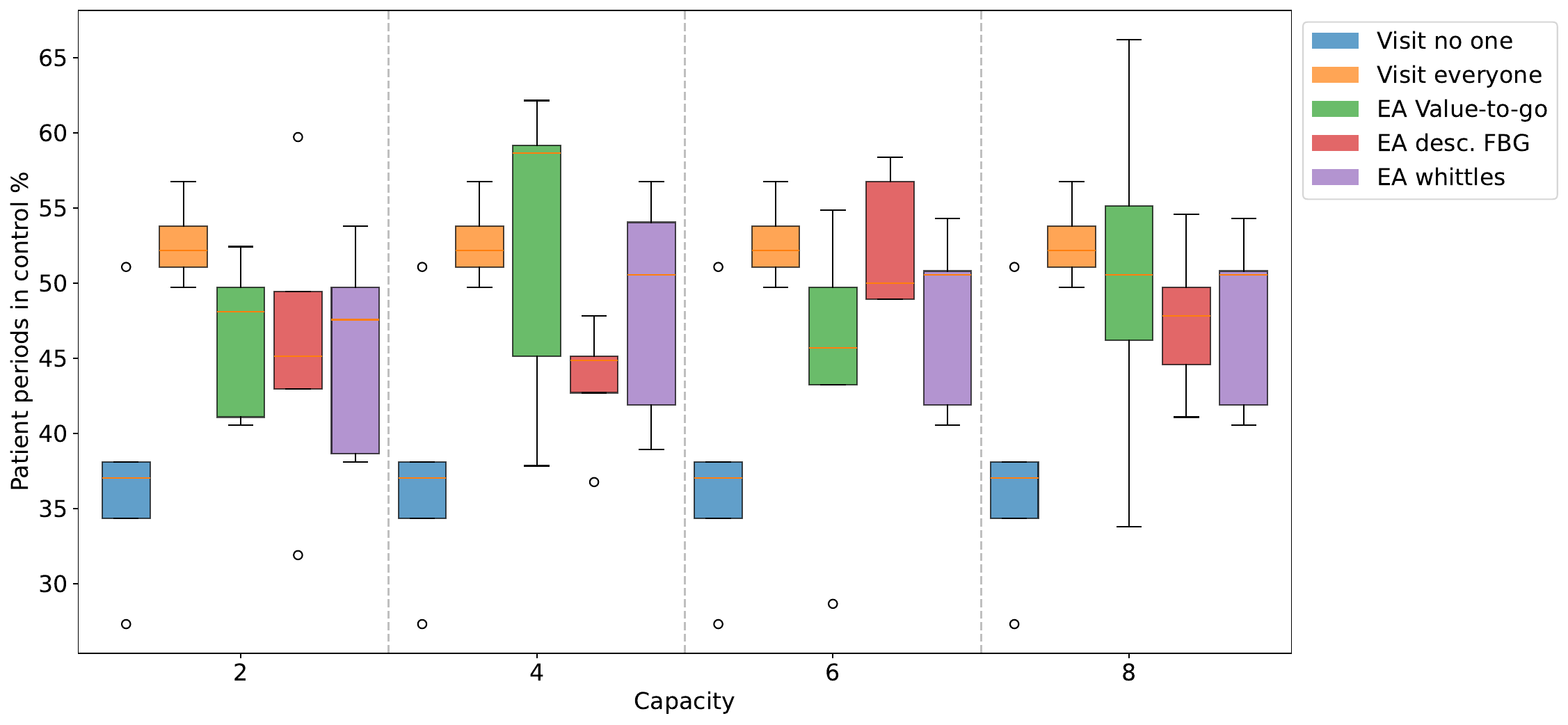}
\caption{Boxplots of the average number of patients in control across all periods for each repetition for 10 patient experiment. \label{fig:CE_1_PPC}} 
\end{figure}

Figure~\ref{fig:CE_1_PPC} displays boxplots of the average number of patients in control across all periods for each repetition. The visit everyone and visit no one policies had average PPC values of 37.6\% and 52.7\%, respectively. For a 20\% capacity level (2 out of 10), the EA value-to-go, EA descending FBG, and EA whittles index had average PPC values of 46.4\%, 45.8\%, and 45.6\%, respectively. The EA value-to-go policy improved upon visit no one by 23.5\% and comprised 88\% of the visit everyone policy with 80\% fewer resources.

\begin{figure}[ht]
\centering
\includegraphics[width=0.9\textwidth]{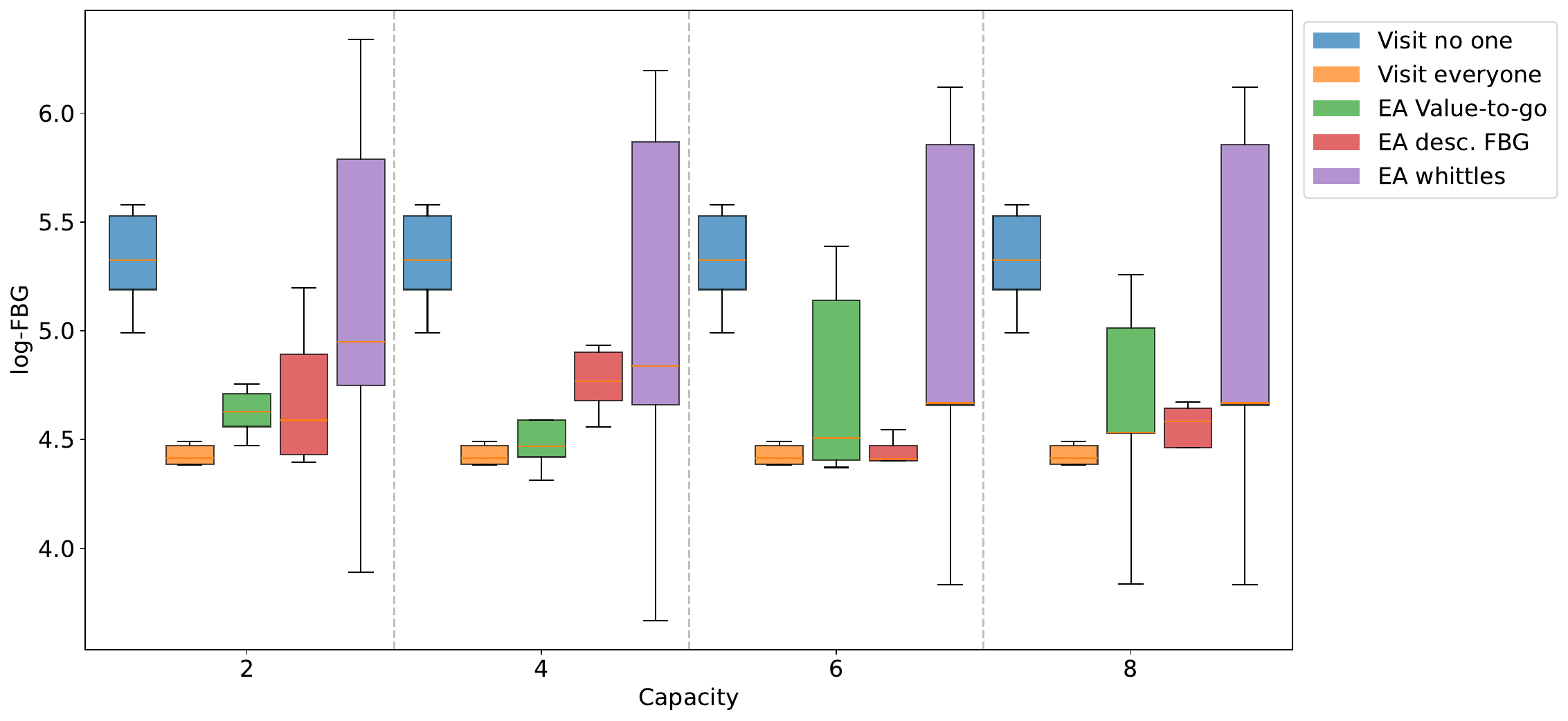}
\caption{Boxplots of the average FBG across all periods for each repetition for 10 patient experiment. \label{fig:CE_1_FBG}} 
\end{figure}

Figure~\ref{fig:CE_1_FBG} displays boxplots of the average FBG across all periods for each repetition. The visit everyone and visit no one policies had average FBG values of 209.5 and 84.0, respectively. For a 20\% capacity level (2 out of 10), the EA value-to-go, EA descending FBG, and EA Whittle index had average FBG values of 102.6, 115.5, and 239.6, respectively. The EA value-to-go policy improved upon visit no one by 51.0\% and captured 85.2\% of improvement from visit no one to visit everyone with 80\% fewer resources.

\subsubsection{Temporal performance.}

\begin{figure}[ht]
\centering
\begin{subfigure}[ht]{0.45\textwidth}
\centering
\includegraphics[width=\textwidth]{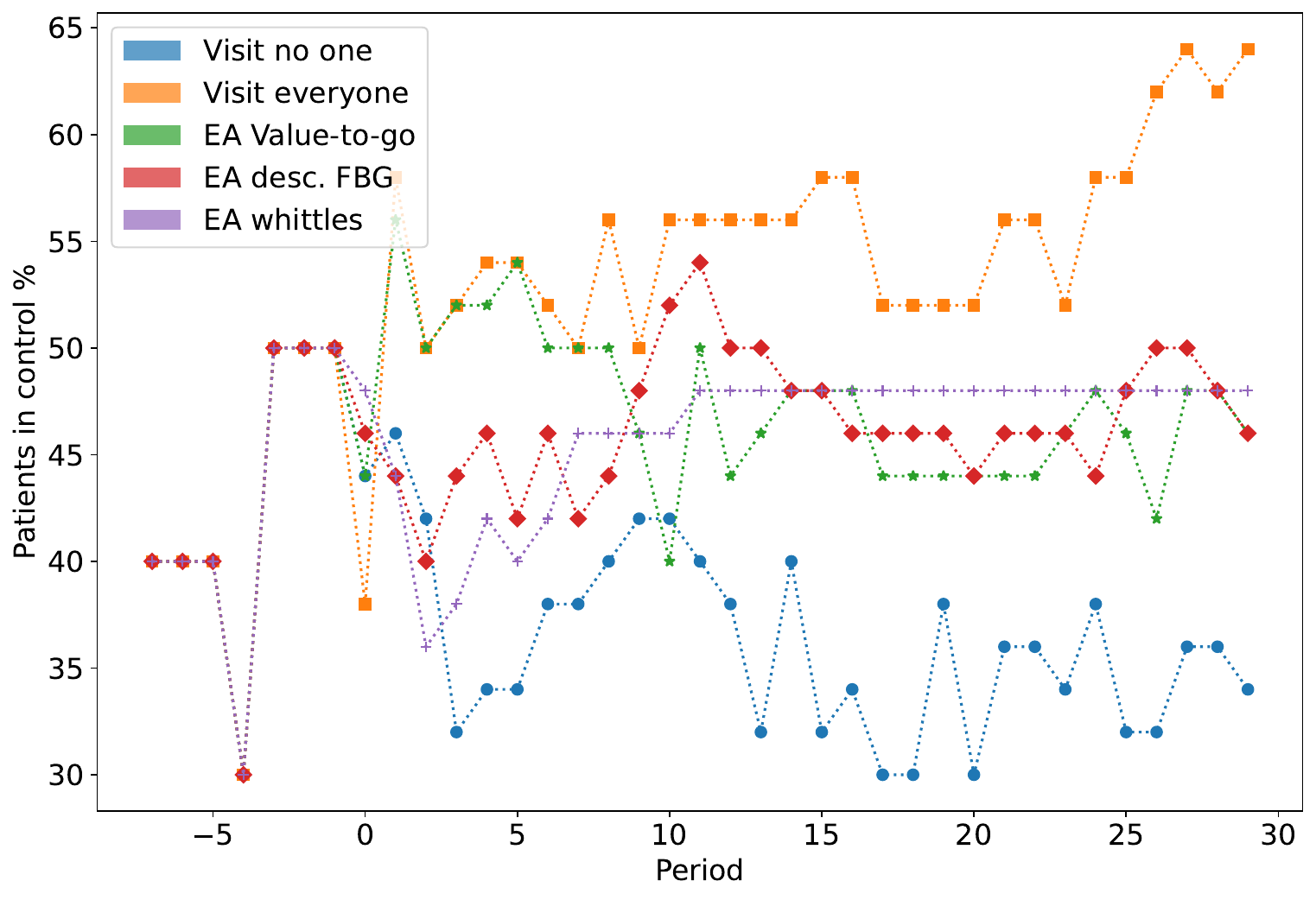}
\caption{\label{fig:CE_2a}}
\end{subfigure}
\hfill
\begin{subfigure}[ht]{0.45\textwidth}
\centering
\includegraphics[width=\textwidth]{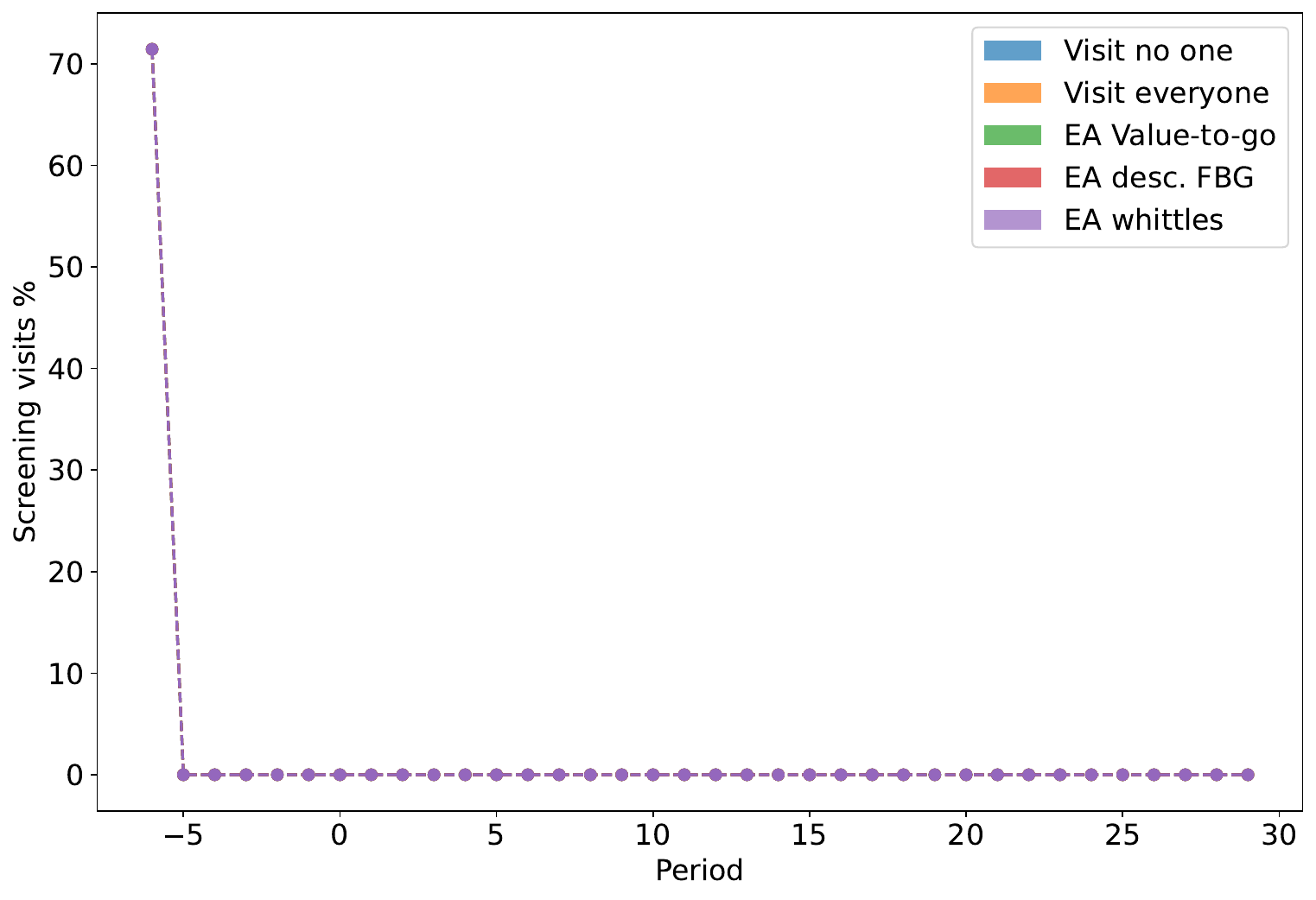} 
\caption{\label{fig:CE_2b}}
\end{subfigure}

\begin{subfigure}[ht]{0.45\textwidth}
\centering
\includegraphics[width=\textwidth]{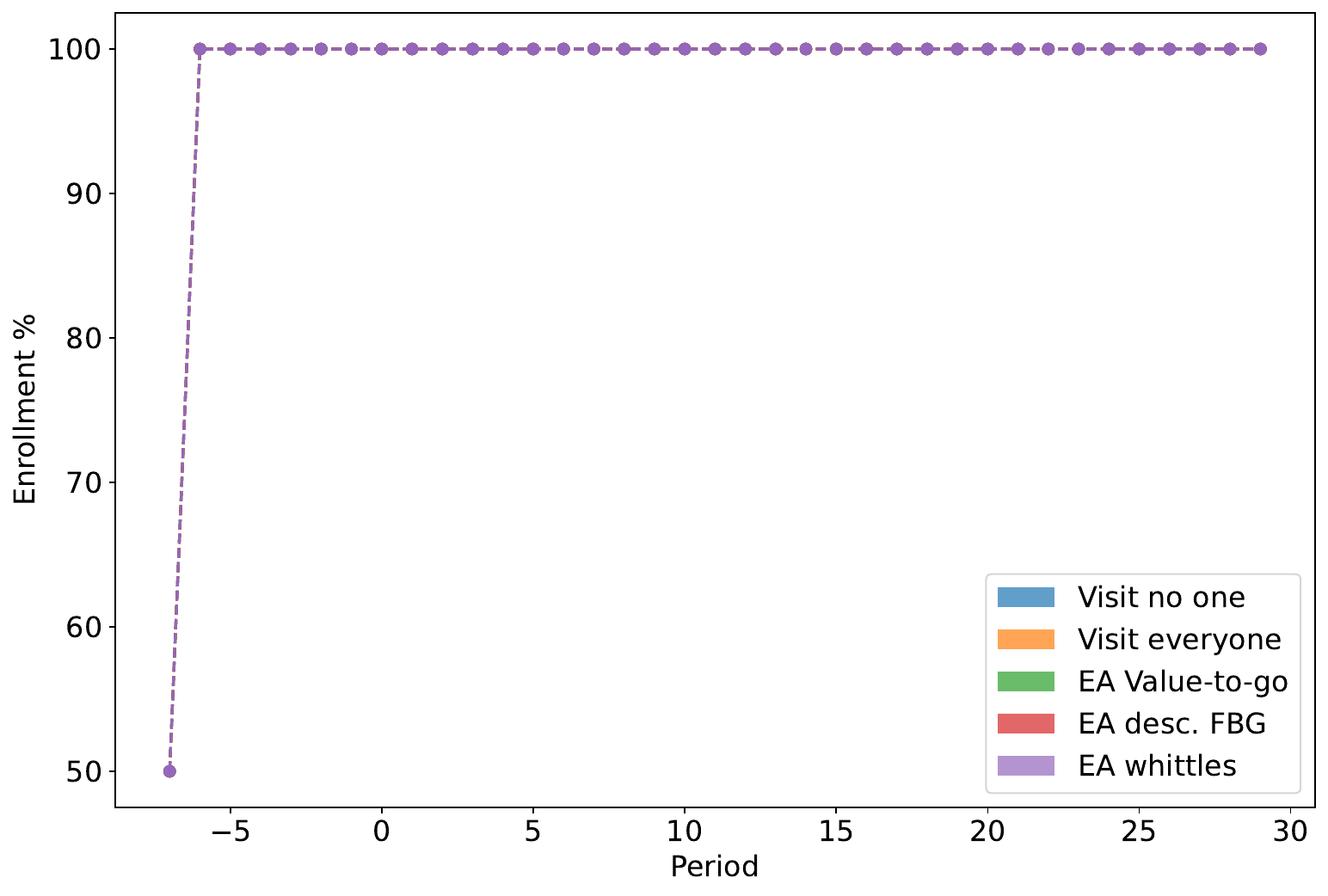} 
\caption{\label{fig:CE_2c}}
\end{subfigure}\vspace{10pt}
\caption{(a) Line plots of the average patient periods in control as a function of the period, (b) Line plots of the average number of screening visits as a function of the period, and (c) Line plots of the average number of enrolled patients as a function of the period. All plots for the 10 patient experiment \label{fig:CE_2}}
\end{figure}

Figure~\ref{fig:CE_2} displays line plots for PPC, enrollment, and visit composition as a function of period for a 20\% capacity level (2 out of 10). Figure~\ref{fig:CE_2a} displays the proportion of patients in control each period for each policy. The x-axis displays the burn in period as negative periods; note that each policy is the same. Over, there is clear separation between visit no one and visit everyone. All EA based methods improve upon visit no one and perform similarly over time. Figure~\ref{fig:CE_2b} displays the proportion of visits each period that were screening visits for each policy. Figure~\ref{fig:CE_2c} displays the proportion of enrolled patients each period for each policy. Note that with the current sample of 10 patients, all patients remain enrolled throughout the time horizon.

\subsection{Results for 25 patients}

\subsubsection{Overall performance.}

\begin{figure}[ht]
\centering
\includegraphics[width=0.9\textwidth]{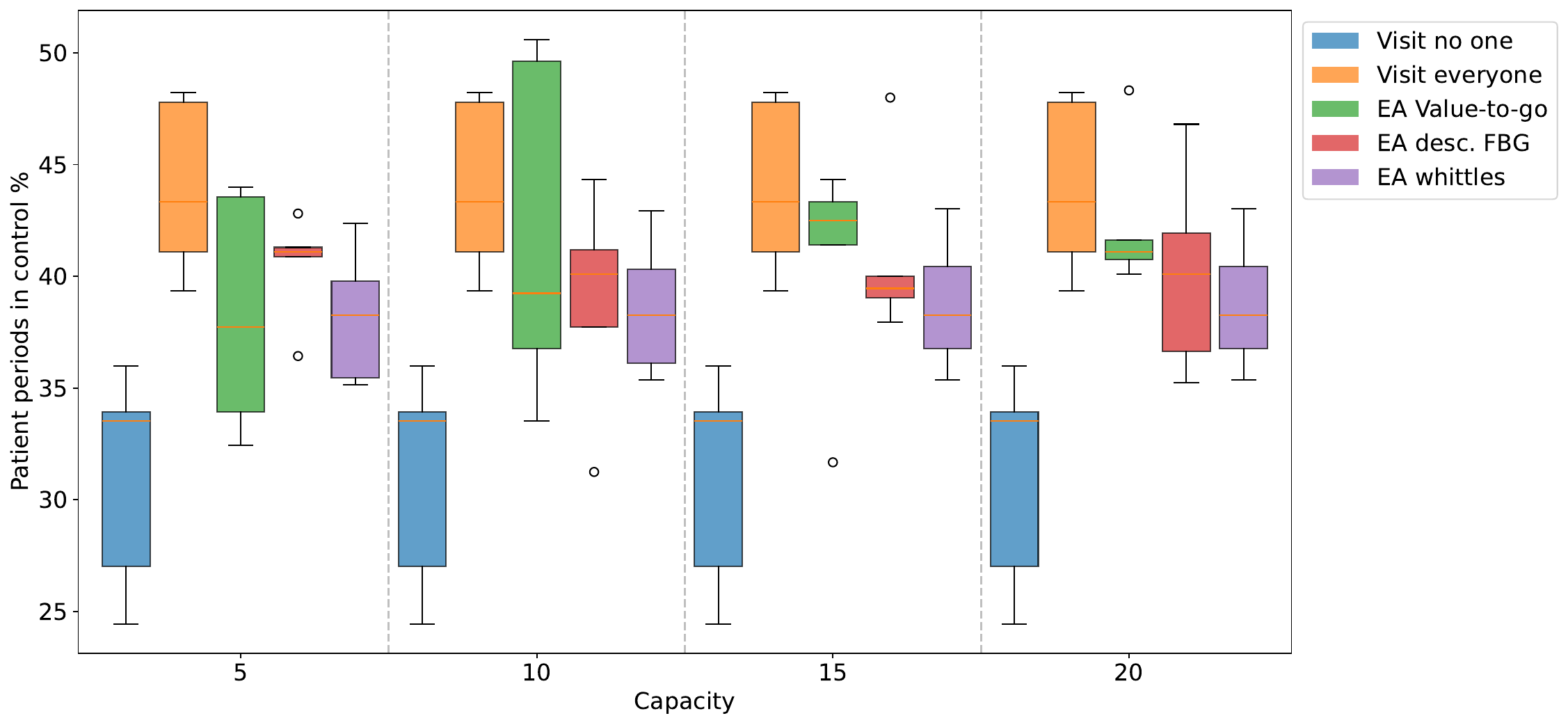}
\caption{Boxplots of the average number of patients in control across all periods for each repetition for the 25 patient experiment. \label{fig:CE_125}} 
\end{figure}

Figure~\ref{fig:CE_125} displays boxplots of the average number of patients in control across all periods for each repetition. The visit everyone and visit no one policies had average PPC values of 31.0\% and 44.0\%, respectively. For a 20\% capacity level (5 out of 25), the EA value-to-go, EA descending FBG, and EA whittles index had average PPC values of 38.3\%, 40.5\%, and 38.2\%, respectively. The EA descending FBG policy improved upon visit no one by 23.7\% and comprised 87.2\% of the visit everyone policy with 80\% fewer resources.

\begin{figure}[ht]
\centering
\includegraphics[width=0.9\textwidth]{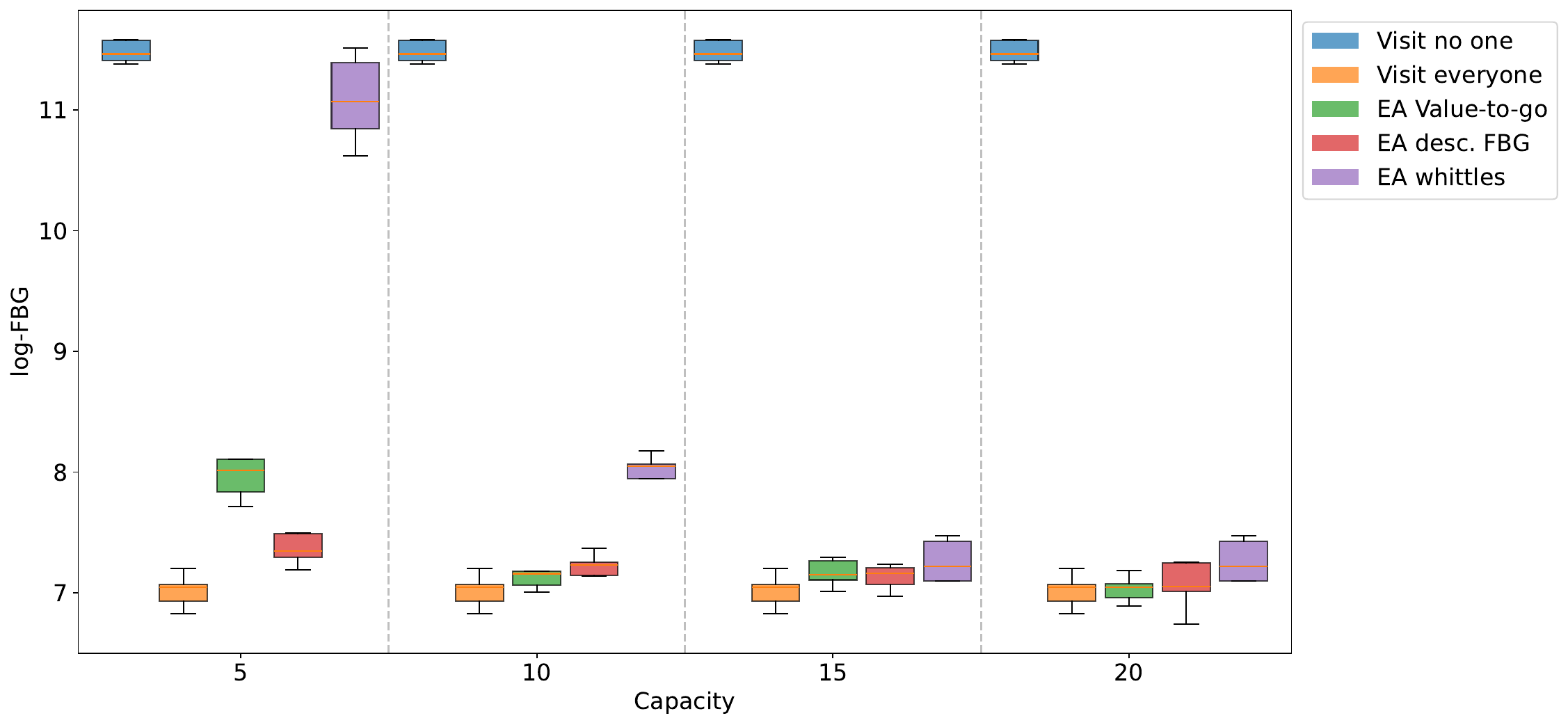}
\caption{Boxplots of the average log-FBG across all periods for each repetition for the 25 patient experiment. \label{fig:CE_FBG_125}} 
\end{figure}

Figure~\ref{fig:CE_FBG_125} displays boxplots of the average log-FBG across all periods for each repetition. We display log-FBG values because the actual FBG values are very large and patients would likely experience an acute event before reaching those levels. The visit everyone and visit no one policies had average log-FBG values of 11.5 and 7.0, respectively. For a 20\% capacity level (5 out of 25), the EA value-to-go, EA descending FBG, and EA whittles index had average FBG values of 8.2, 7.4, and 11.1, respectively. The EA descending FBG improved upon visit no one by 35.9\% (on log scale) and captured 92.2\% of improvement from visit no one to visit everyone with 80\% fewer resources (on log scale). 

\subsubsection{Temporal performance.}

\begin{figure}[ht]
\centering
\begin{subfigure}[ht]{0.45\textwidth}
\centering
\includegraphics[width=\textwidth]{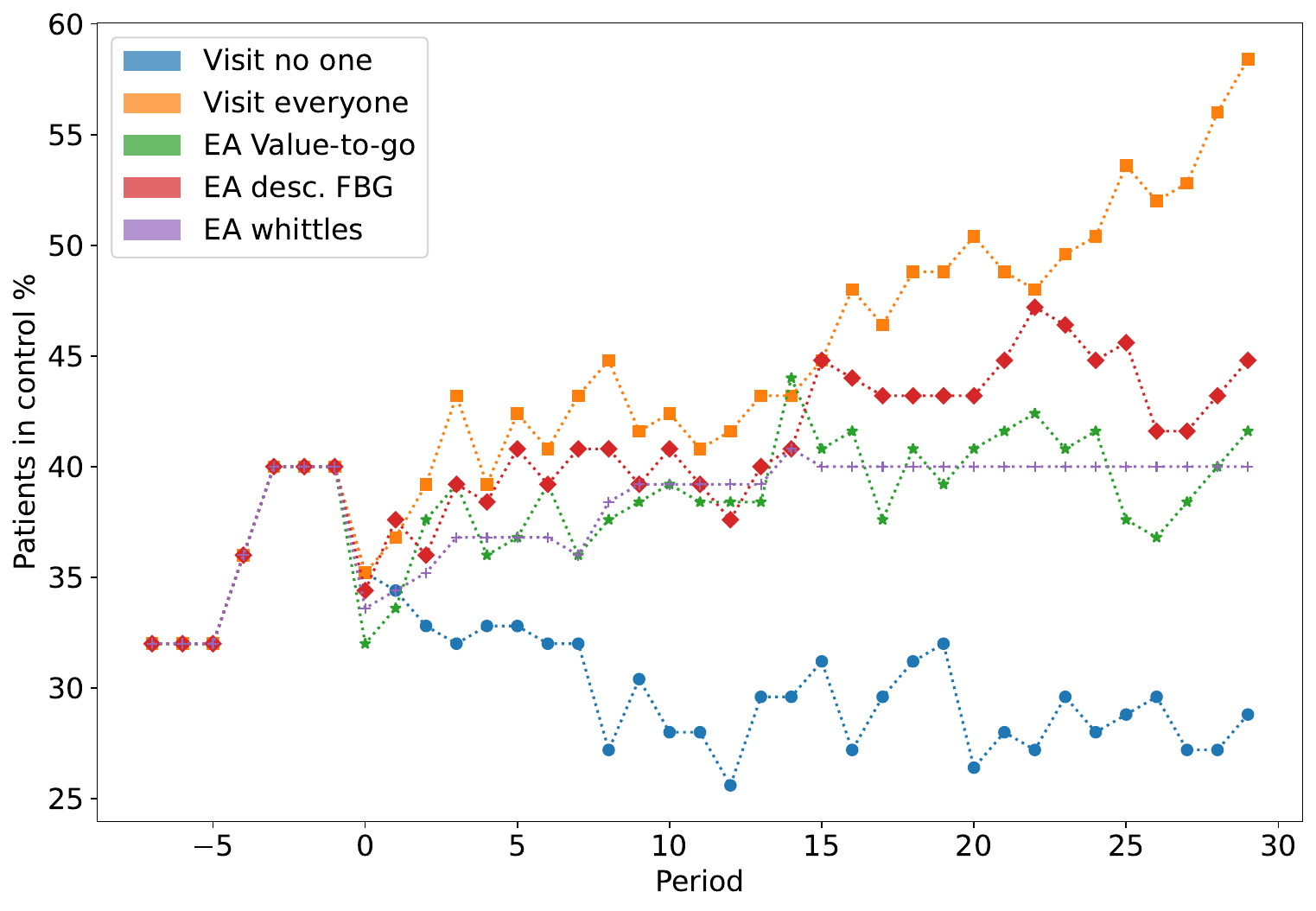}
\caption{\label{fig:CE_2a25}}
\end{subfigure}
\hfill
\begin{subfigure}[ht]{0.45\textwidth}
\centering
\includegraphics[width=\textwidth]{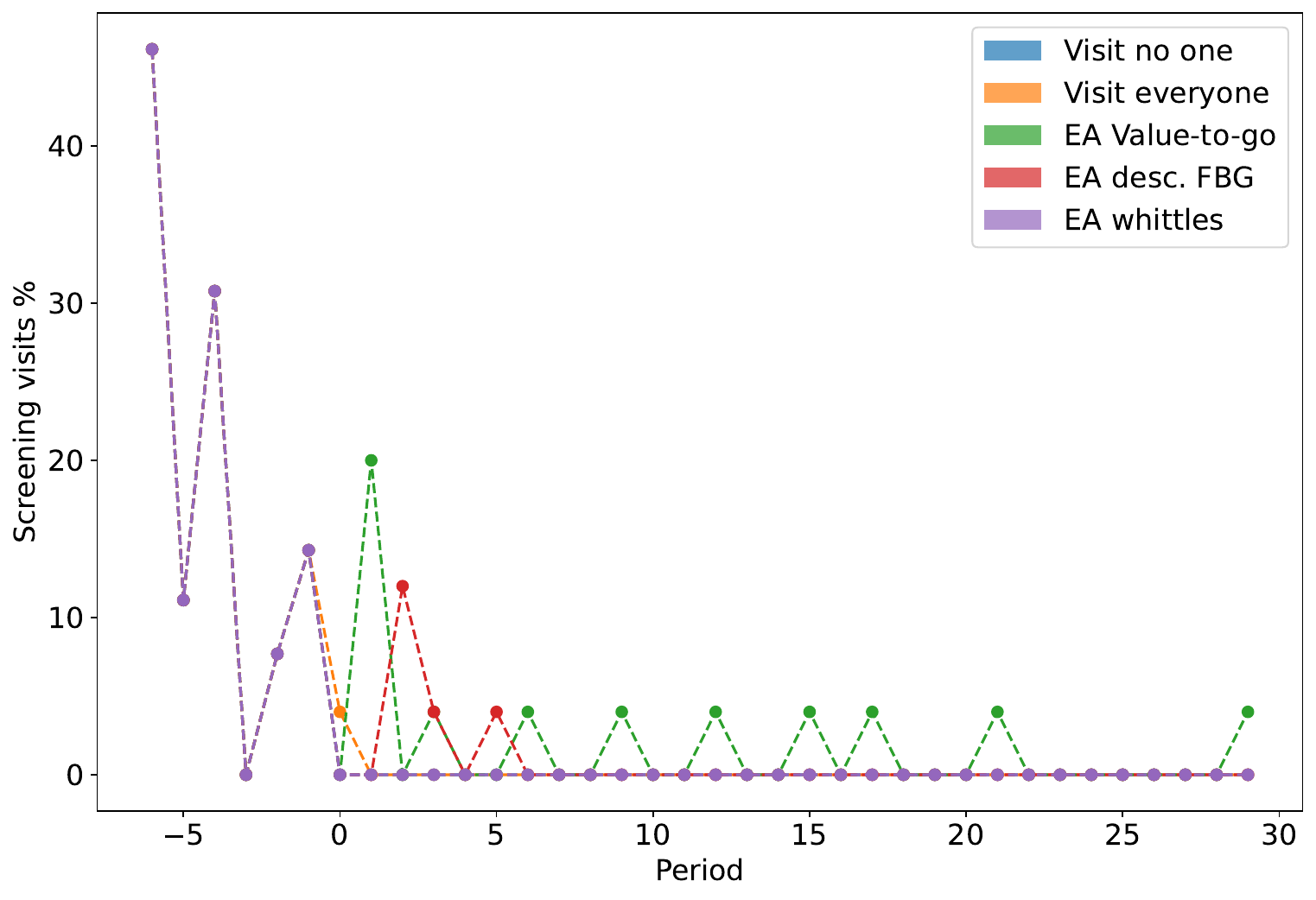} 
\caption{\label{fig:CE_2b25}}
\end{subfigure}

\begin{subfigure}[ht]{0.45\textwidth}
\centering
\includegraphics[width=\textwidth]{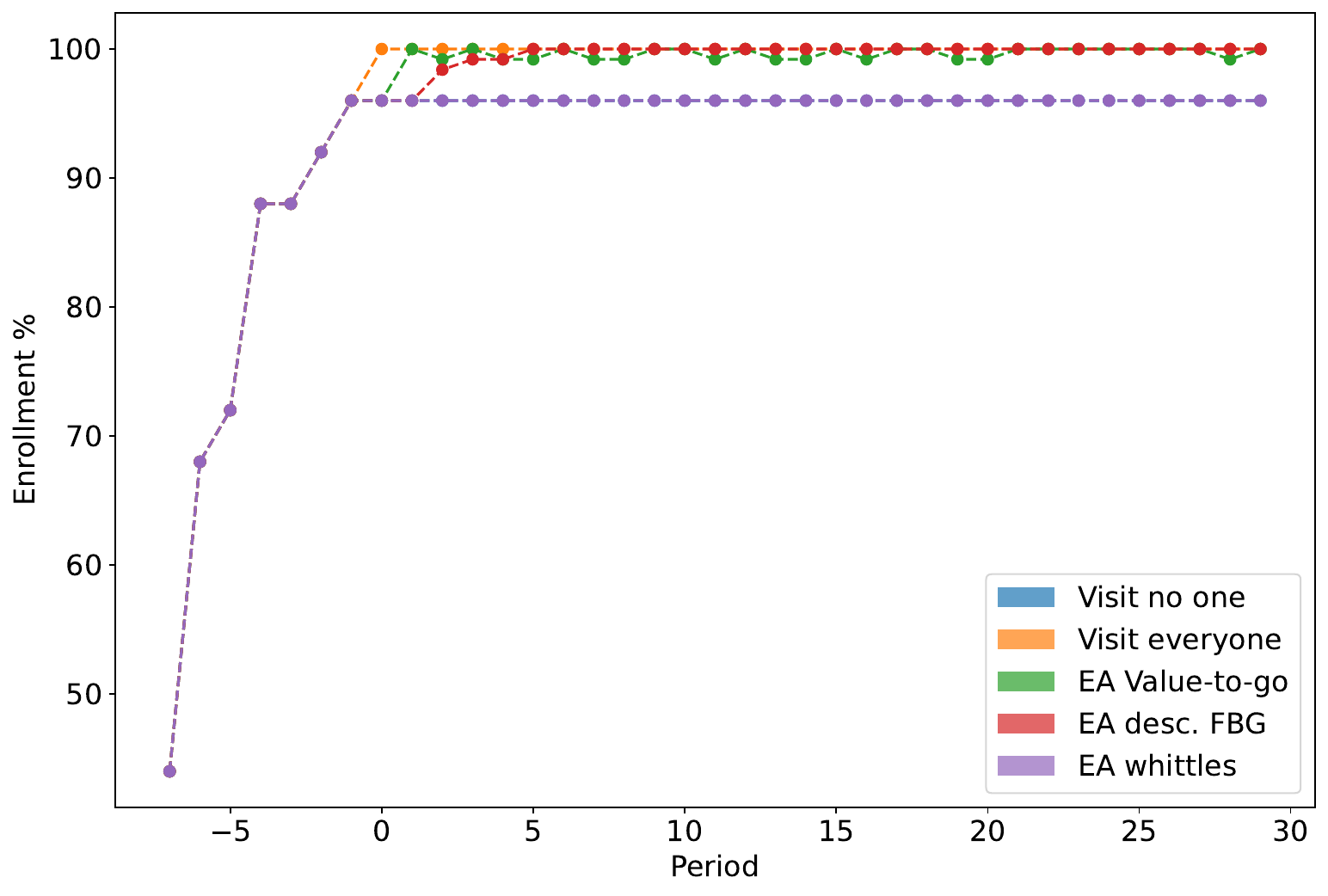} 
\caption{\label{fig:CE_2c25}}
\end{subfigure}\vspace{10pt}
\caption{(a) Line plots of the average patient periods in control as a function of the period, (b) Line plots of the average number of screening visits as a function of the period, and (c) Line plots of the average number of enrolled patients as a function of the period. All plots for the 25 patient experiment.\label{fig:CE_225}}
\end{figure}

Figure~\ref{fig:CE_225} displays line plots for PPC, enrollment, and visit composition as a function of period for a 20\% capacity level (5 out of 25). Figure~\ref{fig:CE_2a25} displays the proportion of patients in control each period for each policy. The x-axis displays the burn in period as negative periods; note that each policy is the same. Overall, there is clear separation between visit no one and visit everyone. All EA based methods improve upon visit no one and perform similarly over time. Figure~\ref{fig:CE_2b25} displays the proportion of visits each period that were screening visits for each policy. Figure~\ref{fig:CE_2c25} displays the proportion of enrolled patients each period for each policy. The visit no one policy maintains an average enrollment of 96\%, while visit everyone and EA descending FBG are able to maintain 100\% enrollment. Both EA value to go and EA whittles index oscillate between 98\% and 100\% enrollment, which can also be seen by the oscillation in screening visits.

\section{MLE Consistency and Proof of Proposition \ref{prop:mle_consist}}\label{EC:prop:mle_consist}

Recall the MLE problem given by the following formulation: \begin{subequations}
\begin{align}
    \underset{\bb,\bs,\btheta,\beta, \alpha, \mu, p, \lambda, \rho, \gamma, \bxi}{\mathrm{minimize}} \: & \sum_{t \in \mathcal{K}} \log f_\epsilon(\bar{b}_t - b_t) + \sum_{t\in\mathcal{T}} \log f_\xi( \xi_t) \\
    \mathrm{subject\,to}\quad &  b_{t+1} = b_{t} + p - \mu z_{t} -\alpha y_{t} z_{t} + \xi_t, \quad  \forall  t \in \mathcal{T},  \\
    & s_{t+1} = \gamma z_{t} (s_t - s_0) + z_t s_0 + \beta y_{t} z_{t}, \quad \forall t \in \mathcal{T}, \\
    & \theta_{t+1} = \rho (\theta_{t} - \theta_{0}) + \theta_{0} - \lambda y_{t} z_{t}, \quad \forall t \in \mathcal{T},\\
    & B_{t} \geq -M(1 - z_{t}), \quad \forall t \in \mathcal{T}, \\
    & B_{t} \leq M z_{t}, \quad \forall t \in \mathcal{T},  \\
    & b_{t} \in \mathcal{B},s_{t} \in\mathcal{S},\theta_t \in \Theta, \quad \forall t \in \mathcal{T},  \\
    & \beta \in \mathcal{S}, \alpha, \mu, p \in \mathcal{B}, \lambda \in \Theta, \rho, \gamma \in (0,1).
\end{align}
\end{subequations}
For our analysis we will make the following simplifying assumption.
\begin{assumption}
     $s_0 \in \overline{\mathcal{S}} \subset\mathcal{S},\beta \in \overline{B} \subset \mathcal{S}$, where $|\overline{\mathcal{S}}| < \infty, |\overline{B}| < \infty $
\end{assumption}
Essentially we are assuming that $s_0$, $\beta$ come from finite sets, which correspond to the gird points used in the grid search algorithm. It is clear that under this assumption, we can compute the exact optimal solution to the above problem using grid search across several MILPs, denote this solution as $(\hat{b}_0,\hat{s}_0,\hat{\theta}_0,\hat{\beta},\hat{\alpha},\hat{\mu},\hat{p},\hat{\lambda}) $. Note that we need only keep track of the initial states of the system since given the model dynamics they then define the trajectories of all other quantities. Using this assumption we can show that the estimates provided by the MLE problem are consistent. We will use a Bayesian consistency approach \citep{mintz2023behavioral,li2023adaptive,he2023model}. To complete this analysis we will need the following definition for Bayesian consistency:
\begin{definition}
\label{def:bayes_consist}
For all $(b_0,s_0,\theta_0,\beta,\alpha,\mu,p,\lambda) \in \mathcal{B}^3\times\mathcal{S}^2\times\Theta^2$ and constants $r, \delta >0$, we say the estimate of the posterior distribution $\hat{\mathbb{P}}(\cdot|\{\overline{b}_t\}_{t\in \mathcal{K}},\{z_t,y_t\}_{t \in \mathcal{T}})$ is consistent if $\mathbb{P}_{(b^*_0,s^*_0,\theta^*_0,\beta^*,\alpha^*,\mu^*,p^*,\lambda^*)} (\hat{\mathbb{P}}(\mathcal{E}(\delta)|\{\overline{b}_t\}_{t\in \mathcal{K}},\{z_t,y_t\}_{t \in \mathcal{T}}) \geq r) \rightarrow 0 \text{ as } t \rightarrow \infty $. Here $\mathbb{P}_{(b^*_0,s^*_0,\theta^*_0,\beta^*,\alpha^*,\mu^*,p^*,\lambda^*)}$ is the probability law where $(b^*_0,s^*_0,\theta^*_0,\beta^*,\alpha^*,\mu^*,p^*,\lambda^*))$ are the true initial conditions of the system, and  where
$\mathcal{E}(\delta):=\{(b_0,s_0,\theta_0,\beta,\alpha,\mu,p,\lambda) \not\in \mathcal{B}((b^*_0,s^*_0,\theta^*_0,\beta^*,\alpha^*,\mu^*,p^*,\lambda^*),\delta)\}$, where $\mathcal{B}((b^*_0,s^*_0,\theta^*_0,\beta^*,\alpha^*,\mu^*,p^*,\lambda^*),\delta)$ is an open ball with radius $\delta$ centered around $(b^*_0,s^*_0,\theta^*_0,\beta^*,\alpha^*,\mu^*,p^*,\lambda^*))$.
\end{definition}

This definitions means that for any point other than the true initial parameters $(b^*_0,s^*_0,\theta^*_0,\beta^*,\alpha^*,\mu^*,p^*,\lambda^*)$, the log-likelihood becomes infinitely small as we obtain more observations. We need one additional assumption known as \emph{sufficient excitation} to prove the statistical consistency of our estimates.
\begin{assumption}
\label{ass:sufficient_excitation}
Let $(b^*_0,s^*_0,\theta^*_0,\beta^*,\alpha^*,\mu^*,p^*,\lambda^*)$ be the participant's true parameters. The visits given to the participant $\{y_t\}_{t\in \mathcal{T}}$ are such that:
\begin{equation}
    \max _{\mathcal{E}(\delta)} \lim _{|\mathcal{K}|,|\mathcal{T}| \rightarrow \infty} \sum_{t\in \mathcal{K}} \log \frac{  f_\epsilon(\bar{b}_t - b_t) }{  f_\epsilon(\bar{b}_t - b^*_t)} + \sum_{t \in \mathcal{T}}\frac{ f_\xi( \xi_t)}{f_\xi( \xi^*_t)}=-\infty,
\end{equation}
for any $\delta > 0$, almost surely, where $b^*_t$ are mean log FBG measures under the true parameters and $\xi^*_t$ are the true disturbance realizations, and $b_t$ are the mean log FBG measures under any other parameters with corresponding estimated disturbance $\xi_t$.
\end{assumption} 
This type of assumption is common in the adaptive control literature \citep{craig1987adaptive, aastrom2013adaptive} known as a \emph{sufficient excitation} or a \emph{sufficient richness} condition. This assumption can be satisfied in practice by either injecting noise into the visit schedule, but has also been observed to be implicitly satisfied in settings where decisions rely on participant specific contexts \citep{bastani2021mostly}. With these in mind, consider the following feasibility problem:
\begin{subequations}
\begin{align}
    \psi(\tilde{b}_0,\tilde{s}_0,\tilde{\theta}_0,  \tilde{\beta}, \tilde{\alpha}, \tilde{\mu},\tilde{p}, \tilde{\lambda}) = \underset{\bb,\bs,\btheta,\beta, \alpha, \mu, p, \lambda, \rho, \gamma, \bxi}{\mathrm{minimize}} \: & \sum_{t \in \mathcal{K}} \log f_\epsilon(\bar{b}_t - b_t) + \sum_{t\in\mathcal{T}} \log f_\xi( \xi_t) \\
    \mathrm{subject\,to}\quad &  b_{t+1} = b_{t} + p - \mu z_{t} -\alpha y_{t} z_{t} + \xi_t, \quad  \forall  t \in \mathcal{T},  \\
    & s_{t+1} = \gamma z_{t} (s_t - s_0) + z_t s_0 + \beta y_{t} z_{t}, \quad \forall t \in \mathcal{T}, \label{c_s}\\
    & \theta_{t+1} = \rho (\theta_{t} - \theta_{0}) + \theta_{0} - \lambda y_{t} z_{t}, \quad \forall t \in \mathcal{T}, \label{c_t}\\
    & B_{t} \geq -M(1 - z_{t}), \quad \forall t \in \mathcal{T}, \\
    & B_{t} \leq M z_{t}, \quad \forall t \in \mathcal{T},  \\
    & b_{t} \in \mathcal{B},s_{t} \in\mathcal{S},\theta_t \in \Theta, \quad \forall t \in \mathcal{T}, \\
    & b_0 = \tilde{b}_0, s_0 = \tilde{s}_0, \theta_0 = \tilde{\theta}_0, \beta = \tilde{\beta}, \alpha = \tilde{\alpha}, \mu = \tilde{\mu}, p = \tilde{p}, \lambda =\tilde{\lambda}.
\end{align}
\end{subequations}
By definition it is clear that $\psi(\hat{b}_0,\hat{s}_0,\hat{\theta}_0,\hat{\beta},\hat{\alpha},\hat{\mu},\hat{p},\hat{\lambda}) \leq  \psi(b_0,s_0,\theta_0,\beta,\alpha,\mu,p,\lambda)$ for all $(b_0,s_0,\theta_0,\beta,\alpha,\mu,p,\lambda) \in \mathcal{B}^3\times\mathcal{S}^2\times\Theta^2$. Then one measure of interest is the following likelihood ratio that can be interpreted as probability measure over the parameter space: $\hat{\mathbb{P}}(b_0,s_0,\theta_0,\beta,\alpha,\mu,p,\lambda |\{\overline{b}_t\}_{t\in \mathcal{K}},\{z_t,y_t\}_{t \in \mathcal{T}}) = \frac{\exp(-\psi(b_0,s_0,\theta_0,\beta,\alpha,\mu,p,\lambda))}{\exp(-\psi(\hat{b}_0,\hat{s}_0,\hat{\theta}_0,\hat{\beta},\hat{\alpha},\hat{\mu},\hat{p},\hat{\lambda}))}$. Our first result will be to show that this likelihood satisfies the property of Definition \ref{def:bayes_consist}.

\begin{proposition}
\label{prop:bayes_consist}
Likelihood $\hat{\mathbb{P}}(b_0,s_0,\theta_0,\beta,\alpha,\mu,p,\lambda |\{\overline{b}_t\}_{t\in \mathcal{K}},\{z_t,y_t\}_{t \in \mathcal{T}}) $ is Bayesian consistent.
\end{proposition}

\proof{Proof of \ref{prop:bayes_consist}: } Note that by definition for any $(b_0,s_0,\theta_0,\beta,\alpha,\mu,p,\lambda) \in \mathcal{B}^3\times\mathcal{S}^2\times\Theta^2$ that are not the true parameters, we can express the likelihood as:
\begin{multline}
   \log \hat{\mathbb{P}}(b_0,s_0,\theta_0,\beta,\alpha,\mu,p,\lambda |\{\overline{b}_t\}_{t\in \mathcal{K}},\{z_t,y_t\}_{t \in \mathcal{T}})  =\\ \log \hat{\mathbb{P}}(b^*_0,s^*_0,\theta^*_0,\beta^*,\alpha^*,\mu^*,p^*,\lambda^* |\{\overline{b}_t\}_{t\in \mathcal{K}},\{z_t,y_t\}_{t \in \mathcal{T}}) +\sum_{t \in \mathcal{K}} \log \frac{  f_\epsilon(\bar{b}_t - b_t) }{  f_\epsilon(\bar{b}_t - b^*_t)} + \sum_{t \in \mathcal{T}}\frac{ f_\xi( \xi_t)}{f_\xi( \xi^*_t)}
\end{multline}
Note that by construction $\hat{\mathbb{P}}(b^*_0,s^*_0,\theta^*_0,\beta^*,\alpha^*,\mu^*,p^*,\lambda^* |\{\overline{b}_t\}_{t\in \mathcal{K}},\{z_t,y_t\}_{t \in \mathcal{T}}) \in [0,1]$ thus $\log \hat{\mathbb{P}}(b^*_0,s^*_0,\theta^*_0,\beta^*,\alpha^*,\mu^*,p^*,\lambda^* |\{\overline{b}_t\}_{t\in \mathcal{K}},\{z_t,y_t\}_{t \in \mathcal{T}}) \leq 0$. Thus using Assumption \ref{ass:sufficient_excitation} implies that  $\max_{\mathcal{E}(\delta)}\lim_{|\mathcal{K}|,|\mathcal{T}|\rightarrow \infty}  \log \hat{\mathbb{P}}(b_0,s_0,\theta_0,\beta,\alpha,\mu,p,\lambda |\{\overline{b}_t\}_{t\in \mathcal{K}},\{z_t,y_t\}_{t \in \mathcal{T}}) \rightarrow -\infty $. Therefore, this means $\max_{\mathcal{E}(\delta)}\lim_{|\mathcal{K}|,|\mathcal{T}|\rightarrow \infty}  \hat{\mathbb{P}}(b_0,s_0,\theta_0,\beta,\alpha,\mu,p,\lambda |\{\overline{b}_t\}_{t\in \mathcal{K}},\{z_t,y_t\}_{t \in \mathcal{T}}) \rightarrow 0 $. 
Using this result we can see:
\begin{align}
    &\hat{\mathbb{P}}(\mathcal{E}(\delta) |\{\overline{b}_t\}_{t\in \mathcal{K}},\{z_t,y_t\}_{t \in \mathcal{T}})\\
    &= \int_{\mathcal{E}(\delta)}\hat{\mathbb{P}}(b_0,s_0,\theta_0,\beta,\alpha,\mu,p,\lambda |\{\overline{b}_t\}_{t\in \mathcal{K}},\{z_t,y_t\}_{t \in \mathcal{T}}) \ d(b_0,s_0,\theta_0,\beta,\alpha,\mu,p,\lambda) \\
    &\leq \text{Vol}(\mathcal{B}^3\times\mathcal{S}^2\times\Theta^2)\max_{\mathcal{E}(\delta)}\lim_{|\mathcal{K}|,|\mathcal{T}|\rightarrow \infty}  \hat{\mathbb{P}}(b_0,s_0,\theta_0,\beta,\alpha,\mu,p,\lambda |\{\overline{b}_t\}_{t\in \mathcal{K}},\{z_t,y_t\}_{t \in \mathcal{T}})  \rightarrow 0
\end{align}
Here Vol is the volume operator, which is well defined and finite by assumptions on the parameter sets being bounded or finite. This result holds almost surely for any $\delta>0$ thus completing the proof. \halmos  \endproof

Using this result we can now proceed to prove our main consistency result
\proof{Proof of Proposition \ref{prop:mle_consist}: } Consider the following two events: $E_1 = \{ (\hat{b}_0,\hat{s}_0,\hat{\theta}_0,\hat{\beta},\hat{\alpha},\hat{\mu},\hat{p},\hat{\lambda})  \not\in \mathcal{B}((b^*_0,s^*_0,\theta^*_0,\beta^*,\alpha^*,\mu^*,p^*,\lambda^*),\delta) \}$ and $E_2 = \{\max_{\mathcal{E}(\delta)} \hat{\mathbb{P}}(b_0,s_0,\theta_0,\beta,\alpha,\mu,p,\lambda |\{\overline{b}_t\}_{t\in \mathcal{K}},\{z_t,y_t\}_{t \in \mathcal{T}}) \geq \max_{\mathcal{B}((b^*_0,s^*_0,\theta^*_0,\beta^*,\alpha^*,\mu^*,p^*,\lambda^*),\delta)} \hat{\mathbb{P}}(b_0,s_0,\theta_0,\beta,\alpha,\mu,p,\lambda |\{\overline{b}_t\}_{t\in \mathcal{K}},\{z_t,y_t\}_{t \in \mathcal{T}}) \}$. By construction $E_1 \subset E_2$, therefore $\mathbb{P}(E_1) \leq \mathbb{P}(E_2)$, but note that $\mathbb{P}(E_2) \rightarrow 0 $ by Proposition \ref{prop:bayes_consist} thus proving the desired result. \halmos
\endproof
Note that the implication of these results is that, even if the true parameters do not lie on the grid, the resulting estimates will minimize the likelihood ratio, and thus the KL divergence, between the true distribution and closest set of distributional parameters on the grid.

\section{Subgradient Approach for Lagrangian Relaxation}
\label{app:subgradient-approach}
Using the result of Proposition \ref{prop:lagrange_single_struct}, we can efficiently compute the subgradient of $\mathcal{V}^{\blambda}_{i,t}$, and, by extension, the subgradient of $\mathcal{L}(\lambda)$. 
\begin{proposition}
\label{prop:lambda_subgradient}
    For any $\blambda \in \mathbb{R}^{|\mathcal{T}|_+},b_{i,t},s_{i,t},\theta_{i,t} \geq 0$, $z_{i,t-1} \in \{0,1\}$, and $t,j \leq N-1$, we have that for a fixed policy $\{y_{i,1}, ...,y_{i,N-1} \}$:
    \begin{enumerate}
        \item If $j<t$, then \begin{equation}\frac{\partial\mathcal{V}^{\blambda}_{i,t}(b_{i,t},s_{i,t},\theta_{i,t},z_{i,t-1})}{\partial \lambda_j} = 0
        \end{equation} 
        \item If $j\geq t$ and $y_{i,j} = 0$, then \begin{equation}\frac{\partial\mathcal{V}^{\blambda}_{i,t}(b_{i,t},s_{i,t},\theta_{i,t},z_{i,t-1})}{\partial \lambda_j} = 0
        \end{equation}
        \item If $j\geq t$ and $y_{i,j} = 1$, then \begin{equation}\frac{\partial\mathcal{V}^{\blambda}_{i,t}(b_{i,t},s_{i,t},\theta_{i,t},z_{i,t-1})}{\partial \lambda_j} = -1
        \end{equation} 
        \item If $j\geq 1$ and $y_{i,j} = 1$, then \begin{equation}\frac{\partial\mathcal{V}^{\blambda}_{i,1}(b_{i,1},s_{i,1},\theta_{i,1},z_{i,0})}{\partial \lambda_j} = -1
        \end{equation} 
    \end{enumerate}
\end{proposition}
The proof of Proposition \ref{prop:lambda_subgradient} is trivial and follows immediately form Proposition \ref{prop:lagrange_single_struct}. This result implies that the subgradient for any fixed open-loop policy has a very simple structure.  If we consider a  closed-loop policy $\pi$, this implies that $\frac{\partial\mathcal{V}^{\blambda}_{i,1}(b_{i,1},s_{i,1},\theta_{i,1},z_{i,0})}{\partial \lambda_j} = - \mathbb{P}(y_{i,j} = 1 | b_{i,1},s_{i,1},\theta_{i,1},z_{i,0}, \pi )$. Thus the subgradient can be computed efficiently using backwards induction for any given policy extracted from dynamic programming.  This leads us to the sub-gradient method presented in Algorithm \ref{alg:lagrange}. 
\begin{algorithm2e}[H]
    \DontPrintSemicolon
    \KwIn{${\blambda}_0 = (\lambda_{0,1},....,\lambda_{0,N-1}),\bb_1,\bs_1,\btheta_1,\bz_{0}, \epsilon, MR, \{\gamma_k\}_{k=0}^{MR}$,} \;
    \For{$k \in \{0,...,MR\}$}{
    \For{$i \in \mathcal{P}$}{
    Compute $\mathcal{V}_{i,1}^{\blambda_k}(b_{i,1},s_{i,1},\theta_{i,1},z_{i,0})$ using Equation~\eqref{eq:langrange_singele_patient} \;
    Compute $\partial_{{\blambda}_k}\mathcal{V}_{i,1}^{\blambda_k}(b_{i,1},s_{i,1},\theta_{i,1},z_{i,0})$\;
    }
    Set: $\mathcal{V}_1^{\blambda_k}(\bb_1,\bs_1,\btheta_1,\bz_{0}) = C \sum_{t = 0}^N\lambda_{k,t} + \sum_{i \in \mathcal{P}} \mathcal{V}_{i,1}^{\blambda_k}(b_{i,1},s_{i,1},\theta_{i,1},z_{i,0})$\;
    Set: $\partial_{\blambda_k}\mathcal{V}_1^{\blambda_k}(\bb_1,\bs_1,\btheta_1,\bz_{0}) = C\mathbbm{1} + \sum_{i \in \mathcal{P}}\partial_{\blambda_k}\mathcal{V}_{i,1}^{\blambda_k}(b_{i,1},s_{i,1},\theta_{i,1},z_{i,0})$ \;
    \If{$\|\partial_{\blambda_k}\mathcal{V}_1^{\blambda_k}(\bb_1,\bs_1,\btheta_1,\bz_{0})\| \leq \epsilon$}{ \Return $\mathcal{V}_1^{\blambda_k}(\bb_1,\bs_1,\btheta_1,\bz_{0})$}
    Update ${\blambda}_{k+1} \gets {\blambda}_{k} - \gamma_k  \partial_{\blambda_0}\mathcal{V}_1^{\blambda_k}(\bb_1,\bs_1,\btheta_1,\bz_{0})$
    }
    \Return $\mathcal{V}_1^{\blambda_{MR}}(\bb_1,\bs_1,\btheta_1,\bz_{0})$
    \caption{Subgradient Procedure for Lagrangian Relaxation \label{alg:lagrange}}
\end{algorithm2e}

Algorithm \ref{alg:lagrange} uses the structural results from \ref{prop:lagrange_struct} to compute the subgradient by first computing the subgradient for each individual participant and then combining them according to the decomposability result in Proposition~\ref{prop:lagrange_struct}. Step sizes $\gamma_k$ should be chosen using standard assumptions to ensure convergence. While Algorithm  \ref{alg:lagrange} presents one form of stopping criterion, other criterion can be implemented in conjunction with block halving to improve performance. In practice, the approximate value functions generated from this algorithm can be used in a look ahead policy, to ensure that actions taken in the present period satisfy the budget constraint.

\section{Comparison with Exact Solution}
While in general, the CHW problems of interest have far too many participants to solve optimally with dynamic programming, in this section we consider a small 2 patient problem with a budget of 1 across 
 5 periods to compare the performance of our various approximation methods against the optimal policy. The parameters we used for the two patients are given in Table \ref{tab:exact_params}
\begin{table}
\caption{Parameters of patients used in exact policy comparison.}
\label{tab:exact_params}
\centering
\resizebox{\textwidth}{!}{
 \begin{tabular}{|c|ccccccccccc|}
 \hline
     Patient Number & $p$ & $\mu$ & $\alpha$ & $\theta$ & $\lambda$ & $s_0$ & $\beta$ & $\gamma$ & $\rho$ & $b$ & $z$  \\ \hline
     0 & 0.2666645031 &  0.02316177714 & 0.3174730391 & 0.005663521508 & 0.001816217979 & 0.9097943152 & 1.66084819 & 0.2 & 0.2 & 5.308267697 & 0  \\
     1 & 0.06650739423 & 0.02148926302 &  0.088861680261 & 0.005867785106 & 0.00133048806 & 0.9097943152 & 1.66084819 & 0.2 & 0.2 & 5.159055299 & 0 \\ \hline
 \end{tabular}}
\end{table}
Using this setting we ran each of our heuristics and compared them to the optimal policy (which is easy to compute in this setting using enumeration). For each algorithm we ran 10 replicates and computed the average number of patient periods in control (PPC) and the standard deviation in Table \ref{tab:toy_problem_comparison_results}. Interestingly in this case  where the number of patients is so small it is hard to distinguish between the methods enhanced by EA and other heuristics. However, EA with descending Value/Visits and descending Value tend to perform closest to the optimal policy. What is also clear is that the visit everyone policy is far from optimal for this case. Both the Lagrangian and Whittle's index approaches do not perform as well in this setting, which is to be expected. Both of these methods are better suited for larger patient populations where the duality gap is reduced.

 \begin{table}
 \caption{Patient periods in control and standard deviation for various policies in the 2 patient 5 period problem.}
     \label{tab:toy_problem_comparison_results}
     \centering
     
 \begin{tabular}{|c|cc|}
 \hline
          Policy& Average PPC & Standard Deviation  \\ \hline
          Optimal Policy& 6.2 & 2.14 \\
          Ascending FBG & 3.8 & 2.57 \\
        Descending FBG & 4.1 & 2.56 \\
 EA ascending FBG & 3.8 & 2.57 \\
 EA descending FBG & 4.1 & 2.56 \\
 EA value-to-go & 4.2 & 2.66 \\
 EA value-to-go/visits & 4.2 & 2.66 \\
 EA Lagrangian & 3.7 & 2.54 \\
 EA Whittle's Index & 3.7 & 2.54 \\
 Visit everyone & 4.3 & 2.71 \\
 Visit no one & 3.7 & 2.54 \\ \hline
 \end{tabular}  
 \end{table}

\end{document}